%% file: main.tex
\documentclass[10pt,journal,compsoc]{IEEEtran}

\ifCLASSOPTIONcompsoc
  \usepackage[nocompress]{cite}
\else
  \usepackage{cite}
\fi

\usepackage[caption=false,font=footnotesize,labelfont=rm,textfont=rm]{subfig}
\usepackage{textcomp}
\usepackage{stfloats}
\usepackage{url}
\usepackage{verbatim}
\usepackage{graphicx}
\usepackage{multirow}
\usepackage{amsmath,amssymb,amsfonts}
\usepackage[linesnumbered,ruled,vlined]{algorithm2e}
\usepackage{float}
\usepackage{booktabs}
\usepackage{bm}
\usepackage{mathtools}
\usepackage{amsthm}
\usepackage{pifont}
\usepackage{xcolor}
\usepackage{enumitem}
\usepackage{balance}
\usepackage{hyperref}
\usepackage{cleveref}
\usepackage{tabularx}
\usepackage{adjustbox}

\usepackage[utf8]{inputenc}
\usepackage{booktabs}
\usepackage{array}
\usepackage{multirow}
\usepackage{makecell}
\usepackage{xcolor}
\usepackage{colortbl}
\usepackage{graphicx}

\newcolumntype{L}[1]{>{\raggedright\arraybackslash}p{#1}}
\newcolumntype{C}[1]{>{\centering\arraybackslash}p{#1}}

\definecolor{ieeeblue}{RGB}{0,72,138}
\definecolor{bandgray}{RGB}{240,240,240}
\definecolor{highlight}{RGB}{226,236,248}

\renewcommand{\arraystretch}{1.25}

\newtheorem{theorem}{Theorem}

\newtheorem{lemma}[theorem]{Lemma}
\newtheorem{remark}{Remark}
\newtheorem{assumption}{Assumption}

\newcommand{\cL}{\mathcal{L}}
\newcommand{\cD}{\mathcal{D}}

\newcommand{\cO}{\mathcal{O}}
\newcommand{\R}{\mathbb{R}}

\newcommand{\xmark}{\ding{55}}
\newcommand{\E}{\mathbb{E}}

\definecolor{clcolor}{RGB}{31,119,180}    
\definecolor{flcolor}{RGB}{44,160,44}     
\definecolor{hycolor}{RGB}{214,39,40}     
\newcommand{\CL}[1]{\textcolor{clcolor}{#1}}
\newcommand{\FL}[1]{\textcolor{flcolor}{#1}}
\newcommand{\HY}[1]{\textcolor{hycolor}{#1}}

\hypersetup{
    colorlinks=true, 
    linkcolor=blue,  
    filecolor=blue,  
    urlcolor=black,   
    citecolor=blue,  
    pdfborder={0 0 0} 
}

\newcommand{\verticaltext}[3][0pt]{%
  \raisebox{#1}{\parbox[t]{1em}{\rotatebox[origin=c]{90}{#2 #3}}}%
}

\Crefname{figure}{Fig.}{Figs.}
\Crefname{table}{Tab.}{Tabs.}
\Crefname{algorithm}{Algo.}{Algos.}
\Crefname{equation}{Eq.}{Eqs.}

\providecommand{\BIBentryALTinterwordspacing}{\spaceskip=\fontdimen3\font plus 0.63\fontdimen3\font minus 0.33\fontdimen3\font}
\providecommand{\BIBentrySTDinterwordspacing}{\spaceskip=0pt\relax}

\begin{document}

\title{
OmniISR: A Unified Framework for Centralized and Federated Learning via Intermediate Supervision and Regularization
}

\author{Wei-Bin Kou, Guangxu Zhu, Ming Tang, Chen Zhang, Lisheng Wu, Lei Zhou, Yujiu Yang$^*$%
\IEEEcompsocitemizethanks{
\IEEEcompsocthanksitem Wei-Bin Kou and Yujiu Yang are with Tsinghua Shenzhen International Graduate School, Tsinghua University, Shenzhen, China
\IEEEcompsocthanksitem Guangxu Zhu is with Shenzhen Research Institute of Big Data, Shenzhen, China.
\IEEEcompsocthanksitem Ming Tang is with the Department of Computer Science and Engineering, Southern University of Science and Technology, Shenzhen, China.
\IEEEcompsocthanksitem Chen Zhang is with the Department of Electrical and Electronic Engineering, The University of Hong Kong, Hong Kong, China.
\IEEEcompsocthanksitem Lisheng Wu and Lei Zhou are with Yinwang Intelligent Technology Co. Ltd., Shenzhen, China
\IEEEcompsocthanksitem Corresponding authors: Yujiu Yang.}%
}

\markboth{IEEE Transactions on Pattern Analysis and Machine Intelligence, 2026}%
{Kou \MakeLowercase{\textit{et al.}}: On the Role of Intermediate Supervision in Deep and Federated Learning}

\IEEEtitleabstractindextext{%
\begin{abstract}
The global deployment of edge intelligence (e.g., autonomous driving) operates across heterogeneous legal frameworks. While some regions permit centralized learning (CL) via cloud data aggregation, others enforce strict data localization, necessitating federated learning (FL). 
This operational dichotomy introduces two fundamentally incompatible optimization regimes (i.e., unbiased global gradients yet coupled with internal covariate shift in CL versus biased, drift-prone local updates in FL), resulting in that any naive integration of the two lacks rigorous theoretical guarantees.
To fill this gap, we propose OmniISR, a unified framework that fuses pure CL, pure FL, and hybrid CL--FL training modes via equipping intermediate supervision and regularization (ISR) signals at multiple hidden layers. Specifically, we propose (i) to use mutual-information (MI) as intermediate supervision to align shifting internal covariate in CL and client-drifting representations in FL, and (ii) to adopt negative-entropy (NE) as intermediate regularizer to penalize overconfident prediction, preserve representational uncertainty, and avoid device-specific collapse. 
On the theory side, we derive (i) a unified, ISR-agnostic, and non-asymptotic \(\mathcal{O}(1/\sqrt{T})\) convergence bound that shows the introduced ISR does not violate standard SGD convergence, (ii) a federated drift-bound that quantifies the ISR-reduced client drift, (iii) a gradient-alignment guarantee that ensures non-conflicting CL and FL updates under mild bias, and (iv) an explicit escape-time bound that indicates that CL--FL hybrid mixing enlarges effective stochasticity and accelerates escape from strict saddles.
Extensive experiments across multiple model architectures, datasets, and FL algorithms demonstrate that OmniISR consistently improves model performance in both centralized and federated paradigms, reduces the CL–FL gap by 22.60\%, and yields 37/48 paired metric wins across multiple FL algorithms.
\end{abstract}

}

\maketitle
\IEEEdisplaynontitleabstractindextext
\IEEEpeerreviewmaketitle

\ifCLASSOPTIONcompsoc
\IEEEraisesectionheading{\section{Introduction}\label{sec:intro}}
\else
\section{Introduction}\label{sec:intro}
\fi
\IEEEPARstart{T}{he} advent of edge intelligence has revolutionized large-scale distributed systems, with autonomous driving (AD) serving as a paramount application \cite{10205502,10891659,10007064}. To continuously refine AD models, AD fleets require to collect vast amounts of driving data. However, the data collection is increasingly constrained by divergent data governance and privacy regulations. In certain jurisdictions, data can be aggregated in the cloud for centralized learning (CL). Conversely, in regions governed by stringent privacy frameworks, such as the General Data Protection Regulation (GDPR) in the European Union and the Data Security Law in China, raw data is classified as sensitive information and is strictly prohibited from cross-device or cross-border transmission. In these regulatory domains, Federated Learning (FL) \cite{mcmahan2017fedavg,10696955,11429609} emerges as the legally compliant methodology for model enhancement. Consequently, globally deployed edge intelligence systems must operate under a compatible training paradigm. This necessitates a unified optimization framework that seamlessly integrates CL and FL to ensure consistent and robust model performance across diverse international markets.

However, building such a unified framework is far from straightforward. The two paradigms differ fundamentally in their data distribution and optimization dynamics. Naively alternating between centralized and federated updates without principled coordination offers no convergence guarantees and may introduce pathological gradient interference. A rigorous unification must confront three deeply intertwined challenges that span both the theoretical and structural dimensions of the learning process.

\textbf{First, fundamentally divergent optimization dynamics between CL and FL impede the unified training.} In pure CL, data is generally assumed to be independent and identically distributed (IID), allowing optimizers (e.g., Adam, SGD) to descend smoothly along the loss landscape. In contrast, FL in edge scenarios is characterized by highly non-IID data distributions \cite{11173936,10908045} and system heterogeneity (e.g., stragglers) \cite{wang2023fluid,11127200,11162622}. Formulating a unified mechanism requires rigorously answering: \textit{How can we mathematically guarantee convergence across such disparate optimization landscapes?} Furthermore, it is imperative to prove that the exact gradients derived from centralized data do not engage in "gradient conflict" with the aggregated pseudo-gradients from federated clients, but rather operate synergistically to accelerate to escape local optima. 

\begin{figure*}[!t]
\includegraphics[width=\linewidth]{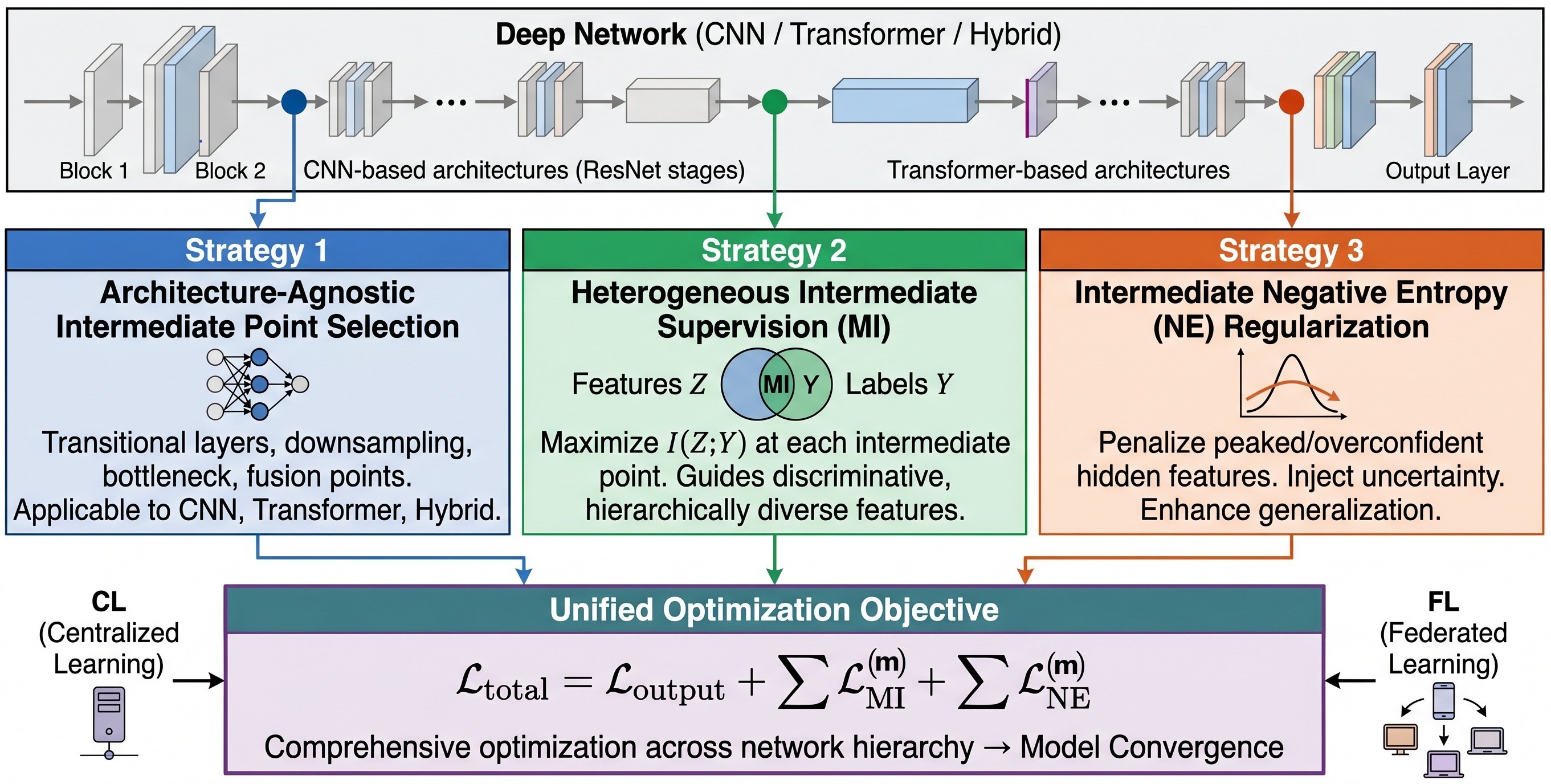}
\vspace{-0.7cm}
\caption{Overview of the mechanism of intermediate supervision and regularization in the proposed OmniISR framework.}
\label{Fig:CL_paradigm}
\vspace{-0.5cm}
\end{figure*}

\textbf{Second, non-IID data exacerbates the independent drift of latent representations across distributed clients in FL} \cite{karimireddy2020scaffold}. In deep learning architectures, supervision is exclusively applied at the output layer. While this single-point supervision suffices for centralized IID data, in a federated non-IID context, the lack of explicit constraints on hidden layers causes the latent features of different clients to drift independently. This representation drift severely degrades the aggregated global model's performance, an issue that output-layer-only supervision is mathematically ill-equipped to resolve. 

\textbf{Third, mere intermediate supervision compromises generalization.} A seemingly intuitive solution to representation drift is the introduction of supervision at  intermediate layers to anchor the hidden feature. However, enforcing strict intermediate supervision forces the hidden representations to become highly scenario- or task-informed. This deterministic feature alignment reduces the model's flexibility, causing it to overfit to the specific distributions of the training clients and drastically reducing its generalization capability in unseen scenarios.

To systematically address these three challenges, we propose OmniISR, a unified training framework for edge intelligence that seamlessly operates across pure CL, pure FL, and hybrid CL--FL paradigms. The core design principle of OmniISR is that deep networks deployed with heterogeneous distributions require explicit, diversified guidance at hidden layers rather than merely at the output. Concretely, OmniISR embodies intermediate supervision and regularization (ISR) mechanism that contains following three integral strategies. 

\begin{enumerate}
    \item \textbf{Architecture-Agnostic Intermediate Layer Selection:} we select multiple ISR layers within the network based on architecture-agnostic criteria, such as transitional layer between blocks (e.g., ResNet stages or Transformer layers), downsampling layers, bottleneck layers, or feature fusion layers, ensuring applicability across CNN-based, Transformer-based, and hybrid architectures without requiring architecture-specific redesign.
    \item \textbf{Heterogeneous Supervision at Intermediate Layers:} at each selected intermediate point, OmniISR computes the mutual information (MI) between the latent features and ground-truth, using this as a heterogeneous supervision signal distinct from the output layer's cross-entropy objective. By maximizing the shared information between hidden representations and task labels, MI supervision guides intermediate layers to learn discriminative yet hierarchically diverse features, effectively constraining representation drift without compromising to learn premature hidden features.
    \item \textbf{Negative Entropy (NE) Regularization on Intermediate Activations:} OmniISR imposes NE regularization on the latent activation distributions at each selected intermediate layer, explicitly penalizing peaked, overconfident hidden features and injecting necessary uncertainty and significantly enhancing its generalization to unseen scenarios.
\end{enumerate}
These three integral strategies are illustrated in \Cref{Fig:CL_paradigm}.

\begin{figure*}[tp]
\centering
\includegraphics[width=\linewidth]{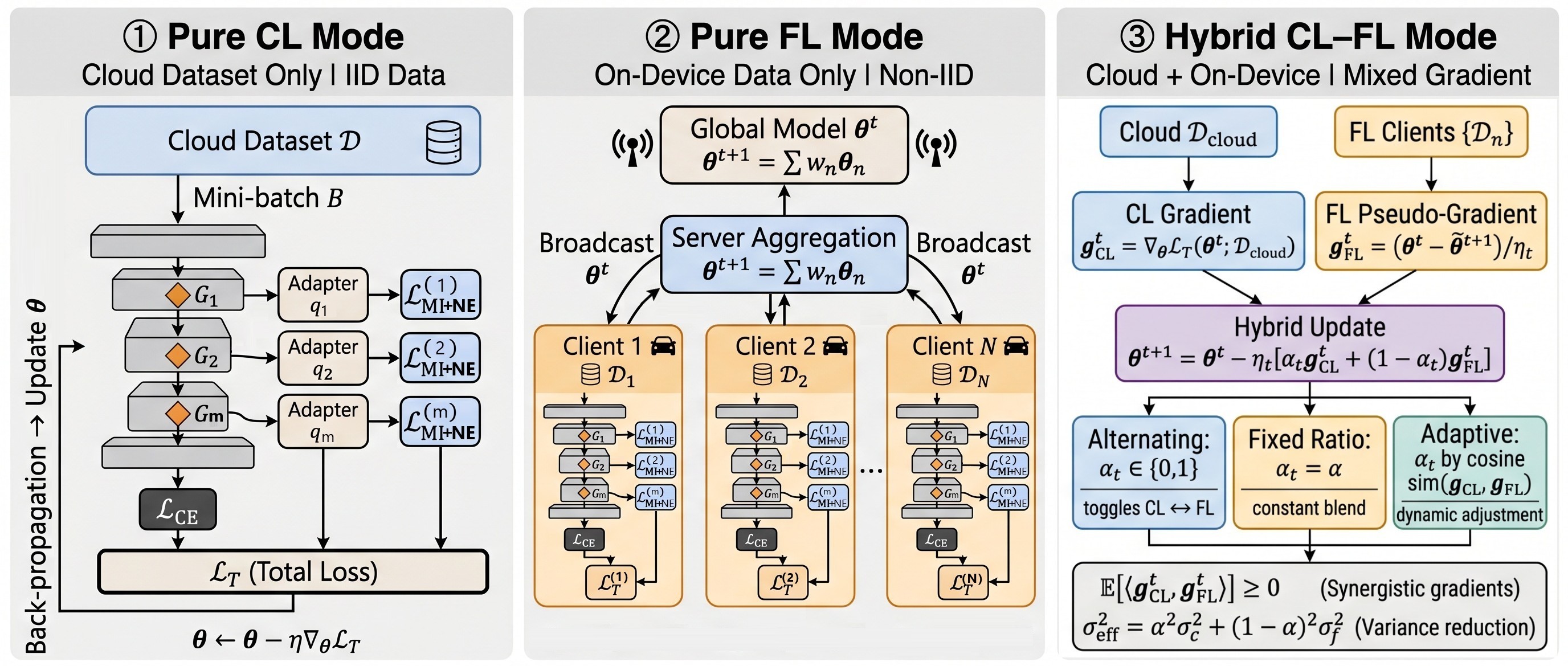}
\vspace{-0.7cm}
\caption{Illustration of the three modes of the proposed OmniISR framework.}
\label{Fig:FL_paradigm}
\vspace{-0.5cm}
\end{figure*}

The proposed OmniISR operates aforementioned ISR mechanism across three modes: Pure CL, Pure FL, and Hybrid CL--FL. In the CL mode, all intermediate losses and regularizers are combined with the output-layer loss into a single objective, which is optimized until convergence. In the FL mode, training proceeds over multiple communication rounds. Within each round, every client first minimizes a weighted combination of intermediate losses, regularizers, and the final loss on its private data for several local iterations, after which the central server aggregates the updated local models across all participants without exposing raw data. In the Hybrid CL--FL mode, each update blends an exact cloud gradient with a federated pseudo-gradient through a mixing weight, scheduled via one of three strategies: alternating between pure CL and FL rounds, a fixed mixing ratio, or an adaptive weight adjusted by gradient similarity. Beyond leveraging larger data volumes, this hybrid procedure combines centralized and federated noise sources, increasing effective stochasticity and thereby helping the model escape saddle points and explore the loss landscape more effectively. The three modes are illustrated in \Cref{Fig:FL_paradigm}.

On the theoretical front, we derive a unified, ISR-independent, and non-asymptotic \(\mathcal{O}(1/\sqrt{T})\) convergence bound for OmniISR across CL, FL, and hybrid modes, proving that OmniISR maintains standard non-convex SGD rates. Our analysis explicitly characterizes how finite-time constants scale with the number of intermediate points \(M\) and their associated weights \(\{\alpha_m, \lambda_m\}_{m=1}^M\). In the FL setting, we establish an \(\mathcal{O}(E^{2}H)\) drift bound, where \(E\) denotes local epochs and \(H\) quantifies non-IID data heterogeneity, and demonstrate that ISR reduces effective client drift by stabilizing hidden representations. For the hybrid mode, the bound isolates an explicit bias floor \(B_{\mathrm{eff}}\) caused by cloud--device representativeness gaps, alongside an effective variance \(\sigma_{\mathrm{eff}}^{2}\) resulting from mixed noise sources. Furthermore, we prove that CL and FL gradients satisfy \(\mathbb{E}[\langle \mathbf{g}_{\mathrm{CL}}, \mathbf{g}_{\mathrm{FL}} \rangle] \geq 0\) under mild assumptions on the overlap between centralized and federated data distributions. This confirms that both gradient sources cooperate to escape local optima, providing first formal justification for mixed-paradigm training. Ultimately, these theoretical guarantees translate hyperparameters (e.g., \(M\), \(E\), and the mixing weight \(\alpha\)) into actionable design principles for stable and unified training.

In summary, the main contributions of this work are highlighted as follows:
\begin{itemize}
    \item We propose OmniISR, a unified optimization framework for CL, FL, and hybrid CL--FL training under one objective, with architecture-agnostic introduction of ISR. This directly targets real deployments where training mode is policy-dependent across regions.
    \item We introduce a coupled intermediate design that combines heterogeneous MI supervision (not output-layer CE replication) and NE regularization. This coupling is intended to jointly address two competing requirements in non-IID training: representation-drift suppression and generalization preservation.
    \item Theoretically, we demonstrate that (i) OmniISR guarantees ISR-not-violated SGD convergence regardless of working modes, (ii) OmniISR reduces heterogeneity-caused client-drift, (iii) cloud and on-device hybrid updates operate synergistically rather than destructively, and (iv) the combined stochasticity of the hybrid CL--FL accelerates to escape suboptimal saddle points.
    \item Empirically, we evaluate OmniISR across multiple model architectures, datasets, and FL algorithms. Beyond absolute gains, OmniISR narrows the CL--FL performance gap by 22.60\%, shows broad cross-FL-algorithm positive transferability with 37/48 metric wins in paired comparisons, and clarifies how intermediate point number, spacing, and placement impact OmniISR's effectiveness via comprehensive ablations.
\end{itemize}

The remainder of this paper proceeds as follows. \Cref{sec:related} reviews related works. \Cref{sec:methodology} details the proposed OmniISR and its theoretical guarantees. \Cref{sec:experiments} presents experiments and ablations. \Cref{sec:conclusion} concludes this paper.

\input{related_works}

\input{methodology}

\input{experiments}


\end{document}

%% file: related_works.tex
\section{Related Works}
\label{sec:related}

\subsection{Centralized Learning Optimization}
Centralized learning (CL) optimization encompasses a broad suite of algorithms for minimizing loss functions over high-dimensional parameter spaces.  Stochastic gradient descent (SGD) and back-propagation lay the groundwork~\cite{refinetti2023sgd}, while adaptive methods such as Adam~\cite{kingma2014adam} and RMSprop provide adaptive learning rates that improve convergence stability across architectures.  Regularization strategies, including dropout~\cite{hinton2012dropout}, L1/L2 penalties~\cite{tibshirani1996lasso}, and sharpness-aware minimization~\cite{foret2021sam}, are critical for preventing overfitting and ensuring robust generalization.  
Normalization techniques such as batch normalization~\cite{ioffe2015batch} and layer normalization~\cite{ba2016layer} further stabilize training dynamics.

Despite these advances, training very deep networks still faces gradient gradually weakening~\cite{guo2024gradient,hochreiter2001gradient} and under-optimized intermediate features~\cite{liu2024detection}, particularly when supervision is applied solely at the output layer.  This motivates the study of intermediate supervision, discussed next.

\subsection{Federated Learning Optimization}
Federated Learning (FL) \cite{10833756,11314751,10255290} enables distributed devices to collaboratively train a shared model while keeping raw data private~\cite{mcmahan2017fedavg}.  FedAvg~\cite{mcmahan2017fedavg} is the de-facto baseline, where clients perform local SGD and upload updates for server-side aggregation.  However, the non-IID nature of distributed data, arising from diverse geographic environments, weather conditions, and traffic patterns, causes significant performance degradation and slow convergence.

A rich line of work has been proposed to mitigate these challenges.  FedProx~\cite{li2020fedprox} adds a proximal term to penalize local deviation from the global model.  SCAFFOLD~\cite{karimireddy2020scaffold} introduces control variates to correct client drift.  FedDyn~\cite{acar2021feddyn} uses dynamic regularization.  FedAvgM~\cite{hsu2019fedavgm} applies server-side momentum.  FedIR~\cite{hsu2020fedir} addresses class imbalance through importance reweighting.  MOON~\cite{li2021moon} leverages model-contrastive learning.  BalanceFL~\cite{9825928} addresses long-tail class imbalance.  In the AD-specific context, FedRC~\cite{kou2024fedrc} and FedGau~\cite{kou2025fedgau} accelerate hierarchical FL convergence; FedEMA~\cite{kou2025fedema} integrates exponential moving averaging with negative entropy regularization; pFedLVM~\cite{kou2025pfedlvm} personalizes federated models via large vision model features; and FedDrive~\cite{fantauzzo2022feddrive} generalizes FL to semantic segmentation in AD.  Communication-constrained and hierarchical FL for AD has also been studied~\cite{kou2023communication}, along with contrastive-divergence approaches to non-IID mitigation~\cite{do2024reducing}.

Despite these algorithmic advances, existing FL methods address non-IID challenges primarily through aggregation-level or loss-level corrections \emph{at the output layer}.  None of these works introduces supervision and regularization at intermediate layer within the federated training loop.  OmniISR fills this gap.

\subsection{Intermediate Supervision}
Intermediate supervision \cite{lee2015deeply} augments output-layer-only supervision by injecting auxiliary losses at intermediate layers, thereby providing explicit guidance for hidden representations.  Early instances include GoogleNet's two auxiliary classifiers at intermediate stages~\cite{szegedy2015going} and DSN's auxiliary supervision branches~\cite{wang2015dsn}.  Subsequent work has extended the paradigm to dense-prediction tasks. For example, PSPNet~\cite{zhao2017pspnet} adds an auxiliary classifier for pixel-wise cross-entropy on pyramid pooling features. BiSeNet~\cite{yu2018bisenet} applies supervised branches to spatial and context paths. Gated-SCNN~\cite{takikawa2019gated} introduces shape-based intermediate losses. ICNet~\cite{zhao2018icnet} attaches auxiliary losses to low-resolution intermediate predictions in a cascaded framework.  More recently, contrastive intermediate supervision~\cite{zhang2022contrastive} has been explored, and a comprehensive review of intermediate supervision theories and applications is provided in~\cite{li2022deepsupervision}.

While intermediate supervision has proven effective, three limitations still persist.  First, existing techniques are tightly coupled with specific model architectures, lacking generality across CNN-based and Transformer-based model architectures. Second, the auxiliary losses applied to intermediate layers are typically \emph{identical} to the output-layer loss (e.g., cross-entropy), which forces intermediate layers to prioritize output-specific features prematurely and limits the learning of generalizable representations.  Third, no explicit regularization is imposed on hidden activations, risking overconfident predictions that degrade out-of-distribution generalization.  OmniISR addresses all three issues by introducing architecture-agnostic intermediate point selection, heterogeneous mutual-information supervision, and negative-entropy regularization.

\subsection{Unified Centralized and Federated Learning}
A fundamental yet under-explored challenge in practical edge intelligence is the coexistence of centralized and federated training paradigms.  In globally deployed systems (for instance, AD fleets operating across jurisdictions with different data-privacy regulations), some regions permit cloud-based centralized training while others mandate strictly local, federated training.  This bifurcation necessitates a framework that can seamlessly support both modes under a single optimization formulation.
To the best of our knowledge, no prior work has provided such a unified framework with formal convergence guarantees.  Existing methods treat centralized and federated learning as separate paradigms, each with its own optimization analysis and algorithmic pipeline.  The potential synergy between exact centralized gradients and aggregated federated pseudo-gradients has not been formally analyzed.

The proposed OmniISR bridges this gap. The novelty of OmniISR is not a standalone use of intermediate supervision and intermediate regularization in isolation. The contribution lies in their \emph{joint coupling across paradigms}: (i) heterogeneous intermediate MI supervision (not output-layer CE replication), (ii) intermediate NE uncertainty regularization, and (iii) a unified CL/FL/hybrid optimization with fundamentally theoretical guarantees. This three-fold coupling defines the technical distinction from prior FL corrections and architecture-specific intermediate supervision designs.

%% file: methodology.tex
\section{Methodology}
\label{sec:methodology}

We first introduce the key notations of the proposed OmniISR framework in \Cref{tab:notation}, and then present the ISR mechanism in \Cref{sec:framework_ISR}, the three working modes of OmniISR in \Cref{sec:three_modes}, the theoretical guarantees in \Cref{sec:convergence}, and its $\epsilon$-stationarity complexity analysis in \Cref{sec:complexity_analysis}. Finally, we compare OmniISR's working modes in \Cref{sec:modes_comparison}.

\begin{table}[t]
\centering
\setlength{\tabcolsep}{3.5pt}
\caption{Key notations used throughout this paper}
\vspace{-0.2cm}
\label{tab:notation}
\renewcommand{\arraystretch}{1.15}
\begin{tabularx}{\linewidth}{ll}
\toprule
Symbol & Description \\
\midrule
\(\theta \in \R^d\) & Global model parameters \\
\(\theta_t\) & Model parameters at iteration/round \(t\) \\
\(D\) & The depth (layer number) of model \(\theta\)   \\
\(\cD\) & Centralized training dataset \\
\(\cD_n\) & Local private dataset on client \(n\) \\
\(N\) & Number of clients in FL \\
\(M\) & Number of intermediate supervision points \\
\(G_m\) & \(m\)-th intermediate point (layer index) \\
\(z^m_i = G_m(\theta; x_i)\) & Latent feature map at point \(G_m\) for input \(x_i\) \\
\(q^m(\cdot\,;\varphi_m)\) & Lightweight dimension adapter at \(G_m\) \\
\(C_m, w_m, h_m\) & Channel count, width, and height of \(z^m\) \\
\(K\) & Number of semantic classes \\
\(\cL_{\mathrm{CE}}\) & Output-layer cross-entropy loss \\
\(\cL_{\mathrm{MI}}^{(m)}\) & MI-based intermediate supervision at \(G_m\) \\
\(\cL_{\mathrm{NE}}^{(m)}\) & Negative entropy regularizer at \(G_m\) \\
\(\alpha_m, \lambda_m\) & Weights for MI loss and NE regularizer at \(G_m\) \\
\(E\) & Number of local training epochs per FL round \\
\(T\) & Total CL training iterations or FL rounds \\
\(w_n = |\cD_n|/\sum_k |\cD_k|\) & Aggregation weight for client \(n\) \\
\(\alpha_t \in [0,1]\) & Hybrid CL--FL mixing weight at round \(t\) \\
\(g_{\mathrm{CL}}^t, g_{\mathrm{FL}}^t\) & Cloud gradient and federated pseudo-gradient \\
\(\eta, \eta_t\) & Learning rate (constant or scheduled) \\
\bottomrule
\end{tabularx}
\vspace{-0.3cm}
\end{table}

\subsection{The Proposed ISR Mechanism}
\label{sec:framework_ISR}

The core philosophy of OmniISR is that deep networks deployed across CL and FL paradigms require \emph{explicit, diversified guidance at hidden layers} rather than mere supervision at the output. This intermediate guidance is supposed to balance \textbf{representation alignment} (constraining drift across clients or training stages) and \textbf{representational flexibility} (preserving the uncertainty needed for generalization). OmniISR instantiates this principle through three integral strategies: (i) architecture-agnostic intermediate layer selection, (ii) intermediate heterogeneous mutual-information (MI) supervision, and (iii) intermediate negative-entropy (NE) regularization. We elaborate each below.

\subsubsection{Architecture-Agnostic Intermediate Layer Selection}
\label{sec:point_selection}

Let the network \(f_\theta\) be composed of \(D\) layers. We select \(M\) ISR points \(\{G_1,\dots,G_M\}\) (\(M \ll D\)) at key architectural transition boundaries. Unlike prior intermediate supervision methods that are tightly coupled to a specific backbone (e.g., ICNet's cascaded branches~\cite{zhao2018icnet}), OmniISR defines \emph{architecture-agnostic} selection criteria:
\begin{itemize}[leftmargin=*,itemsep=2pt]
    \item \textbf{Scale transitions:} before or after spatial downsampling (pooling, strided convolutions), capturing changes in spatial resolution and feature granularity.
    \item \textbf{Block boundaries:} between major computational blocks (e.g., ResNet stages, Transformer encoder layers), leveraging hierarchical feature abstraction.
    \item \textbf{Bottleneck layers:} where feature dimensionality is compressed, highlighting the critical information pathways.
    \item \textbf{Attention or fusion points:} before/after attention mechanisms or multi-branch feature fusion, capturing how information is redistributed.
\end{itemize}
This design ensures that OmniISR is directly applicable to CNN-based architectures (e.g., DeepLabv3+~\cite{chen2018deeplabv3p}), Transformer-based architectures (e.g., TopFormer~\cite{zhang2022topformer}), and CNN-Transformer hybrid architectures (e.g., SeaFormer~\cite{wan2023seaformer}) without any architecture-specific redesign. In practice, we recommend \(M = \cO(\log D)\) to balance supervision granularity against introduced overhead.

\begin{figure*}[!t]
\includegraphics[width=\linewidth]{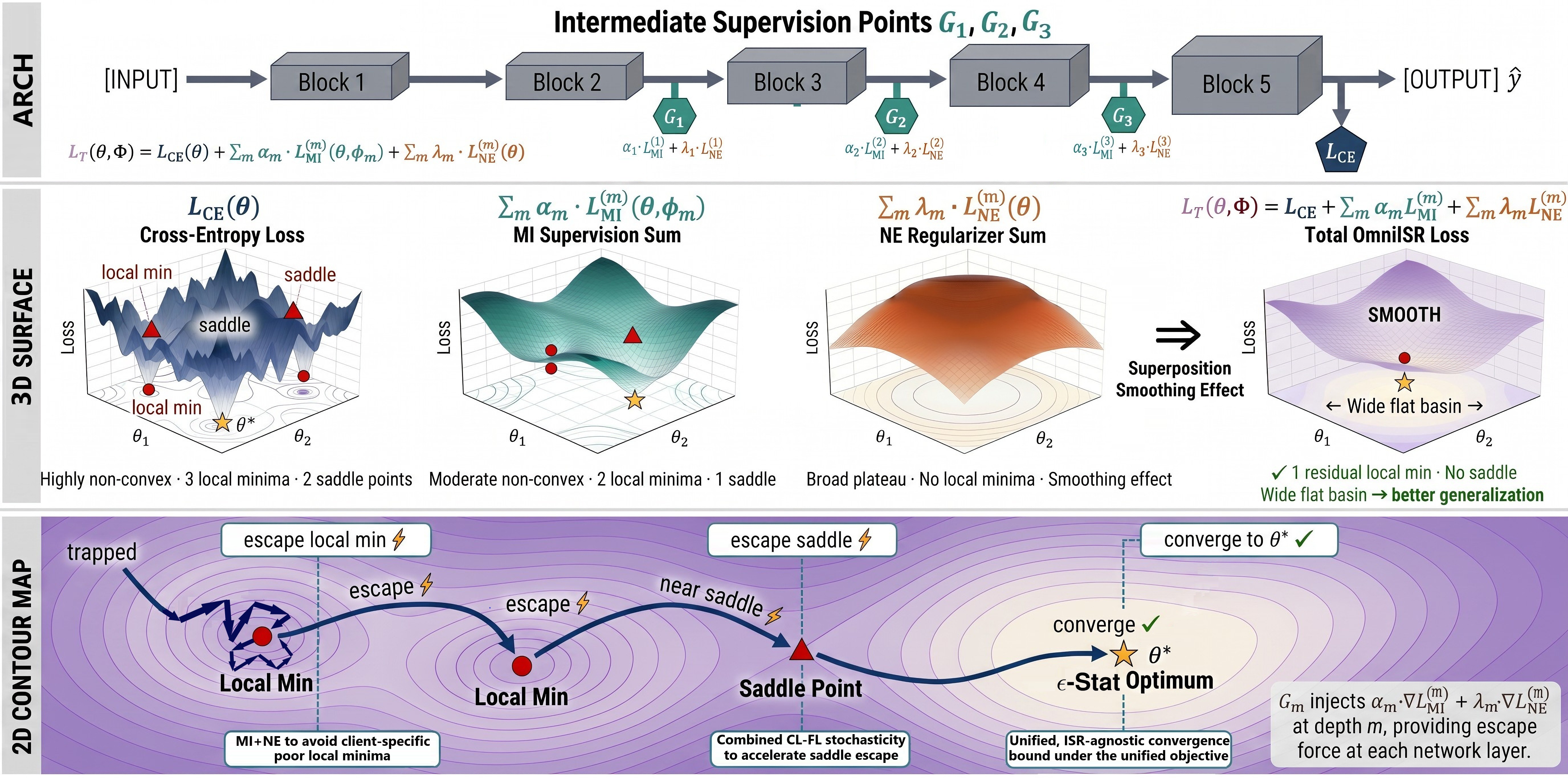}
\vspace{-0.7cm}
\caption{Illustration of benefits of ISR mechanism, supposing three intermediate points within the model.}
\label{Fig:benefits_ISR}
\vspace{-0.3cm}
\end{figure*}

For each selected point \(G_m\), we denote the latent feature map extracted from input \(x_i\) as
\begin{equation}\label{eq:latent_feature}
    z^m_i = G_m(\theta; x_i) \in \R^{C_m \times w_m \times h_m},
\end{equation}
where \(C_m\), \(w_m\), and \(h_m\) are the channel count, width, and height of the feature map at \(G_m\), respectively. Since \(z^m_i\) typically differs in spatial resolution and channel dimensionality from the ground truth \(y_i \in \{1,\dots,K\}^{W \times H}\), we attach a lightweight \emph{dimension adapter} \(q^m(\cdot\,; \varphi_m)\) at each point. The adapter consists of a \(1\times 1\) convolution followed by bilinear upsampling to \(W \times H\), producing class-probability maps \(q^m_k(z^m_{i,p}; \varphi_m)\) for each pixel \(p\) and class \(k\).

\subsubsection{Intermediate Mutual Information Supervision}
\label{sec:mi_supervision}

A fundamental limitation of classical intermediate supervision~\cite{lee2015deeply,szegedy2015going,wang2015dsn} is that it applies the \emph{same} loss function (typically cross-entropy) to both intermediate and output layers. While this provides gradient signal to hidden layers, it also forces intermediate representations to converge prematurely toward output-layer decision boundaries. From an information-theoretic perspective~\cite{shwartz2017opening}, an ideal intermediate representation should preserve maximal information about the task label \(Y\) while maintaining diverse feature structures that differ from the final prediction head. This motivates the use of a \emph{heterogeneous} intermediate loss that maximizes the shared information between hidden features and labels without collapsing onto the output-layer objective.

Let \(\Omega\) denote the set of all pixels in an image. The output-layer cross-entropy (CE) loss is defined as
\begin{equation}\label{eq:L_CE}
    \cL_{\mathrm{CE}}(\theta) = -\frac{1}{|\cD|}\sum_{(x_i,y_i)\in\cD}\sum_{p\in\Omega}\sum_{k=1}^{K} y_{i,p,k}\,\log\, p_k(x_{i,p};\theta),
\end{equation}
where $\mathcal{D}$ denotes the training dataset on a device, \(p_k(x_{i,p};\theta)\) is the output-layer softmax probability for class \(k\) at pixel \(p\).

For intermediate point \(G_m\), we define the MI supervision loss. Recall that the MI between the latent feature \(Z^m\) and label \(Y\) is \(I(Z^m;Y) = H(Y) - H(Y|Z^m)\), where \(H(Y)\) is a data-dependent constant. Maximizing \(I(Z^m;Y)\) is therefore equivalent to minimizing \(H(Y|Z^m)\). Since the true conditional \(P(Y|Z^m)\) is intractable, we introduce a variational approximation \(q^m(\cdot\,;\varphi_m)\) and optimize the variational upper bound on \(H(Y|Z^m)\)~\cite{barber2003im}:
\begin{equation}\label{eq:L_MI}
    \cL_{\mathrm{MI}}^{(m)}(\theta,\varphi_m) \!=\! -\frac{1}{|\cD|}\!\sum_{(x_i,y_i)\in\cD}\!\sum_{p\in\Omega}\!\sum_{k=1}^{K} y_{i,p,k}\,\!\log\, \!q^m_k\big(z^m_{i,p};\varphi_m\big).
\end{equation}
Although \Cref{eq:L_MI} superficially resembles a CE loss, it differs from \(\cL_{\mathrm{CE}}\) in three critical respects:
\begin{enumerate}[leftmargin=*,itemsep=2pt]
    \item \textbf{Operating on latent features instead of predictions:} \(\cL_{\mathrm{MI}}^{(m)}\) is computed on the intermediate feature map \(z^m\) passed through a small adapter \(q^m(\cdot\,;\varphi_m)\), rather than on the full network's output. This means the gradient signal flows directly into the hidden layers at point \(G_m\), providing localized supervision that does not traverse the entire downstream subnetwork.
    \item \textbf{Distinct parameterization:} The adapter \(q^m(\cdot\,;\varphi_m)\) has its own parameters \(\varphi_m\), decoupled from the output prediction head. This prevents the intermediate loss from merely replicating the output-layer objective and instead encourages \(G_m\) to learn representations that are \emph{independently discriminative}.
    \item \textbf{Hierarchical diversity:} Because each \(G_m\) operates at a different depth and spatial resolution, the MI losses collectively enforce a hierarchy of increasingly abstract yet task-relevant features, aligning with the information-theoretic principle of progressive information refinement~\cite{shwartz2017opening}.
\end{enumerate}

\subsubsection{Intermediate Negative Entropy Regularization}
\label{sec:ne_regularization}
While MI supervision ensures that intermediate features are well-optimized, it does not prevent them from becoming \emph{overconfident}, i.e., producing peaked, low-entropy distributions that are overly specialized to the training data distribution. This problem is amplified in FL, where each client's local training can push intermediate features toward client-specific overconfidence, exacerbating representation drift upon aggregation.

To counteract this tendency, we impose a NE regularizer on the latent activation distribution at each intermediate point. For point \(G_m\), the NE regularizer is defined as
\begin{equation}\label{eq:L_NE}
    \cL_{\mathrm{NE}}^{(m)}(\theta) = \frac{1}{|\cD|}\sum_{(x_i,y_i)\in\cD}\sum_{p\in\Omega}\sum_{c=1}^{C_m} p^m_c(z^m_{i,p};\theta)\,\log\, p^m_c(z^m_{i,p};\theta),
\end{equation}
where \(p^m_c(z^m_{i,p};\theta) = \mathrm{softmax}(z^m_{i,p})_c\) is the softmax probability over the \(C_m\) channels at pixel \(p\) of the latent feature. Minimizing \(\cL_{\mathrm{NE}}^{(m)}\) is equivalent to \emph{maximizing the entropy} of the channel-wise activation distribution, thereby penalizing peaked activations and injecting controlled uncertainty into hidden features.

\subsubsection{The MI--NE synergy}
The MI supervision and NE regularization play complementary roles that resolve a fundamental tension in intermediate layer optimization:
\begin{itemize}[leftmargin=*,itemsep=2pt]
    \item \textbf{MI supervision} pulls features toward \emph{discriminativeness}: it ensures that hidden representations carry sufficient information about the task labels, preventing under-optimized or task-irrelevant intermediate features.
    \item \textbf{NE regularization} pushes features toward \emph{uncertainty}: it prevents features from collapsing into overconfident, overly specialized representations.
\end{itemize}
Together, they carve out a ``sweet spot'' in the representation space, where features are \emph{well-optimized, task-informed but not task-overfit}. This synergy is the key design principle underlying OmniISR's effectiveness in both CL and FL settings.

\subsubsection{OmniISR's Unified Optimization Objective}
\label{sec:unified_objective}

Combining the output-layer loss with all intermediate terms, the total optimization objective of OmniISR at each device is
\begin{equation}\label{eq:total_loss}
    \cL_T(\theta,\Phi) = \cL_{\mathrm{CE}}(\theta) + \sum_{m=1}^{M}\Big(\alpha_m\,\cL_{\mathrm{MI}}^{(m)}(\theta,\varphi_m) + \lambda_m\,\cL_{\mathrm{NE}}^{(m)}(\theta)\Big),
\end{equation}
where \(\Phi = \{\varphi_1,\dots,\varphi_M\}\) denotes the collective adapter parameters and \(\alpha_m, \lambda_m > 0\) are weights controlling the relative contribution of MI supervision and NE regularization at each point \(G_m\).

The total loss \(\cL_T(\cdot, \cdot)\) induces a richer gradient landscape than output-layer-only training. The gradient with respect to model parameters $\theta$ decomposes as
\begin{equation}\label{eq:gradient_decompose}
    \nabla_\theta \cL_T = \nabla_\theta \cL_{\mathrm{CE}} + \sum_{m=1}^{M}\Big(\alpha_m\,\nabla_\theta\cL_{\mathrm{MI}}^{(m)} + \lambda_m\,\nabla_\theta\cL_{\mathrm{NE}}^{(m)}\Big).
\end{equation}
The intermediate gradient terms \(\nabla_\theta\cL_{\mathrm{MI}}^{(m)}\) and \(\nabla_\theta\cL_{\mathrm{NE}}^{(m)}\) inject supervision directly at \(G_m\), effectively shortening the back propagation path for layers preceding \(G_m\). This addresses the gradient vanishing problem while simultaneously providing diverse optimization signals that prevent all layers from converging to the same output-layer-centric features.
The benefits of ISR mechanism are illustrated in \Cref{Fig:benefits_ISR}.

\begin{figure*}[t]
\centering
\subfloat[\footnotesize Gradient drift comparison at intermediate layers]{\includegraphics[width=0.48\linewidth]{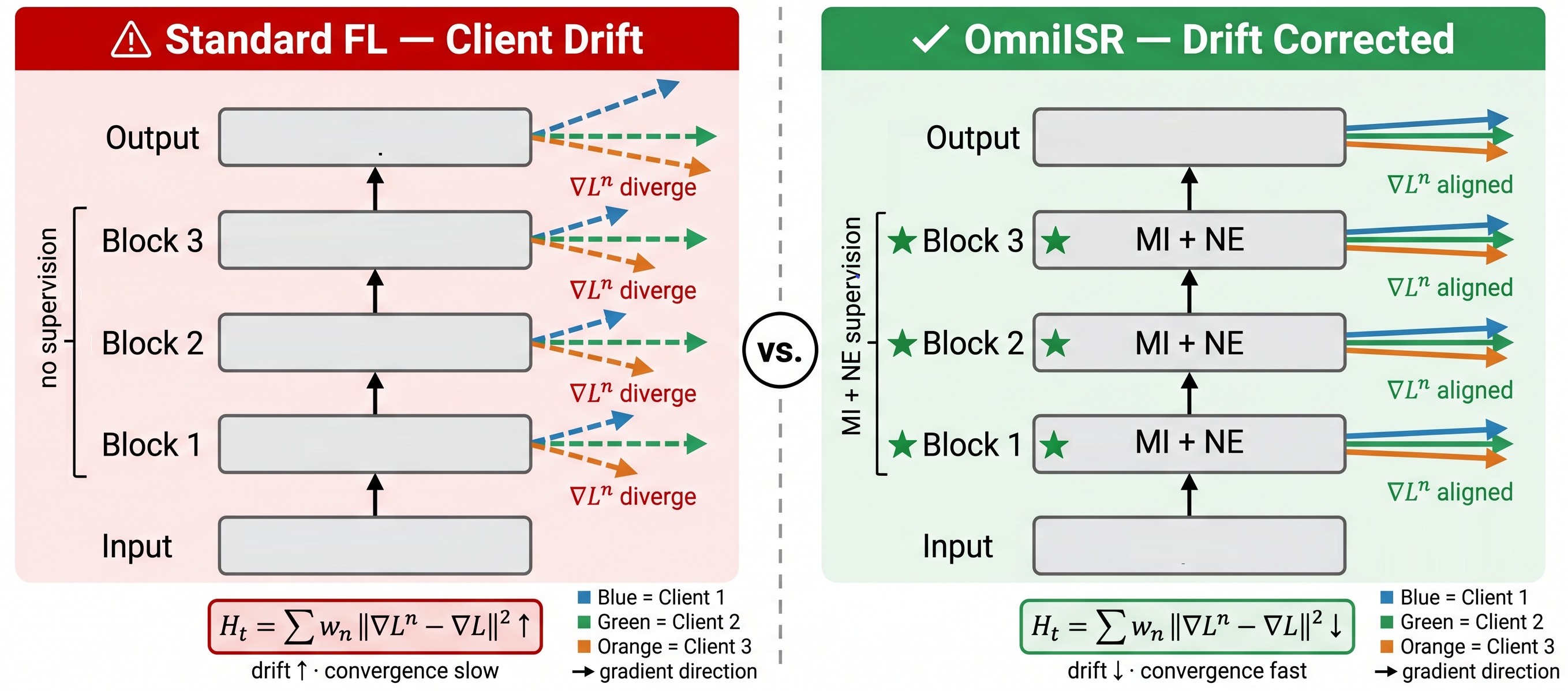}
\label{Fig.client_drift_arch}
}
\subfloat[\footnotesize Convergence trajectory comparison on objective landscapes]{\includegraphics[width=0.49\linewidth]{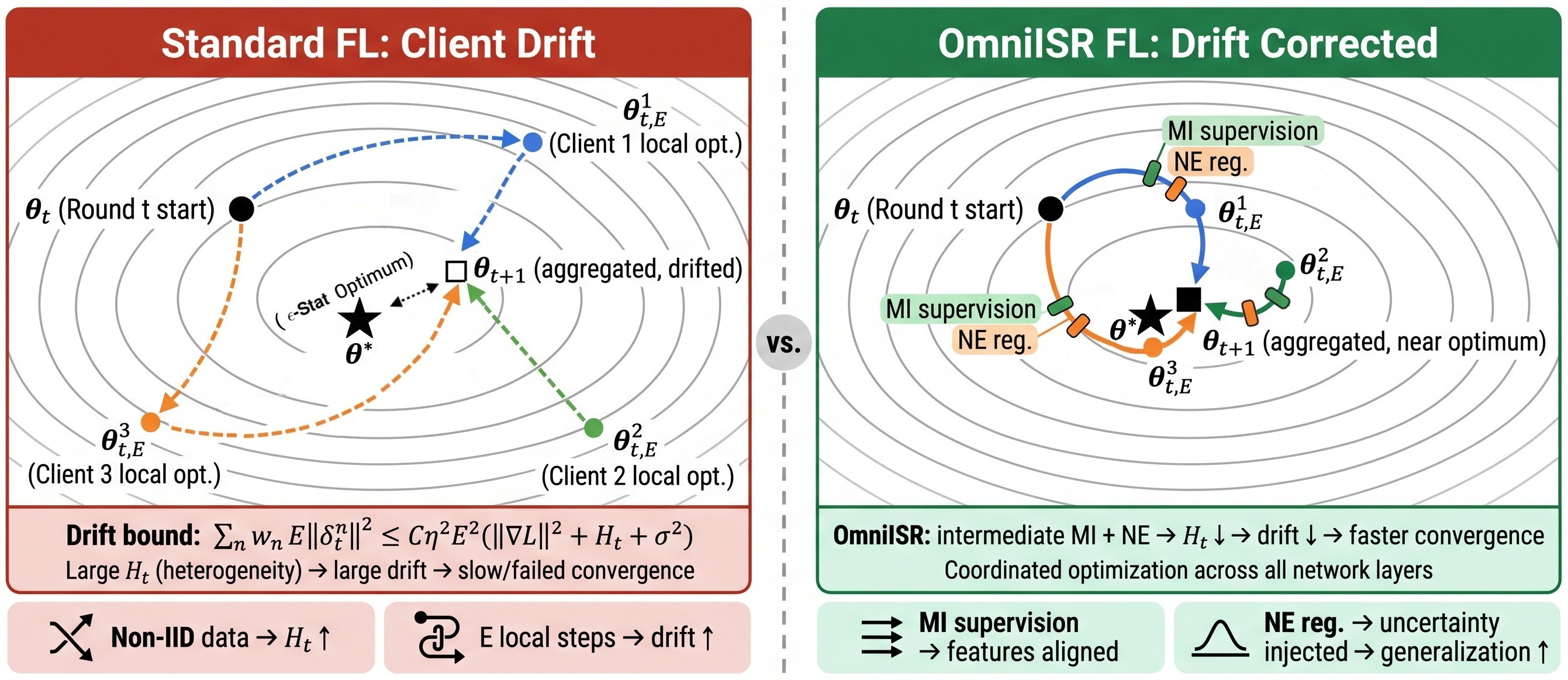}
\label{Fig.client_drift_landscape}
}
\vspace{-0.15cm}
\caption{Illustration of why the proposed OmniISR framework can reduce client drift in federated setting, taking three clients in this toy example.}
\label{Fig.overcome_FL_client_drift}
\vspace{-0.3cm}
\end{figure*}

\subsection{Three Working Modes of OmniISR} \label{sec:three_modes}
\subsubsection{Pure Centralized Training (CL Mode)}
\label{sec:cl_mode}

When a centralized dataset $\mathcal{D}$ is available, OmniISR optimizes $\mathcal{L}_T(\cdot,\cdot)$ via stochastic gradient descent until convergence. Concretely, each training iteration proceeds through four steps: (i)~\textbf{Forward Pass:} For an input image $x_i$, the network computes intermediate features $\{z_i^1, \dots, z_i^M\}$ and the final prediction $\hat{y}_i$. (ii)~\textbf{Loss Computation:} The total loss $\mathcal{L}_T$ is computed via \Cref{eq:total_loss}, i.e., aggregating the output-layer loss $\mathcal{L}_{\mathrm{CE}}$, the intermediate MI losses $\{\mathcal{L}_{\mathrm{MI}}^{(m)}\}_{m=1}^{M}$, and the NE regularizers $\{\mathcal{L}_{\mathrm{NE}}^{(m)}\}_{m=1}^{M}$. (iii)~\textbf{Back-Propagation:} Gradients of $\mathcal{L}_T$ are computed with respect to both the model parameters $\theta$ and the auxiliary dimension adapters $\{\varphi_m\}_{m=1}^M$. (iv)~\textbf{Parameter Update:} All parameters are updated using the Adam optimizer~\cite{kingma2014adam}:
\begin{align}
    &\theta \leftarrow \theta - \eta\,\nabla_\theta \mathcal{L}_T, \quad \nonumber \\
    &\varphi_m \leftarrow \varphi_m - \eta\,\nabla_{\varphi_m} \mathcal{L}_T, \quad \forall\, m=1,\dots,M,
\end{align}
where $\eta$ is the learning rate. This procedure iterates over all mini-batches for $T$ epochs until convergence.

\subsubsection{Pure Federated Training (FL Mode)}
\label{sec:fl_mode}

In the federated setting, for \(N\) clients, each holds a private local dataset \(\cD_n\) reflecting their respective working environments. The global objective becomes
\begin{equation}\label{eq:fl_global_obj}
    \min_{\theta,\Phi}\; \cL_T(\theta,\Phi) = \sum_{n=1}^{N} w_n\,\cL_T^n(\theta,\Phi_n;\cD_n),
\end{equation}
where \(\cL_T^n\) is the total OmniISR loss (\Cref{eq:total_loss}) evaluated on \(\cD_n\), and \(w_n = |\cD_n|/\sum_k |\cD_k|\) is the aggregation weight for client $n$.

In standard FL with output-layer-only supervision, the non-IID nature of client data causes a well-documented problem of \emph{intermediate representation drift}~\cite{li2021moon,karimireddy2020scaffold}. Because hidden layers receive supervision only indirectly via back-propagation from the output, clients in diverse environments (e.g., urban vs.\ rural) develop independently drifting latent features. When these models are aggregated on the server, the misaligned intermediate representations produce a global model whose hidden features represent an incoherent ``average'' that serves no client well.

OmniISR-embodied intermediate MI supervision directly addresses this problem by providing \emph{explicit, task-relevant anchor points} at each \(G_m\). Because all clients optimize toward the same MI objective at each intermediate depth, their hidden representations are guided toward a shared semantic subspace. Simultaneously, the NE regularizer prevents any single client from developing overconfident, locally specialized features that would resist aggregation. The combination ensures that the aggregated global model inherits coherent, generalizable features across the network hierarchy. \Cref{Fig.overcome_FL_client_drift} illustrates why the proposed OmniISR framework can reduce client drift in federated setting.

\begin{algorithm}[tp]
\caption{Unified OmniISR (CL / FL / Hybrid)}
\label{alg:UnifiedOmniISR}
\SetKwInOut{KwIn}{Input}
\SetKwInOut{KwOut}{Output}
\SetKwFor{ForPar}{for}{do in parallel}{end}

\KwIn{$\mathtt{mode}\in\{\CL{\mathrm{CL}},\FL{\mathrm{FL}},\HY{\mathrm{Hybrid}}\}$;
\CL{centralized dataset $\mathcal{D}_{\mathrm{C}}$}; \FL{clients $\{1,\dots,N\}$ with datasets $\{\mathcal{D}_n\}_{n=1}^N$}; intermediate points $\{G_m\}_{m=1}^M$; adapters $\Phi=\{\varphi_m\}_{m=1}^M$; learning rates $\{\eta_t\}$; \FL{local epochs $E$}; total rounds $T$;
\HY{mixing schedule $\{\alpha_t\}$, adaptation rate $\beta>0$}}
\KwOut{Trained global model $\theta^*$;}

\textbf{Initialization:} Initialize $\theta^0$, $\Phi^0$, $\{\alpha_m,\lambda_m\}_{m=1}^M$\;

\For{round $t = 0$ \KwTo $T-1$}{

  \tcp{\CL{Used by CL and Hybrid}}
  \If{$\mathtt{mode}\in\{\CL{\mathrm{CL}},\HY{\mathrm{Hybrid}}\}$}{
    Sample minibatch $\mathcal{B}_{\mathrm{c}} \subseteq \mathcal{D}_{\mathrm{C}}$\;
    Forward: $\hat{y}_i,\{z_i^1,\dots,z_i^M\}\leftarrow f_{\theta^t}(x_i)$ for $(x_i,y_i)\in\mathcal{B}_{\mathrm{c}}$\;
    Compute $\mathcal{L}_{\mathrm{CE}}$ (\Cref{eq:L_CE}); for $m=1,\dots,M$ compute $\mathcal{L}_{\mathrm{MI}}^{(m)}$ (\Cref{eq:L_MI}), $\mathcal{L}_{\mathrm{NE}}^{(m)}$ (\Cref{eq:L_NE});
    aggregate $\mathcal{L}_T$ (\Cref{eq:total_loss})\;
    Backward: $g_{\mathrm{CL}}^{t} \leftarrow \nabla_\theta \mathcal{L}_T(\theta^t,\Phi^t;\mathcal{B}_{\mathrm{c}})$;\quad
    $g_{\varphi_m,\mathrm{CL}}^{t} \leftarrow \nabla_{\varphi_m}\mathcal{L}_T(\theta^t,\Phi^t;\mathcal{B}_{\mathrm{c}})$\;
    \If{$\mathtt{mode}=\CL{\mathrm{CL}}$}{
      \CL{$\theta^{t+1} \leftarrow \theta^t - \eta_t\, g_{\mathrm{CL}}^{t}$;\quad
          $\varphi_m^{t+1} \leftarrow \varphi_m^{t} - \eta_t\, g_{\varphi_m,\mathrm{CL}}^{t}$ for $\forall m$}\;
      \textbf{continue}\;
    }
  }

  \tcp{\FL{Used by FL and Hybrid}}
  Sample participating clients $S^t \subseteq \{1,\dots,N\}$;\quad broadcast $(\theta^t,\Phi^t)$ to $n\in S^t$\;
  \ForPar{each client $n \in S^t$}{
    $\theta_n \leftarrow \theta^t$;\quad $\varphi_{m,n} \leftarrow \varphi_m^{t}$ for $\forall m$\;
    \For{local epoch $e=1$ \KwTo $E$}{
      \For{each minibatch $\mathcal{B}\subseteq \mathcal{D}_n$}{
        Forward: $\hat{y}_i,\{z_i^1,\dots,z_i^M\}\leftarrow f_{\theta_n}(x_i)$ for $(x_i,y_i)\in\mathcal{B}$\;
        Compute $\mathcal{L}_{\mathrm{CE}}^n$ (\Cref{eq:L_CE}); $\mathcal{L}_{\mathrm{MI}}^{(m),n}$, $\mathcal{L}_{\mathrm{NE}}^{(m),n}$ for $\forall m$; aggregate $\mathcal{L}_T^n$ (\Cref{eq:total_loss})\;
        $\theta_n \leftarrow \theta_n - \eta_t\,\nabla_{\theta_n}\mathcal{L}_T^n$;\quad
        $\varphi_{m,n} \leftarrow \varphi_{m,n} - \eta_t\,\nabla_{\varphi_{m,n}}\mathcal{L}_T^n$ for $\forall m$\;
      }
    }
    Send $(\theta_n, |\mathcal{D}_n|, \{\varphi_{m,n}\}_{m=1}^M)$ to server\;
  }
  $w_n \leftarrow |\mathcal{D}_n|\big/\!\!\sum_{k\in S^t}|\mathcal{D}_k|$;\quad
  $\tilde{\theta}_{t+1} \leftarrow \sum_{n\in S^t} w_n\,\theta_n$;\quad
  $\tilde{\varphi}_m^{t+1} \leftarrow \sum_{n\in S^t} w_n\,\varphi_{m,n}$ \emph{(optional)}\;

  \If{$\mathtt{mode}=\FL{\mathrm{FL}}$}{
    \FL{$\theta^{t+1}\leftarrow \tilde{\theta}_{t+1}$;\quad $\Phi^{t+1}\leftarrow \tilde{\Phi}_{t+1}$}\;
    \textbf{continue}\;
  }

  \tcp{\HY{Used by Hybrid Only}}
  \HY{$g_{\mathrm{FL}}^{t} \leftarrow (\theta^t - \tilde{\theta}_{t+1})/\eta_t$
      $g_{\Phi,\mathrm{FL}}^{t} \leftarrow (\Phi^t - \tilde{\Phi}_{t+1})/\eta_t$ \emph{(optional)} 
      }\;
  \HY{$\theta^{t+1} \leftarrow \theta^t - \eta_t\big(\alpha_t\,g_{\mathrm{CL}}^{t} + (1-\alpha_t)\,g_{\mathrm{FL}}^{t}\big)$}\;
  \HY{$\varphi_m^{t+1} \leftarrow \varphi_m^t - \eta_t\big(\alpha_t\,g_{\varphi_m,\mathrm{CL}}^{t} + (1-\alpha_t)\,g_{\varphi_m,\mathrm{FL}}^{t}\big)$ for $\forall m$ \emph{(optional)}}\;

  \If{\HY{adaptive mixing enabled}}{
    \HY{$s \leftarrow sim\!\big(g_{\mathrm{CL}}^{t},\,g_{\mathrm{FL}}^{t}\big)$;\quad
        $\alpha_{t+1} \leftarrow \mathrm{clip}\!\big(\alpha_t + \beta(1-s),\,0,\,1\big)$}\;
  }
}
\Return $\theta^* \leftarrow \theta^{T}$\;
\end{algorithm}

\subsubsection{Hybrid CL--FL Training (Hybrid Mode)}
\label{sec:hybrid_mode}

In real-world edge intelligence deployments, it is common for a system to have access to both a cloud-curated dataset (e.g., a representative replay buffer collected from diverse regions) and distributed on-device data that must remain local. OmniISR naturally supports a hybrid update that combines centralized and federated gradient information:
\begin{equation}\label{eq:hybrid_update}
    \theta_{t+1} = \theta_t - \eta_t\Big(\alpha_t\,g_{\mathrm{CL}}^t + (1-\alpha_t)\,g_{\mathrm{FL}}^t\Big),
\end{equation}
where \(g_{\mathrm{CL}}^t = \nabla_\theta \cL_T(\theta_t; \cD_{\mathrm{cloud}})\) is the exact gradient on cloud data, \(g_{\mathrm{FL}}^t = (\theta_t - \theta_{t+1}^{\mathrm{fed}})/\eta_t\) is the pseudo-gradient derived from the federated aggregation step, and \(\alpha_t \in [0,1]\) is the mixing weight.
The mixing weight \(\alpha_t\) controls the balance between centralized and federated contributions. 

We identify three practical mixing regimes:
\begin{enumerate}[leftmargin=*,itemsep=2pt]
    \item \textbf{Alternating schedule} (\(\alpha_t \in \{0,1\}\)): The system alternates between pure CL rounds and pure FL rounds. This is simplest to implement and incurs no synchronization overhead between the two data sources.
    \item \textbf{Fixed mixing} (\(\alpha_t = \alpha\)): Both gradient sources contribute in every round with a constant ratio. This provides stable optimization dynamics and is recommended when both data sources are always available.
    \item \textbf{Adaptive mixing}: \(\alpha_t\) is adjusted based on the similarity between \(g_{\mathrm{CL}}^t\) and \(g_{\mathrm{FL}}^t\). When the two directions are well-aligned (high similarity), a larger federated weight captures more data diversity; when they diverge, a larger centralized weight provides corrective guidance.
\end{enumerate}

Beyond the intuitive benefit of combining more data, the hybrid scheme has a deeper theoretical advantage. That is, the mixture of two independent noise sources (centralized sampling noise and federated aggregation noise) increases the effective stochasticity of the optimization trajectory, which accelerates escape from saddle points (see \Cref{thm:escape}) and improves exploration of the loss landscape. 

The training procedure of the proposed OmniISR framework is given in \Cref{alg:UnifiedOmniISR}, including pure CL mode (blue), pure FL mode (green), and hybrid CL--FL mode (red).

\subsection{Theoretical Guarantees}
\label{sec:convergence}

We now provide comprehensive theoretical guarantees of OmniISR under all three training modes. The strict theoretical proofs, detailed explanations, and valuable remarks are deferred to the appendixes in the supplementary material, and here we just discuss the assumptions, state the results, and analyze the most significant implications.

\subsubsection{Common Assumptions}
\label{sec:assumptions}

To conduct the theoretical analysis, we adopt the below standard assumptions for non-convex stochastic optimization.

\begin{assumption}[$L$-Smoothness]\label{ass:smooth}
For each component $\mathcal{L}_s$ with $s \in \{\mathrm{CE}, \{_\mathrm{MI}^m\}_{m=1}^M, \{_\mathrm{NE}^m\}_{m=1}^M\}$, there exists a constant $L_s > 0$ such that for all $\theta, \theta'$,
\[
\|\nabla\mathcal{L}_s(\theta) - \nabla\mathcal{L}_s(\theta')\| \leq L_s\|\theta - \theta'\|.
\]
Consequently, the total objective $\mathcal{L}_T(\cdot)$ is $L_{\max}$-smooth, where 
$L_{\max} = \max\{L_{\mathrm{CE}},\, \{\alpha_m L_{\mathrm{MI}}^{(m)}\}_{m=1}^M,\, \{\lambda_m L_{\mathrm{NE}}^{(m)}\}_{m=1}^M\}$.
\end{assumption}

\begin{assumption}[Bounded Stochastic Gradients]\label{ass:bounded_grad}
For each component $\mathcal{L}_s$ with $s \in \{\mathrm{CE}, \{_\mathrm{MI}^m\}_{m=1}^M, \{_\mathrm{NE}^m\}_{m=1}^M\}$, there exists a constant $G_s > 0$ such that for all $\theta$ and any minibatch,
\[
\mathbb{E}\|\nabla\mathcal{L}_s(\theta)\|^2 \leq G_s^2.
\]
The aggregate bound of $\mathcal{L}_T(\cdot)$ is therefore 
$G_T^2 = G_{\mathrm{CE}}^2 + \sum_{m=1}^{M}\big(\alpha_m^2 (G_{\mathrm{MI}}^{(m)})^2 + \lambda_m^2 (G_{\mathrm{NE}}^{(m)})^2\big)$.
\end{assumption}

\begin{assumption}[Unbiasedness and Bounded Variance]\label{ass:variance}
For each component $\mathcal{L}_s$ with $s \in \{\mathrm{CE}, \{_\mathrm{MI}^m\}_{m=1}^M, \{_\mathrm{NE}^m\}_{m=1}^M\}$, the stochastic gradient $g_s$ is unbiased and has bounded variance: there exists $\sigma_s^2 > 0$ such that for all $\theta$,
\[
\mathbb{E}[g_s \mid \theta] = \nabla\mathcal{L}_s(\theta),
\qquad
\mathbb{E}\|g_s - \nabla\mathcal{L}_s(\theta)\|^2 \leq \sigma_s^2.
\]
Consequently, the stochastic gradient $g_t$ of $\mathcal{L}_T(\cdot)$ satisfies $\mathbb{E}[g_t \mid \theta_t] = \nabla\mathcal{L}_T(\theta_t)$ and $\mathbb{E}\|g_t - \nabla\mathcal{L}_T(\theta_t)\|^2 \leq \sigma_T^2$, where the aggregate variance is
$\sigma_T^2 = \sigma_{\mathrm{CE}}^2 + \sum_{m=1}^{M}\big(\alpha_m^2 (\sigma_{\mathrm{MI}}^{(m)})^2 + \lambda_m^2 (\sigma_{\mathrm{NE}}^{(m)})^2\big)$.
\end{assumption}

\subsubsection{Convergence of Centralized OmniISR}

\begin{theorem}[Centralized OmniISR Convergence]\label{thm:central}
Under Assumptions~\ref{ass:smooth}--\ref{ass:variance}, let the optimization iterates $T$ rounds with being \(\theta_{t+1} = \theta_t - \eta_t g_t\) and step-size \(\eta_t = \eta/\sqrt{T}\). Define \(\Delta = \cL_T(\theta_0) - \cL_T^*\), where $\theta_0$ is the initial model parameters, $\mathcal{L}_{\text{T}}^*$ is the theoretical optimal loss. Then we have
\begin{align}
\label{eq:cl_bound}
    \frac{1}{T}\sum_{t=1}^{T}\E\|\nabla\cL_T(\theta_t)\|^2 &\le \underbrace{\frac{2\Delta}{\eta\sqrt{T}}}_{\text{initial gap}} + \underbrace{\frac{L_{\max}\,\eta}{\sqrt{T}}\,(G_T^2+\sigma_T^2)}_{\text{variance term}} \nonumber \\ 
    &= \cO\!\left(\frac{1}{\sqrt{T}}\right).
\end{align}
\end{theorem}

The bound in \Cref{eq:cl_bound} consists of two terms. The \emph{initial gap} term \(2\Delta/(\eta\sqrt{T})\) reflects the distance between the initial model parameters and the optimum one, and it decreases with larger step-size or more iterations. The \emph{variance term} reflects the noise introduced by mini-batch sampling and scales with the aggregate variance \(G_T^2+\sigma_T^2\). Crucially, the asymptotic rate $\mathcal{O}(1/\sqrt{T})$ matches standard non-convex SGD~\cite{refinetti2023sgd}, proving that the introduction of ISR does \emph{not} degrade the convergence bound, and it only affects the finite-\(T\) constants via \(\sigma_T^2\), \(G_T^2\), and \(L_{\max}\).

\begin{remark}[Role of $M$ in the constants]
The aggregation constants \(G_T^2\) and \(\sigma_T^2\) scale linearly with \(M\) through the weighted sum of component-wise bounds. This means that while the asymptotic convergence rate \(\cO(1/\sqrt{T})\) is independent of \(M\), the \emph{finite-$T$ constants} grow with the number of intermediate points. In practice, this scaling can be controlled by choosing moderate \(M\) (e.g., \(M \le 5\)) and appropriately setting the weights \(\{\alpha_m, \lambda_m\}_{m=1}^M\). 
\end{remark}

\subsubsection{Convergence of Federated OmniISR}

In the federated setting of OmniISR, it runs for $T$ rounds until the global model converges. Specifically, in round $t$, the central server first broadcasts $\theta_t$ to all clients. Then, each client runs $E$ local SGD steps starting from $\theta_t$ with step-size $\eta_t$, producing $\theta_{t,E}^n$. Finally, the server aggregates models received from all participated clients, i.e.,
\begin{align}
\theta_{t+1}&=\sum_{n=1}^N w_n\theta_{t,E}^n = \theta_t + \sum_{n=1}^N w_n\delta_t^n, \\  
\delta_t^n&=\theta_{t,E}^n-\theta_t.
\end{align}
We also define the heterogeneity measure as
\begin{align}
H_t=\sum_{n=1}^N w_n\|\nabla \mathcal{L}_T^n(\theta_t)-\nabla \mathcal{L}_T(\theta_t)\|^2.
\end{align}

Building upon the aforementioned assumptions and definitions, Lemma \ref{lem:drift} establishes a \textit{drift bound} for each federated round. This bound guarantees that the aggregated model does not deviate significantly from its initial state during any given round.
\begin{lemma}[Drift bound]
There exists an absolute constant $C$ such that for every round $t$
\begin{equation}\label{eq:drift-bound}
\sum_{n=1}^N w_n \E\|\delta_t^n\|^2 \le C\,\eta_t^2 E^2 \bigl(\E\|\nabla \mathcal{L}_T(\theta_t)\|^2 + H_t + \sigma_T^2\bigr).
\end{equation}
\label{lem:drift}
\end{lemma}

By constructing the \textit{drift bound}, we now offer the convergence upper bound of the federated OmniISR in \Cref{thm:federated}. 

\begin{theorem}[Federated OmniISR Convergence]\label{thm:federated}
Under Assumptions~\ref{ass:smooth}--\ref{ass:variance} applied per-client, with \(E\) local SGD steps, step-size \(\eta_t = \eta/\sqrt{T}\), and the drift-control condition \(L_{\max}\eta\sqrt{T}\,C\,E^2 < 1\) (where constant $C>0$). Then after $T$ federated aggregation rounds, we have
\begin{align}\label{eq:fl_bound}
    \frac{1}{T}\sum_{t=1}^{T}\E\|\nabla\cL(\theta_t)\|^2 &\le \underbrace{\frac{2\Delta}{\eta\sqrt{T}}}_{\text{initial gap}} + \underbrace{\frac{L_{\max}\,\eta}{\sqrt{T}}\big(G_T^2 + \sigma_T^2\big)}_{\text{variance}} + \nonumber \\ &\underbrace{\frac{L_{\max}\,\eta}{\sqrt{T}}\,\Gamma_{\mathrm{drift}}}_{\text{client drift}},
\end{align}
where the drift contribution satisfies \(\Gamma_{\mathrm{drift}} \le c\,E^2(A + \bar{H})\) with \(A = \frac{1}{T}\sum_t \E\|\nabla\cL(\theta_t)\|^2\) and \(\bar{H} = \frac{1}{T}\sum_t H_t\).
\end{theorem}

The federated bound (\Cref{eq:fl_bound}) differs from the centralized bound (\Cref{eq:cl_bound}) by the additional drift term \(\Gamma_{\mathrm{drift}}\). This term reveals a three-fold interaction:
\begin{enumerate}[leftmargin=*,itemsep=2pt]
    \item \textbf{Local epochs \(E\):} The drift scales as \(E^2\), quantifying the cost of communication. Doubling \(E\) quadruples the drift, motivating the choice of moderate \(E\) in practice.
    \item \textbf{Data heterogeneity \(\bar{H}\):} The drift is amplified by inter-client gradient dissimilarity. In highly non-IID scenarios, \(\bar{H}\) is large, making convergence slower. OmniISR's intermediate MI supervision helps here by aligning client representations (reducing the effective \(\bar{H}\) at hidden layers), which is empirically confirmed by the faster convergence observed in our experiments.
    \item \textbf{Controllability:} Under the drift-control condition \(L_{\max}\eta\sqrt{T}\,C\,E^2 < 1\), the drift term can be absorbed into the left-hand side, preserving the \(\cO(1/\sqrt{T})\) rate. This condition is satisfied for sufficiently small step-size or moderate \(E\).
\end{enumerate}

\subsubsection{Convergence of Hybrid CL--FL OmniISR}

For the hybrid update (\Cref{eq:hybrid_update}), we additionally model the bias structure of each gradient source.

\begin{assumption}[Bias Decomposition]\label{ass:bias}
There exist bias vectors \(b_c^t, b_f^t\) such that
\(\E[g_{\mathrm{CL}}^t|\theta_t] = \nabla\cL_T(\theta_t) + b_c^t\) and
\(\E[g_{\mathrm{FL}}^t|\theta_t] = \nabla\cL_T(\theta_t) + b_f^t\),
with \(\|b_c^t\| \le B_c\) and \(\|b_f^t\| \le B_f\) for all \(t\). The conditional variances satisfy
\(\E\|g_{\mathrm{CL}}^t - \E[g_{\mathrm{CL}}^t|\theta_t]\|^2 \le \sigma_c^2\) and
\(\E\|g_{\mathrm{FL}}^t - \E[g_{\mathrm{FL}}^t|\theta_t]\|^2 \le \sigma_f^2\).
Furthermore, the centralized and federated noise terms are conditionally independent given~\(\theta_t\), i.e.,
\[
  \mathbb{E}\bigl[\langle g_{\mathrm{CL}}^t - 
  \mathbb{E}_t[g_{\mathrm{CL}}^t],\; g_{\mathrm{FL}}^t - 
                    \mathbb{E}_t[g_{\mathrm{FL}}^t] 
                    \rangle \;\big|\; \theta_t \bigr] = 0,
\]
which is natural since the cloud data and the distributed on-device data are collected from independent sources by using independent sensory devices.
\end{assumption}

The bias \(b_c^t\) captures representativeness gaps between the cloud dataset and the true data distribution; \(b_f^t\) captures client drift, staleness, and non-IID effects. We define the effective quantities as follows
\begin{align}
    B_{\mathrm{eff}} &= \max_t \|\alpha_t b_c^t + (1-\alpha_t) b_f^t\|, \label{eq:B_eff}\\
    \sigma_{\mathrm{eff}}^2 &= \alpha_{\min}^2\,\sigma_c^2 + (1-\alpha_{\min})^2\,\sigma_f^2, \label{eq:sigma_eff}
\end{align}
where \(\alpha_{\min} = \min\limits_t \alpha_t > 0\) ensures the persistent participation of centralized training.

\begin{theorem}[Hybrid OmniISR Convergence]\label{thm:hybrid}
Under Assumptions~\ref{ass:smooth}--\ref{ass:bias}, with step-size \(\eta_t = \eta/\sqrt{T}\), CL--FL mixing weights \(\alpha_t \ge \alpha_{\min} > 0\), and \(C > 0\), we have
\begin{equation}\label{eq:hybrid_bound}
    \frac{1}{T}\sum_{t=1}^{T}\E\|\nabla\cL_T(\theta_t)\|^2 \le \underbrace{\frac{C \Delta}{\alpha_{\min}\,\eta\sqrt{T}}}_{\text{initial gap}} + \underbrace{C\,\frac{\eta\,\sigma_{\mathrm{eff}}^2}{\sqrt{T}}}_{\text{variance}} + \underbrace{C\,B_{\mathrm{eff}}^2}_{\text{bias floor}}.
\end{equation}
\end{theorem}

The hybrid framework introduces a non-vanishing bias floor \(B_{\mathrm{eff}}\), which persists even as the number of iterations \(T \to \infty\). This irreducible error arises from the inherent imperfections in the individual gradient sources, where the centralized cloud data may not fully capture the true global distribution, and the federated updates are typically subject to client drift. However, this fundamental trade-off is manageable. By actively mitigating these biases, such as through representative cloud sampling and the application of drift-reduction algorithms like SCAFFOLD \cite{karimireddy2020scaffold}, the bias floor can be kept sufficiently small. Under these controlled conditions, the hybrid scheme maintains the standard \(\mathcal{O}(1/\sqrt{T})\) convergence rate.

\begin{figure}[!t]
\includegraphics[width=\linewidth]{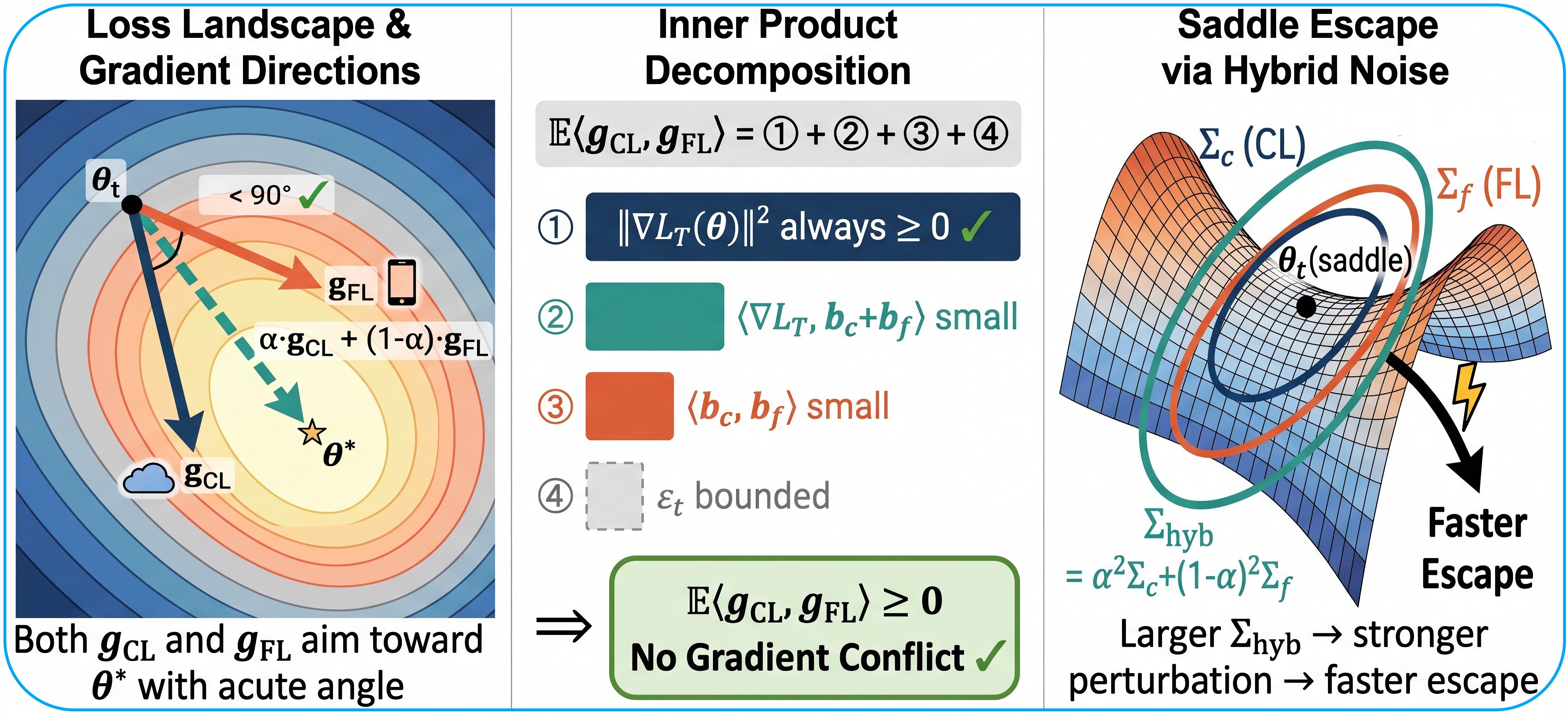}
\vspace{-0.7cm}
\caption{Illustration of how OmniISR helps to escape saddle point.}
\label{Fig:escape_saddle}
\vspace{-0.5cm}
\end{figure}

\subsubsection{Gradient Non-Conflict Between CL and FL}

A critical concern in hybrid CL-FL OmniISR is whether the centralized and federated gradient directions might \emph{conflict}, i.e., pushing the model parameters in opposing directions, leading to oscillated or degraded convergence. The following lemma provides a formal guarantee.

\begin{lemma}[No Gradient Conflict in Hybrid OmniISR]\label{lem:noconflict}
Under Assumption~\ref{ass:bias}, the expected inner product between centralized and federated gradients decomposes as
\begin{align}\label{eq:inner_product}
    \E[\langle g_{\mathrm{CL}}^t, g_{\mathrm{FL}}^t\rangle] = &\underbrace{\|\nabla\cL_T(\theta_t)\|^2}_{\text{positive definite}} + \underbrace{\langle\nabla\cL_T(\theta_t),\, b_c^t + b_f^t\rangle}_{\text{bias--gradient cross term}} + \nonumber \\ &\underbrace{\langle b_c^t, b_f^t\rangle}_{\text{bias--bias interaction}} + \underbrace{\epsilon_t}_{\text{noise residual}},
\end{align}
where \(|\epsilon_t| \le \frac{1}{2}(\sigma_c^2 + \sigma_f^2)\). In particular, when the biases are small relative to the gradient norm (i.e., \(\|b_c^t\| + \|b_f^t\| \le c\,\|\nabla\cL_T(\theta_t)\|\) for some \(c < 1\)), the dominant \(\|\nabla\cL_T(\theta_t)\|^2\) term ensures
\begin{equation}
    \E[\langle g_{\mathrm{CL}}^t, g_{\mathrm{FL}}^t\rangle] \ge 0.
\end{equation}
\end{lemma}

This result provides the first formal justification for mixed CL--FL training in edge intelligence. It demonstrates that, under mild and verifiable conditions, the two gradient sources act synergistically to guide the optimization trajectory toward stationary points, rather than producing conflicting updates. The required condition (i.e., \(\|b_c^t\| + \|b_f^t\| \ll \|\nabla\cL_T(\theta_t)\|\)) naturally holds throughout the majority of the training process when the gradients are large. It is typically only violated near convergence where the gradients diminish and the model approaches a stationary point. This near-convergence regime that the saddle-escape mechanism becomes critical and is discussed below in \Cref{sec:escape_saddle}.

\subsubsection{Accelerated Saddle Escape via Hybrid Stochasticity} \label{sec:escape_saddle}

Near strict saddle points, gradient norms are small and the gradient non-conflict condition may weaken. However, stochasticity becomes beneficial. Specifically, noisy gradients help the optimizer escape the saddle, and more importantly the hybrid OmniISR scheme amplifies this effect.

\begin{assumption}[Hessian Lipschitz]\label{ass:hessian}
There exists a constant $\rho > 0$ such that the Hessian is $\rho$-Lipschitz: for all $\theta, \theta'$,
\[
\|\nabla^2\mathcal{L}_T(\theta) - \nabla^2\mathcal{L}_T(\theta')\| \leq \rho\|\theta - \theta'\|.
\]
\end{assumption}

\begin{theorem}[Saddle Escape]\label{thm:escape}
Suppose Assumptions~\ref{ass:smooth}, \ref{ass:bias}, and \ref{ass:hessian} hold, and let the model \(\theta_t\) be located near a strict saddle point exhibiting negative curvature \(-\gamma\) (where \(\gamma > 0\)). With a probability of at least \(1-\delta\), the hybrid optimization scheme will successfully escape a neighborhood of radius \(R\) around this saddle point in a number of iterations bounded by
\begin{equation}\label{eq:escape_bound}
    T_{\mathrm{esc}}(\delta) \!\le\! \frac{1}{\eta \gamma} \!\ln\!\left(\frac{C_0 R}{y_0 \!-\! \eta C_B \!-\! \eta C_\sigma(\delta)}\right) \!=\! \mathcal{O}\!\left(\!\frac{1}{\eta\gamma}\!\log\!\frac{R}{\delta}\right),
\end{equation}
where \(C_B = B_{\mathrm{eff}}/\gamma\), \(C_\sigma(\delta) = (\sigma_{\mathrm{eff}}/\sqrt{\eta\gamma})\sqrt{\ln(2/\delta)}\), and \(y_0\) represents the initial displacement along the direction of negative curvature. The constant \(C_0\) is determined by \(L_{\max}\) and \(\rho\).
\end{theorem}

The time required to escape the saddle point is governed by the effective second moment \(S^2 = B_{\mathrm{eff}}^2 + \sigma_{\mathrm{eff}}^2\). Notably, the hybrid framework yields a larger \(S^2\) compared to a purely individual approach. This is because the compound gradient introduces additional variance from an independent data distribution. Consequently, this induces a more substantial effective perturbation along the direction of negative curvature, thereby accelerating the escape process. \Cref{Fig:escape_saddle} demonstrates this process. 

Furthermore, our bound on the escape time directly aligns with the perturbed gradient descent (PGD) framework \cite{jin2017escape}. This connection is made by equating their perturbation radius to our effective initial displacement \(y_0\), their noise scale to \(S^2\), and their negative-curvature magnitude to \(\gamma\). By setting the learning rate and incorporating the relevant dimension-dependent geometric factors, we can recover \cite{jin2017escape}'s iteration complexity directly from \Cref{eq:escape_bound}.

\subsection{Complexity Analysis for $\epsilon$-Stationarity} \label{sec:complexity_analysis}

To find the expected number of iteration times $T(\epsilon)$ required for $\epsilon$-stationarity, we convert \Cref{thm:central,thm:federated,thm:hybrid} via
\begin{align}
1/T\sum\nolimits_{t=1}^{T}\mathbb{E}\bigl\|\nabla\mathcal{L}_T(\theta_t)\bigr\|^2 \;\le\; \epsilon.
\end{align}

\subsubsection{Pure Centralized Mode (\Cref{thm:central})}

Setting the right-hand side of \Cref{eq:cl_bound} to $\epsilon$ gives the required round $T_{\text{CL}}^{\text{ISR}}(\epsilon)$ to reach $\epsilon$-stationarity as follow
\begin{equation}\label{eq:cl-complexity}
\;T_{\text{CL}}^{\text{ISR}}(\epsilon)\;=\;\mathcal{O}\!\bigl(\frac{\Delta\,L_{\max}\,(G_T^2+\sigma_T^2)}{\epsilon^{2}}\bigr).
\end{equation}
Owing to $G_T^2=\mathcal{O}(M)$ and $\sigma_T^2=\mathcal{O}(M)$, the absolute iteration count grows \emph{at most linearly in $M$}, while the asymptotic rate $\mathcal{O}(1/\sqrt{T})$ is preserved. 

\begin{remark}[Rationality of the $\mathcal{O}(1/\epsilon^2)$ bound]
\Cref{eq:cl-complexity} is information-theoretically optimal, satisfying \cite{arjevani2023lower}-established $\Omega(1/\epsilon^2)$ lower bound for first-order stochastic methods on $L$-smooth nonconvex objectives with bounded variance. 
\end{remark}

\subsubsection{Pure Federated Mode (\Cref{thm:federated})}

Based on \Cref{thm:federated}, the derived iteration number to reach $\epsilon$-stationarity of FL mode of OmniISR is
\begin{equation}\label{eq:fl-complexity}
\;T_{\mathrm{FL}}^{\text{ISR}}(\epsilon) \;=\; \mathcal{O}\!\bigl(\frac{\Delta\,L_{\max}\,\bigl(G_T^2 + \sigma_T^2 + E^2\,\bar{H}\bigr)}{\epsilon^2}\bigr).
\end{equation}

ISR augments each client's loss with MI supervision and NE regularization that push the learned representations of each client toward a shared representation. This representation alignment translates into a \emph{gradient-level alignment} bound, i.e., there exists a constant $\kappa\in(0,1)$ such that $\bar{H}\le \kappa\bar{H}_0$, where $\bar{H}_0$ is the expected gradient heterogeneity in vanilla FL algorithms without any representation alignment.
The contraction factor $\kappa$ depends on the ISR's strength (i.e., $\{\alpha_m, \lambda_m\}_{m=1}^M$). In general, stronger ISR strength produces smaller $\kappa$, with $\kappa\to 1$ when the ISR weights vanish (recovering vanilla FL) and $\kappa\to 0$ when representations are 
perfectly aligned.

For \emph{vanilla FL} without ISR, by substituting $\bar{H}=\bar{H}_0$ into \Cref{eq:fl-complexity}, we can obtain its iteration complexity 
\begin{equation}\label{eq:TFL0-def}
T_{\text{FL}}^{0}(\epsilon) \;\triangleq\; \mathcal{O}\!\bigl(\frac{\Delta\,L_{\max}\,\bigl(G_T^2 + \sigma_T^2 + E^{2}\,\bar{H}_0\bigr)}{\epsilon^{2}}\bigr).
\end{equation}
With \emph{ISR in place}, replacing $\bar{H}$ by $\kappa\,\bar{H}_0$, it yields
\begin{equation}\label{eq:TFL-isr}
T_{\text{FL}}^{\text{ISR}}(\epsilon) \;=\; \mathcal{O}\!\bigl(\frac{\Delta\,L_{\max}\,\bigl(G_T^2 + \sigma_T^2 + E^{2}\,\kappa\,\bar{H}_0\bigr)}{\epsilon^{2}}\bigr).
\end{equation}
Based on \Cref{eq:TFL0-def} and \Cref{eq:TFL-isr}, we can conclude that
\begin{itemize}
    \item In the heterogeneity-dominated regime where $\bar{H}_0 \gg G_T^2+\sigma_T^2$, this simplifies to
\begin{align}
T_{\text{FL}}^{\text{ISR}}(\epsilon) \;\approx\; \kappa\;T_{\text{FL}}^{0}(\epsilon),
\label{eq:kappa-reduced-FL-round}
\end{align}
indicating that ISR yields a reduction factor of $\kappa<1$ in the number of communication rounds required to reach $\epsilon$-stationarity. 
Equation~\Cref{eq:kappa-reduced-FL-round} formalizes the communication-efficiency gains: ISR does not change the asymptotic rate $\mathcal{O}(1/\sqrt{T})$, but it reduces the \emph{prefactor} by a contraction $\kappa$, directly translating into fewer communication rounds for the same target accuracy.

\item In the noise-dominated regime where $G_T^2+\sigma_T^2 \gg \bar{H}_0$, the ISR's gain is smaller, since heterogeneity is already a sub-leading term.
\end{itemize}

\subsubsection{Hybrid CL--FL Mode (\Cref{thm:hybrid})}

In the hybrid CL--FL regime, the \emph{irreducible bias floor} $B_{\text{eff}}^{2}$ stems from the mismatch between the sampled cloud distribution and the real distribution, and data heterogeneity among FL clients. For target accuracy $\epsilon$, it yields
\begin{equation}\label{eq:hybrid-complexity}
\;T_{\text{Hyb}}^{\text{ISR}}(\epsilon)\;=\;\mathcal{O}\!\bigl(\frac{\Delta\sigma_{\text{eff}}^{2}}{\alpha_{\min}^{2}\,(\epsilon-C\,B_{\text{eff}}^{2})^{2}}\bigr),\;
\end{equation}
revealing two regimes:
\begin{itemize}
    \item \emph{Above the floor} ($\epsilon > C\,B_{\text{eff}}^{2}$): convergence is achievable with $\mathcal{O}\bigl((\epsilon-C B_{\text{eff}}^{2})^{-2}\bigr)$, slightly worse than the centralized $\mathcal{O}(\epsilon^{-2})$ due to the shifted denominator. 
    \item \emph{Below the floor} ($\epsilon \le C\,B_{\text{eff}}^{2}$): convergence is \emph{unattainable} regardless of $T$. The algorithm asymptotically oscillates.
\end{itemize}

The hybrid bound interpolates between CL and FL. Specifically, it inherits the lower variance of CL (via $\alpha_{\min}$) and the richer data coverage of FL (via $1-\alpha_{\min}$), at the cost of a bias floor that reflects the mismatch between the two gradient sources. 
When ISR is active, the federated bias $B_f$ and heterogeneity $\bar{H}$ are both reduced, further tightening the hybrid bound.

\subsection{Comparison among Training Modes of OmniISR} \label{sec:modes_comparison}
\Cref{tab:omniisr_comparison} compares aforementioned three training modes of the proposed OmniISR framework from multiple perspectives, including the role of the intermediate loss and regularizer, and theoretical convergence guarantees.

\begin{table*}[!t]
\renewcommand{\arraystretch}{1.25}
\caption{Comprehensive Comparison of the working modes of \textsc{OmniISR}: Pure CL, Pure FL, and Hybrid CL--FL}
\label{tab:omniisr_comparison}
\centering
\scriptsize
\setlength{\tabcolsep}{4pt}
\begin{tabularx}{\linewidth}{L{2.88cm} C{4.1cm} C{5.05cm} C{5.05cm}}
\toprule
\rowcolor{ieeeblue}
\textcolor{white}{\textbf{Mode}} &
\textcolor{white}{\textbf{Centralized (CL) OmniISR}} &
\textcolor{white}{\textbf{Federated (FL) OmniISR}} &
\textcolor{white}{\textbf{Hybrid CL--FL OmniISR}} \\

\midrule
\rowcolor{white} Update rule &
$\theta_{t+1}=\theta_t-\eta_t g_t$ &
$\theta_{t+1}=\theta_t+\sum_n w_n\delta_t^n$ &
$\theta_{t+1}=\theta_t-\eta_t[\alpha_t g_{\mathrm{CL}}^t+(1-\alpha_t)g_{\mathrm{FL}}^t]$ \\

\multirow{2}{*}{\makecell[l]{Explicit bound on \\ $\tfrac{1}{T}\!\sum_t\!\mathbb{E}\|\nabla\mathcal{L}_T(\theta_t)\|^{2}$}} &\multirow{2}{*}{$\dfrac{2\Delta}{\eta\sqrt{T}}+\dfrac{L_{\max}\eta(G_T^{2}+\sigma_T^{2})}{\sqrt{T}}$} &\multirow{2}{*}{$\dfrac{2\Delta}{\eta\sqrt{T}}+\dfrac{L_{\max}\eta(G_T^{2}+\sigma_T^{2}+\Gamma_{\mathrm{drift}})}{\sqrt{T}}$} &\multirow{2}{*}{$\dfrac{C\Delta}{\alpha_{\min}\eta\sqrt{T}}+\dfrac{C\eta\sigma_{\mathrm{eff}}^{2}}{\sqrt{T}}+C B_{\mathrm{eff}}^{2}$} \\
& & & \\

\rowcolor{white} Convergence rate &
$\mathcal{O}(1/\sqrt{T})$ &
$\mathcal{O}(1/\sqrt{T})$ &
$\mathcal{O}(1/\sqrt{T}) + C\,B_{\mathrm{eff}}^{2}$ \\

\rowcolor{white} Step-size condition &
$L_{\max}\eta \le \sqrt{T}$ &
$L_{\max}\eta\sqrt{T}\,cE^{2}<1$ &
$L_{\max}\eta \le \sqrt{T}/4$ \\

\rowcolor{white} Rounds to $\epsilon$-stationarity & $\mathcal{O}\!\bigl({\Delta L_{\max}(G_T^2+\sigma_T^2)}/{\epsilon^{2}}\bigr)$ & $\mathcal{O}\!\bigl({\Delta L_{\max}\bigl(G_T^2 + \sigma_T^2 + E^2\bar{H}\bigr)}/{\epsilon^2}\bigr)$ & $\mathcal{O}\!\bigl({\Delta\sigma_{\text{eff}}^{2}}/({\alpha_{\min}^{2} (\epsilon-C B_{\text{eff}}^{2})^{2}})\bigr)$ \\

\bottomrule
\end{tabularx}
\raggedright
\vspace{-0.3cm}
\end{table*}

%% file: experiments.tex
\section{Experiments}
\label{sec:experiments}

\begin{table*}[tp]
\centering
\setlength{\tabcolsep}{3.6pt}
\caption{Quantitative performance evaluation of OmniISR across multiple model architectures and datasets}
\vspace{-0.2cm}
\begin{tabularx}{\linewidth}{c|c|c|c|cccc|cccc|cccc}
\toprule
\multirow{2}{*}{Models} & \multirow{2}{*}{Remark} & \multirow{2}{*}{Setting} & \multirow{2}{*}{OmniISR?} & \multicolumn{4}{c|}{Cityscapes Dataset (\%)} & \multicolumn{4}{c|}{CamVid Dataset (\%)} & \multicolumn{4}{c}{SynthiaSF Dataset (\%)} \\ \cline{5-8} \cline{9-12} \cline{13-16} 
 & & & & mIoU & mF1 & mPre & mRec & mIoU & mF1 & mPre & mRec & mIoU & mF1 & mPre & mRec \\ \hline
\multirow{4}{*}{DeepLabv3+} & \multirow{4}{*}{\shortstack{ResNet18\\(Backbone)}} & \multirow{2}{*}{Centralized} & \xmark & 47.91 & 56.31 & 59.99 & 55.58  & 76.02 & 82.43 & 83.07 & 82.46 & 33.28 & 37.27 & 38.98 & 36.45 \\ 
 & & & \checkmark & \textbf{50.49} & \textbf{59.70} & \textbf{65.00} & \textbf{57.82} & \textbf{76.13} & \textbf{82.52} & \textbf{83.09} & \textbf{82.57} & \textbf{34.28} & \textbf{39.13} & \textbf{42.74} & \textbf{37.60} \\ \cline{3-16}
 & & \multirow{2}{*}{Federated} & \xmark & 43.76 & 50.40 & 51.54 & 50.77& 72.78 & 80.24 & 81.33 & 79.47 & 26.65 & 31.94 & 40.30 & 30.46 \\ 
 & & & \checkmark & \textbf{47.78} & \textbf{56.28} & \textbf{59.64} & \textbf{55.64} & \textbf{73.24} & \textbf{80.70} & \textbf{81.60} & \textbf{80.34} & \textbf{29.03} & \textbf{34.90} & \textbf{41.12} & \textbf{32.92} \\ \hline
\multirow{4}{*}{SeaFormer} & \multirow{4}{*}{\shortstack{Axial\\Transformer}} & \multirow{2}{*}{Centralized} & \xmark & 27.40 & 30.99 & 30.55 & 32.14 & 50.69 & 56.00 & 55.40 & 56.89 & \textbf{24.74} & \textbf{29.70} & \textbf{32.68} & \textbf{29.19} \\
 & & & \checkmark & \textbf{29.82} & \textbf{34.19} & \textbf{33.80} & \textbf{35.45} & \textbf{55.83} & \textbf{62.39} & \textbf{64.19} & \textbf{62.54} & 24.20 & 29.23 & 32.20 & 29.00 \\ \cline{3-16}
 & & \multirow{2}{*}{Federated} & \xmark & 27.32 & 30.95 & 32.78 & 31.85 & 47.08 & 53.39 & 53.86 & 53.91 & 16.29 & 21.05 & \textbf{28.98} & 21.29 \\
 & & & \checkmark & \textbf{29.77} & \textbf{34.18} & \textbf{33.84} & \textbf{35.39} & \textbf{51.54} & \textbf{59.40} & \textbf{60.07} & \textbf{60.36} & \textbf{18.43} & \textbf{23.20} & 28.33 & \textbf{22.88} \\ \hline
\multirow{4}{*}{TopFormer} & \multirow{4}{*}{\shortstack{Normal\\Transformer}} & \multirow{2}{*}{Centralized} & \xmark & 32.76 & 37.64 & 36.92 & 39.24 & 63.10 & 70.22 & 71.88 & 70.25 & 28.37 & 33.75 & 36.97 & 32.99 \\
 & & & \checkmark & \textbf{34.28} & \textbf{39.96} & \textbf{40.41} & \textbf{40.60} & \textbf{66.38} & \textbf{74.50} & \textbf{77.47} & \textbf{73.60} & \textbf{28.70} & \textbf{34.04} & \textbf{37.22} & \textbf{33.20} \\ \cline{3-16}
 & & \multirow{2}{*}{Federated} & \xmark & 32.29 & 37.23 & \textbf{36.84} & 38.35 & 56.60 & 63.21 & 71.66 & 63.26 & 21.60 & 26.88 & \textbf{32.22} & \textbf{26.71} \\
 & & & \checkmark & \textbf{32.33} & \textbf{37.24} & 36.75 & \textbf{38.43} & \textbf{58.85} & \textbf{66.20} & \textbf{72.02} & \textbf{65.20} & \textbf{21.63} & \textbf{26.98} & 31.80 & 25.81 \\ 
 \bottomrule
\end{tabularx}
\label{tab:OmniISR_quantitative_comp}
\vspace{-0.5cm}
\end{table*}

\subsection{Datasets, Metrics, and Implementation}
\label{sec:datasets_metrics_impl}

\textbf{Datasets}. We take AD semantic segmentation task as example to evaluate the proposed OmniISR framework on three benchmarks used throughout the literature. The Cityscapes dataset \cite{Cordts2016Cityscapes} consists of 2,975 training images and 500 validation images, each annotated with masks. This dataset encompasses 19 semantic classes, such as vehicles and pedestrians. The CamVid dataset \cite{brostow2008segmentation} comprises a total of 701 images across 11 semantic classes. For our experiments, we randomly selected 600 samples for training and used the remaining 101 samples as a test dataset. The SynthiaSF dataset \cite{ros2016synthia} offers a collection of synthetic, yet photorealistic images that emulate urban scenarios. It provides pixel-level annotations for 23 semantic classes, with 1,596 images designated for training and 628 for testing.

\textbf{Metrics}. We assess the proposed OmniISR framework using four commonly used metrics: \textbf{mIoU}: the mean of intersection over union; \textbf{mPrecision (mPre for short)}: the mean ratio of true positive pixels to the total predicted positive pixels; \textbf{mRecall (mRec for short)}: the mean ratio of true positive pixels to the total positive ground truth pixels; \textbf{mF1}: the mean of harmonic mean of precision and recall, providing a balanced measure of these two metrics. Such metrics are evaluated across all semantic classes, offering a comprehensive view of OmniISR's performance. 

\textbf{Implementation}. All models and baselines are implemented using the PyTorch framework and trained on two NVIDIA GeForce 4090 GPUs. For optimization, we employ the Adam optimizer with beta values of 0.9 and 0.999, alongside a weight decay of 1e-4. Our experiments contain lots of comparisons with comprehensive analyses, utilizing models of CNN-based DeepLabv3+ \cite{chen2018deeplabv3p}, Transformer-based TopFormer~\cite{zhang2022topformer}, and CNN-Transformer hybrid-based SeaFormer \cite{wan2023seaformer}, across the Cityscapes, CamVid, and SynthiaSF datasets. To systematically evaluate the versatility of OmniISR, we contrast its CL training mode against the traditional output-layer supervision approach, while its FL mode is benchmarked against a diverse set of established FL algorithms, including FedProx \cite{li2020federated}, FedDyn \cite{acar2021federated}, FedAvgM \cite{hsu2019measuring}, FedIR \cite{hsu2020federated}, MOON \cite{li2021model}, SCAFFOLD \cite{karimireddy2020scaffold}, FedAvg \cite{mcmahan2017communication}, BalanceFL \cite{9825928}, and FedGau \cite{kou2025fedgau}.

\subsection{Evaluation and Empirical Analyses}
\label{sec:cl_results}
\subsubsection{Quantitative Evaluation}
We carry out extensive experiments to compare the quantitative performance of OmniISR against standard output-layer-only supervision baselines under both centralized and federated settings. The evaluated architectures include the models of CNN-based DeepLabv3+, Transformer-based TopFormer, and CNN-Transformer hybrid-based SeaFormer. The results across the Cityscapes, CamVid, and SynthiaSF datasets are presented in \Cref{tab:OmniISR_quantitative_comp}. Rather than isolated case-by-case number reporting, we substantially reframe the empirical results into reviewer-facing research questions. Specifically, we ask: \textbf{RQ1} whether OmniISR yields consistent gains for both CL and FL paradigms, \textbf{RQ2} whether OmniISR-achieved gains track heterogeneity and optimization difficulty, \textbf{RQ3} how model architecture influences effectiveness, and \textbf{RQ4} whether OmniISR truly improves both paradigms' \emph{unification quality} (not only absolute accuracy). This framing is designed to evaluate this paper's core claims, i.e., cross-paradigm applicability, architecture-agnostic deployment, and drift--generalization trade-off.

\begin{table}[t]
\centering
\setlength{\tabcolsep}{6pt}
\caption{Matrix-level mIoU gain summary derived from \Cref{tab:OmniISR_quantitative_comp}}
\vspace{-0.2cm}
\begin{tabular}{ccc}
\toprule
Setting & Mean Absolute Gain & Mean Relative Gain \\
\midrule
Centralized (9 pairs) & +1.76 & +4.04\% \\
Federated (9 pairs) & +2.03 & +6.06\% \\
\bottomrule
\end{tabular}
\label{tab:macro_miou_gain}
\vspace{-0.4cm}
\end{table}

\subsection*{RQ1: Does OmniISR provide coherent cross-paradigm gains?}

Yes. Matrix-level statistics in \Cref{tab:macro_miou_gain} show that OmniISR improves mIoU in both paradigms, with larger average gain in FL (+2.03 absolute, +6.06\%) than in CL (+1.76 absolute, +4.04\%). This asymmetry of improvement is theoretically meaningful: if intermediate constraints primarily mitigate representation drift, the benefit should be amplified under non-IID FL, which is exactly what we observe. At the exemplar level, DeepLabv3+ on Cityscapes increases from 47.91 to 50.49 in CL and from 43.76 to 47.78 in FL, confirming that OmniISR improves model performance for both paradigms and does not trade one paradigm for the other.

\begin{table*}[tp]
\centering
\renewcommand{\arraystretch}{0.24}
\addtolength{\tabcolsep}{-0.52pt}
\caption{Qualitative performance comparison of the centralized setting of proposed OmniISR against conventional output-supervision-only training method}
\vspace{-0.2cm}
\begin{tabularx}{\linewidth}{|l|lllll|}
\hline
\verticaltext[27.5pt]{Raw RGBs} &
\includegraphics[width=0.188\linewidth, height=0.12\linewidth]{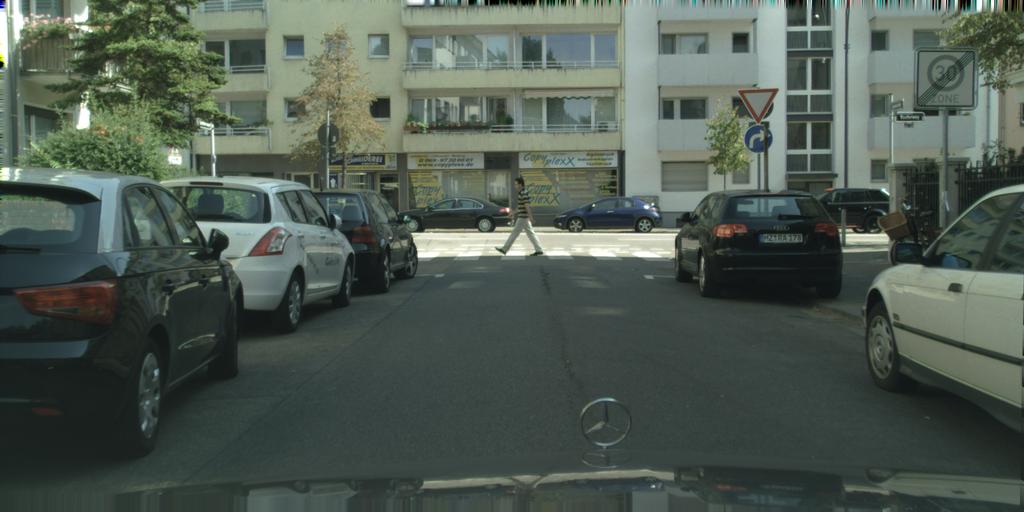} &\hspace{-0.47cm}
\includegraphics[width=0.188\linewidth, height=0.12\linewidth]{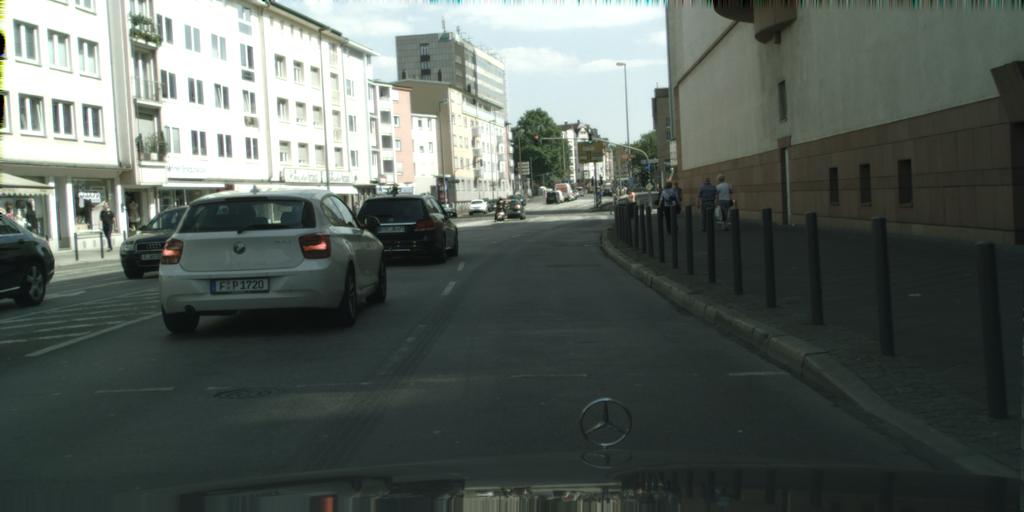} &\hspace{-0.47cm}
\includegraphics[width=0.188\linewidth, height=0.12\linewidth]{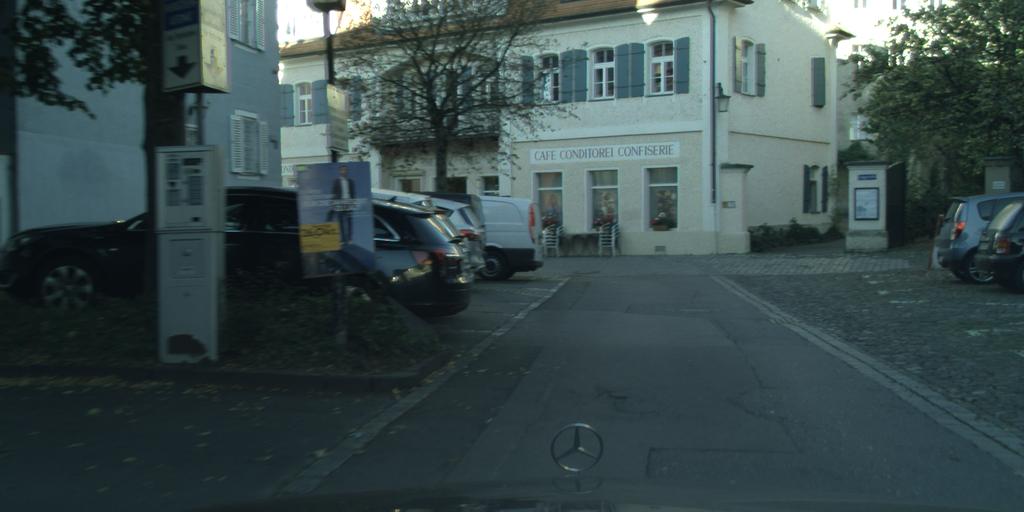   } &\hspace{-0.47cm}
\includegraphics[width=0.188\linewidth, height=0.12\linewidth]{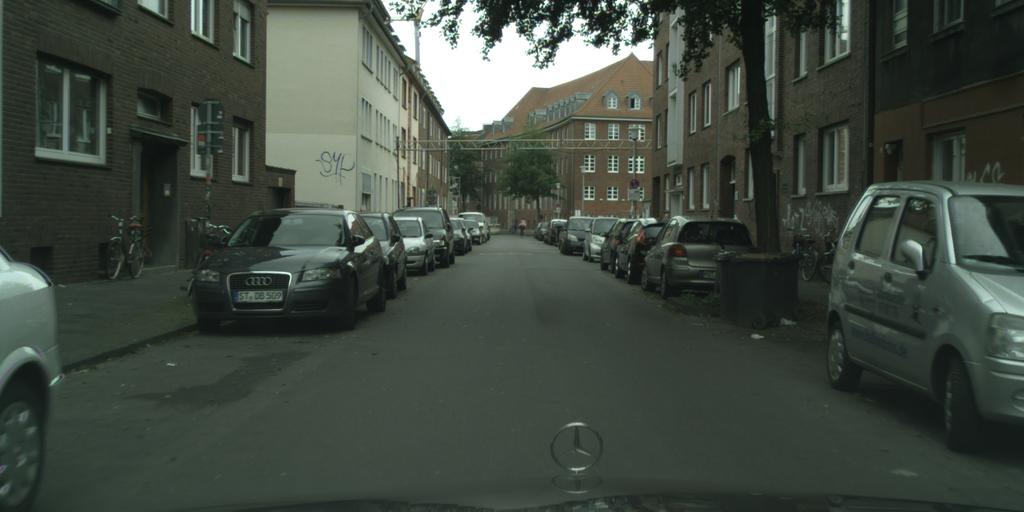  } &\hspace{-0.47cm}
\includegraphics[width=0.188\linewidth, height=0.12\linewidth]{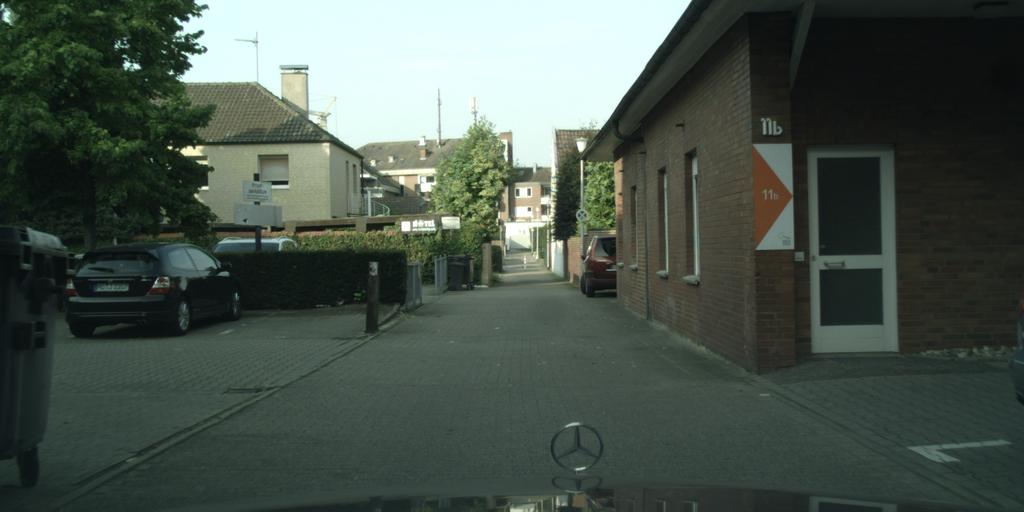  }\\
\hline

\verticaltext[27.5pt]{Ground Truth} &
\includegraphics[width=0.188\linewidth, height=0.12\linewidth]{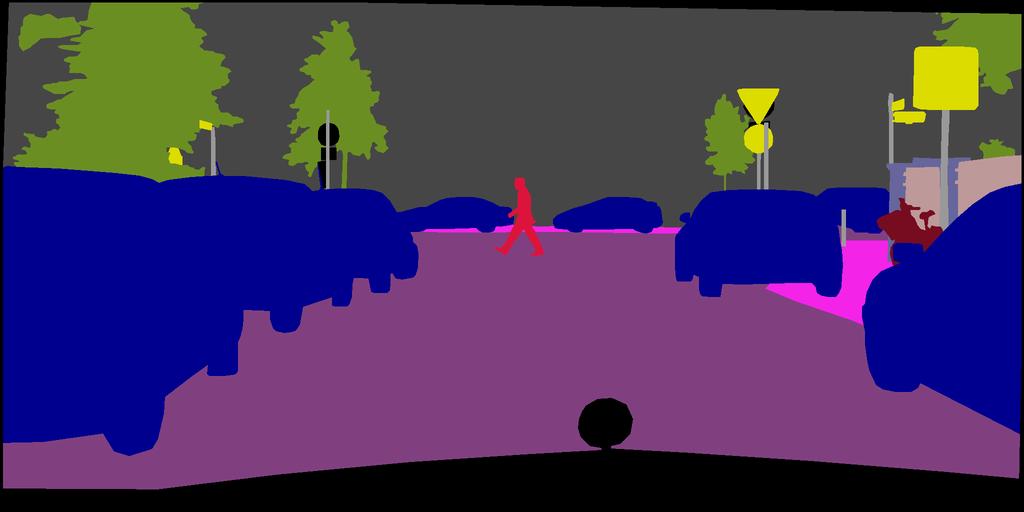} &\hspace{-0.47cm}
\includegraphics[width=0.188\linewidth, height=0.12\linewidth]{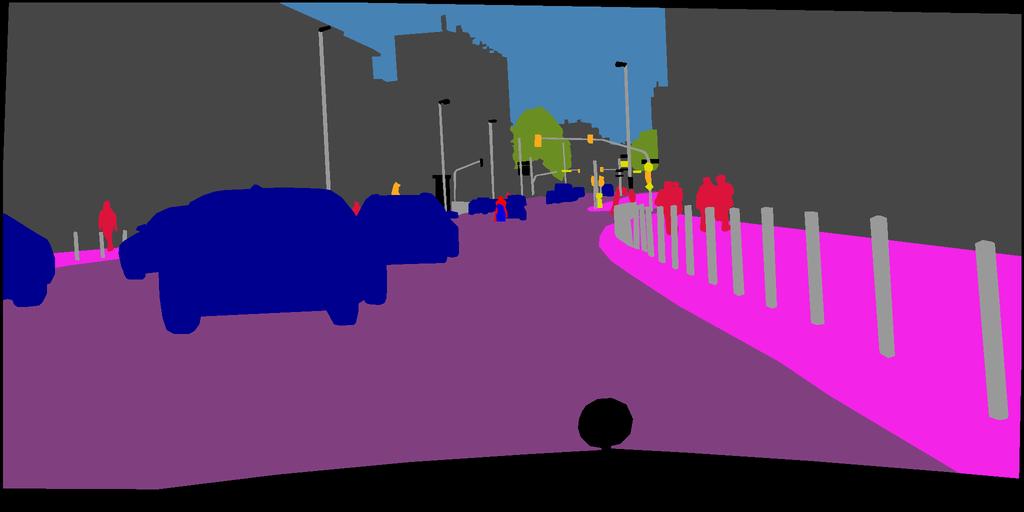} &\hspace{-0.47cm}
\includegraphics[width=0.188\linewidth, height=0.12\linewidth]{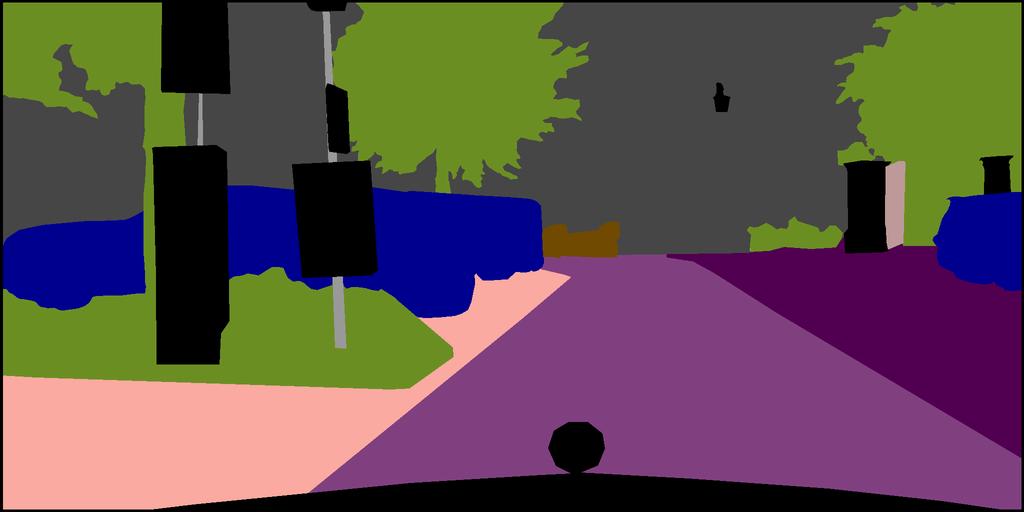   } &\hspace{-0.47cm}
\includegraphics[width=0.188\linewidth, height=0.12\linewidth]{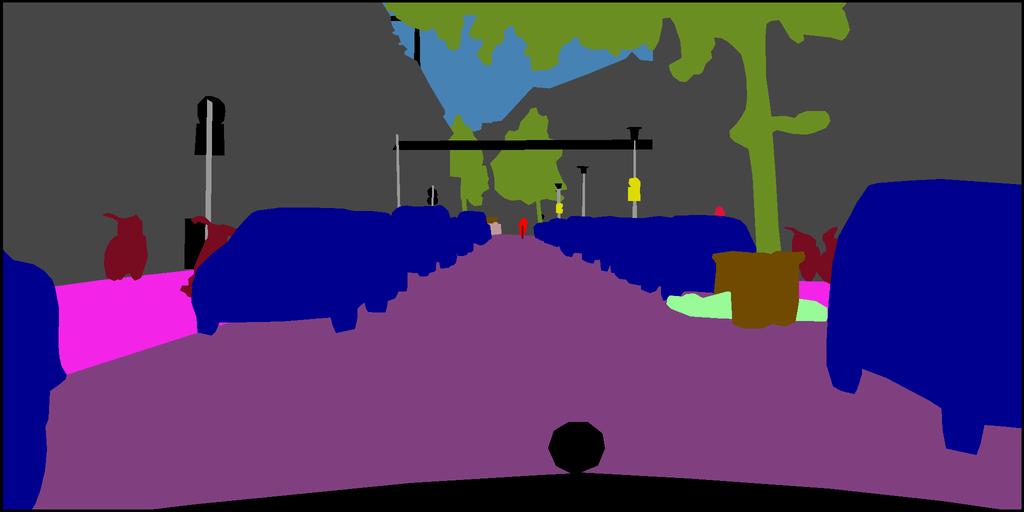  } &\hspace{-0.47cm}
\includegraphics[width=0.188\linewidth, height=0.12\linewidth]{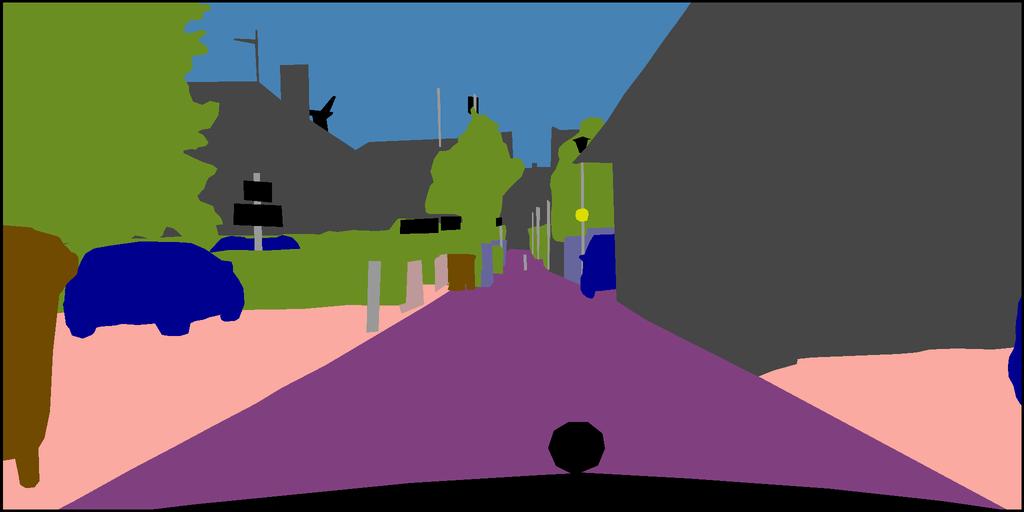  }\\
\hline

\verticaltext[27.5pt]{w/o OmniISR} &
\includegraphics[width=0.188\linewidth, height=0.12\linewidth]{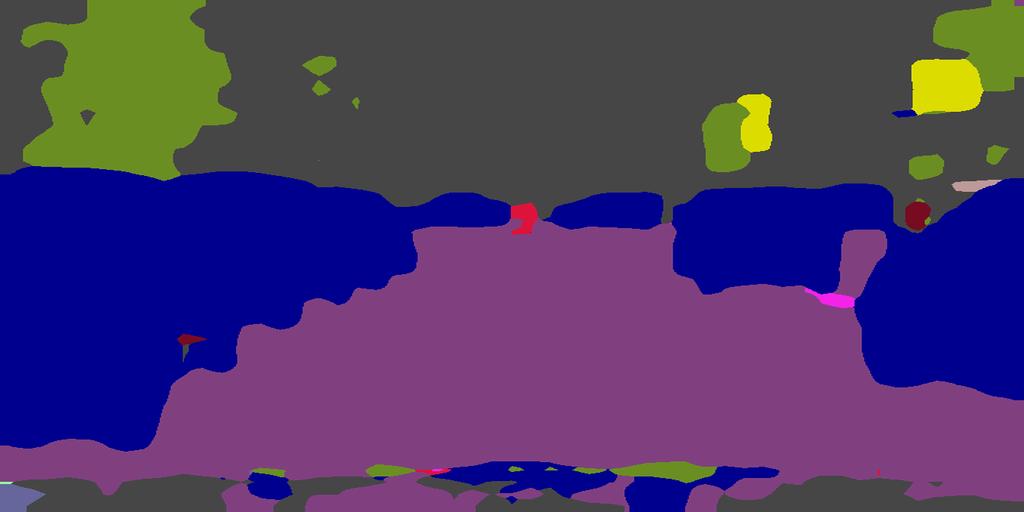} &\hspace{-0.47cm}
\includegraphics[width=0.188\linewidth, height=0.12\linewidth]{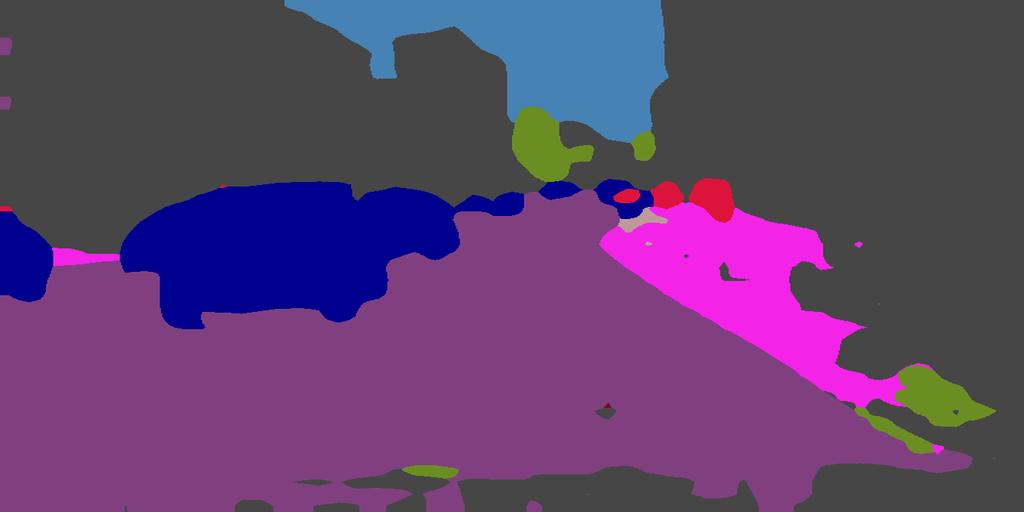} &\hspace{-0.47cm}
\includegraphics[width=0.188\linewidth, height=0.12\linewidth]{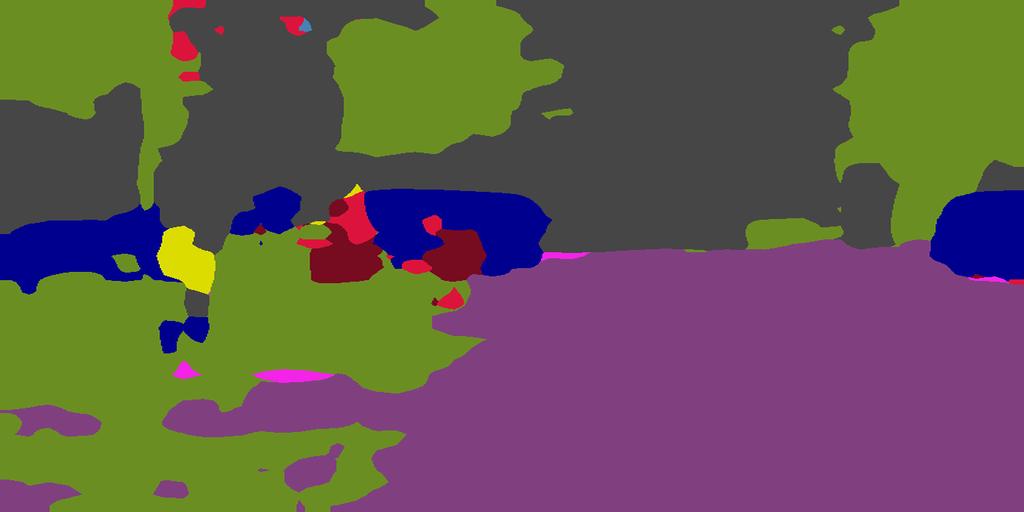   } &\hspace{-0.47cm}
\includegraphics[width=0.188\linewidth, height=0.12\linewidth]{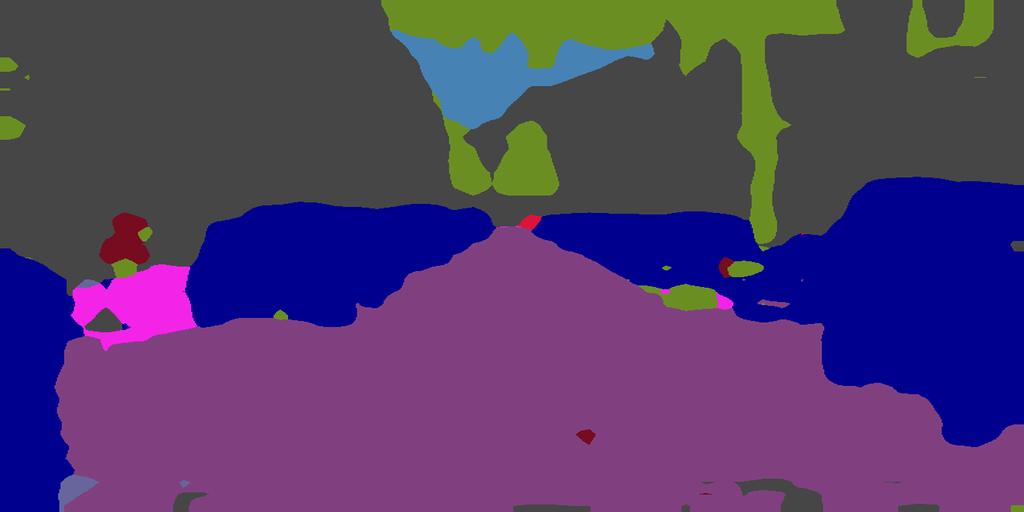  } &\hspace{-0.47cm}
\includegraphics[width=0.188\linewidth, height=0.12\linewidth]{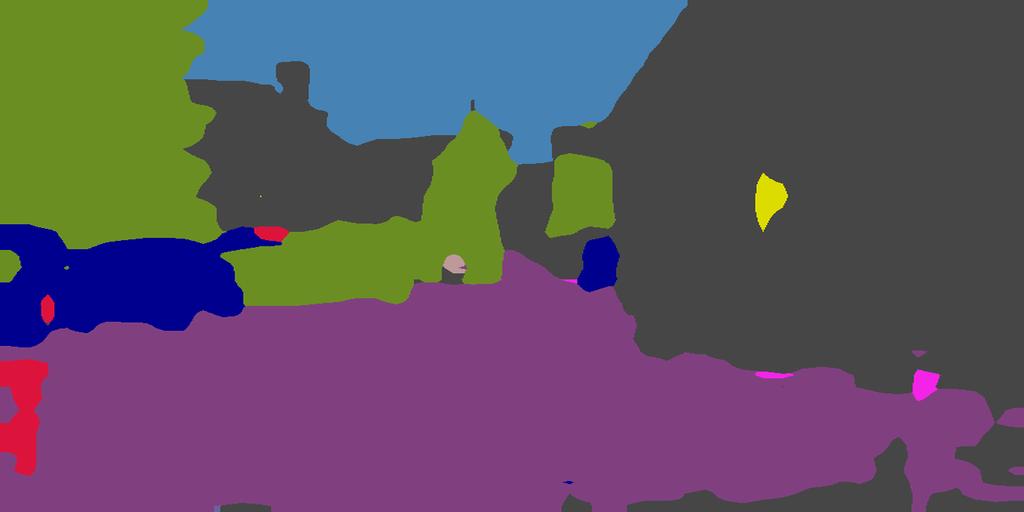  }\\
\hline

\verticaltext[27.5pt]{\textbf{w/ OmniISR}} &
\includegraphics[width=0.188\linewidth, height=0.12\linewidth]{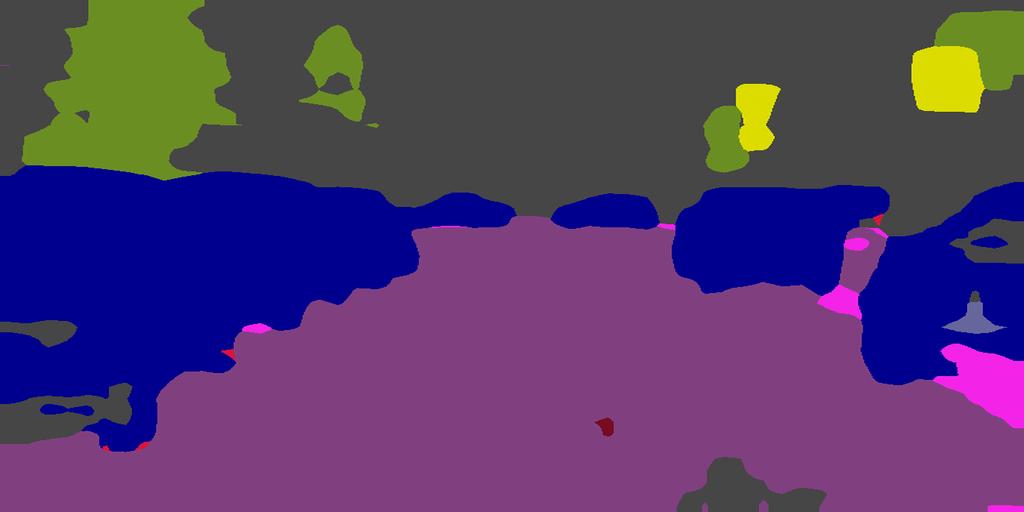} &\hspace{-0.47cm}
\includegraphics[width=0.188\linewidth, height=0.12\linewidth]{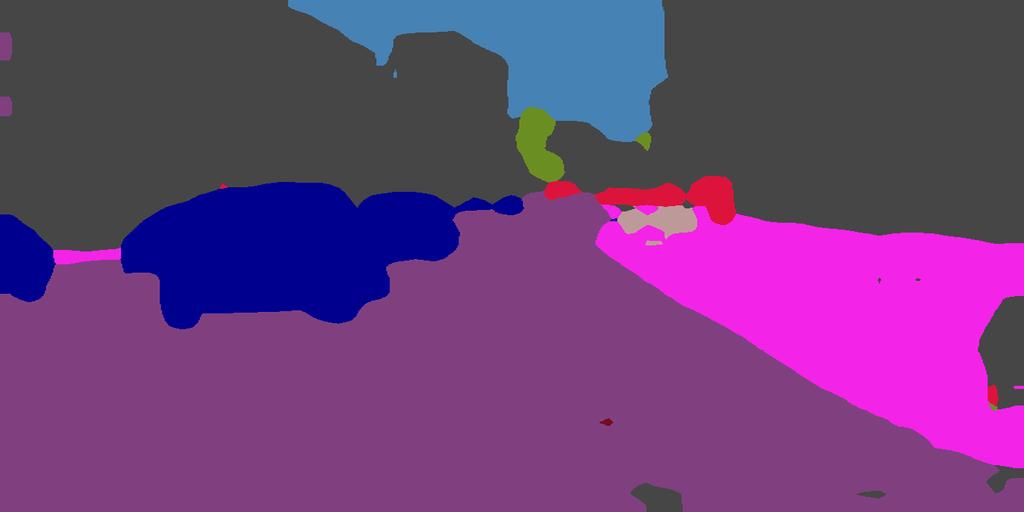} &\hspace{-0.47cm}
\includegraphics[width=0.188\linewidth, height=0.12\linewidth]{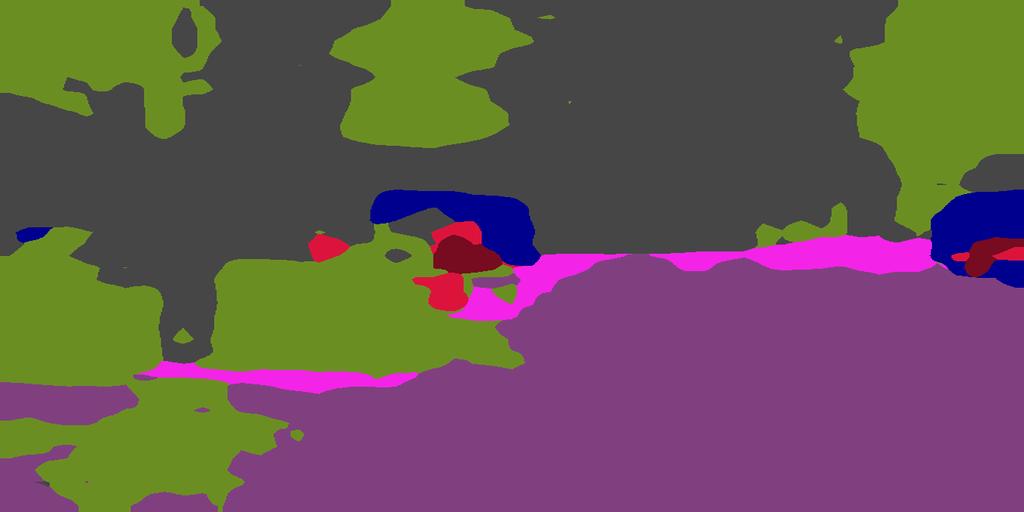   } &\hspace{-0.47cm}
\includegraphics[width=0.188\linewidth, height=0.12\linewidth]{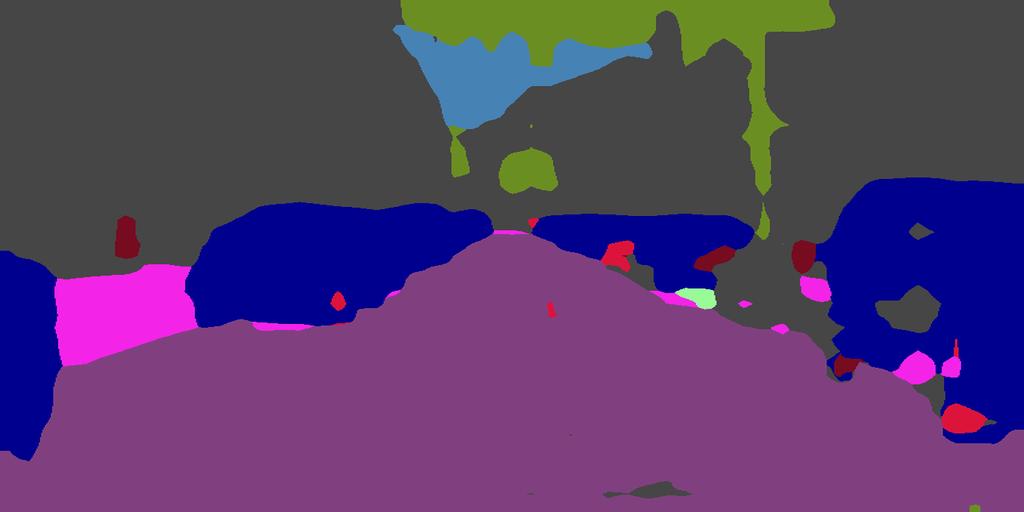  } &\hspace{-0.47cm}
\includegraphics[width=0.188\linewidth, height=0.12\linewidth]{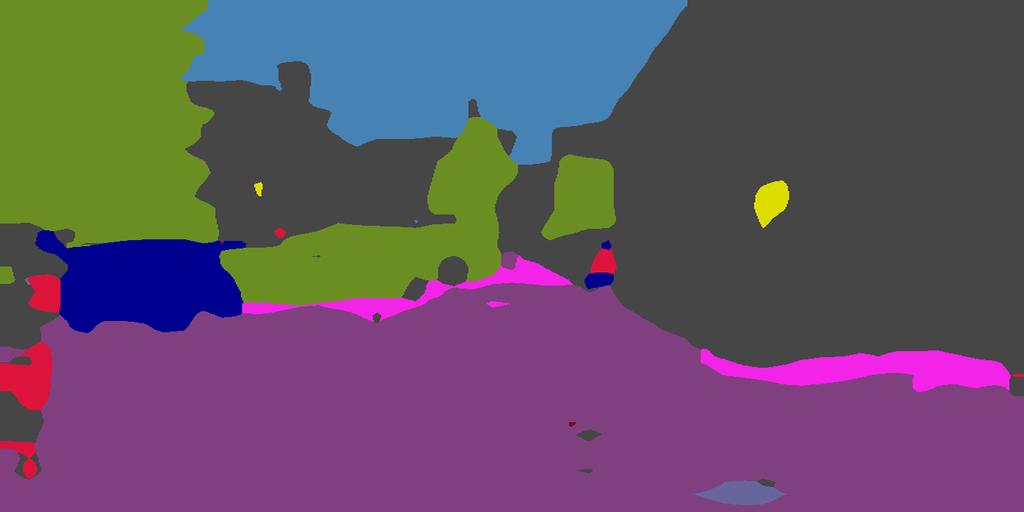  }\\
\hline
\end{tabularx}
\label{tab:iMacRS_qualitive_comp}
\vspace{-0.35cm}
\end{table*}

\subsection*{RQ2: Do OmniISR-achieved gains increase with heterogeneity and optimization difficulty?}

The answer is generally yes, but with informative exceptions. Under DeepLabv3+, Cityscapes exhibits substantially larger gains (CL: +5.39\%, FL: +9.19\%) than CamVid (CL: +0.14\%, FL: +0.63\%), consistent with the hypothesis that OmniISR contributes more obviously when client drift and class-boundary ambiguity are more severe. At the full-matrix level, this pattern is not perfectly monotonic across all models, which is because dataset difficulty/heterogeneity interacts with architecture inductive bias. The key implication is that OmniISR should be interpreted as a \emph{difficulty/heterogeneity-adaptive} mechanism: stronger benefit in harder non-IID regimes, near-saturated benefit in easier regimes.

\begin{figure}[t]
\centering
\subfloat[\footnotesize mIoU]{\includegraphics[width=0.48\linewidth]{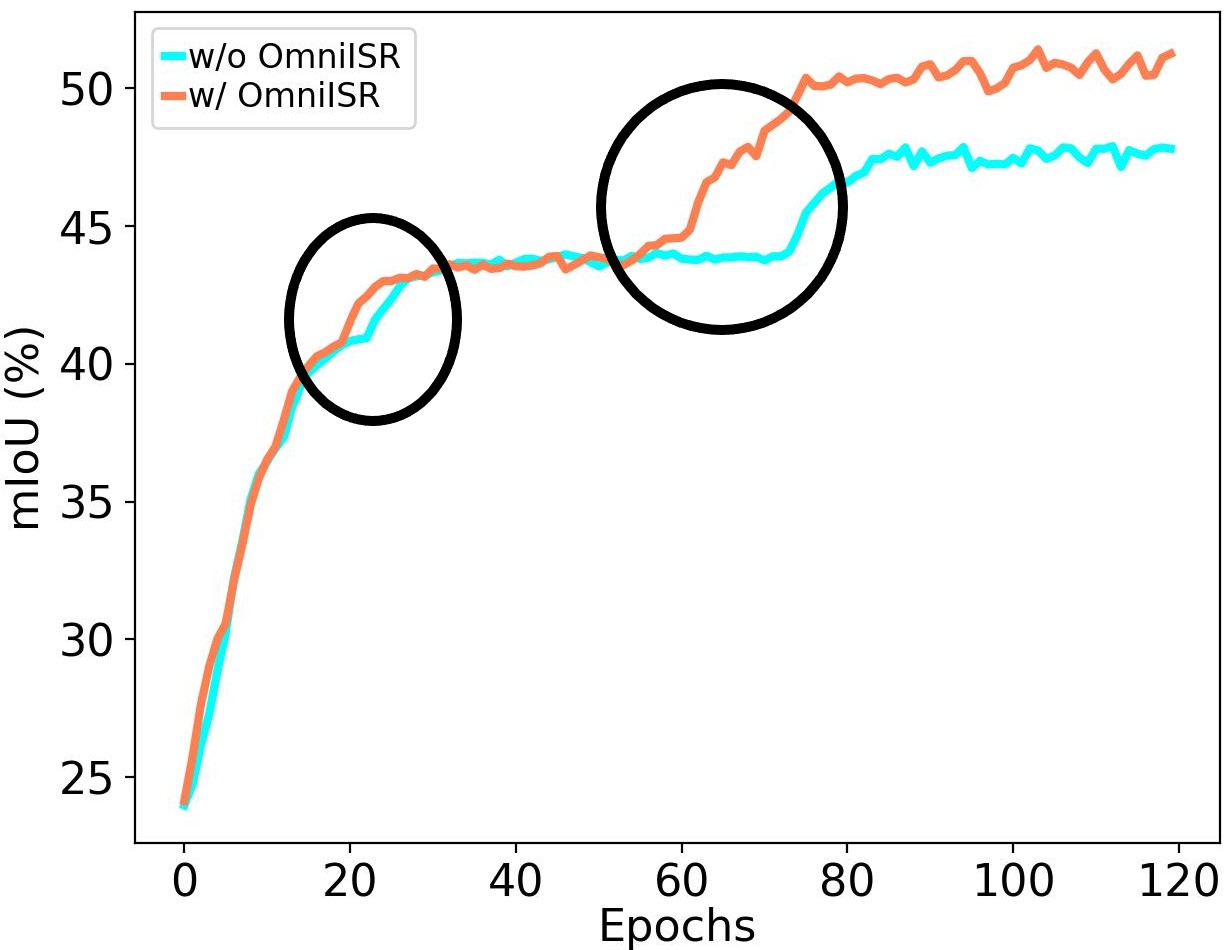}
\label{Fig.federated_mIoU}
}
\subfloat[\footnotesize mF1]{\includegraphics[width=0.48\linewidth]{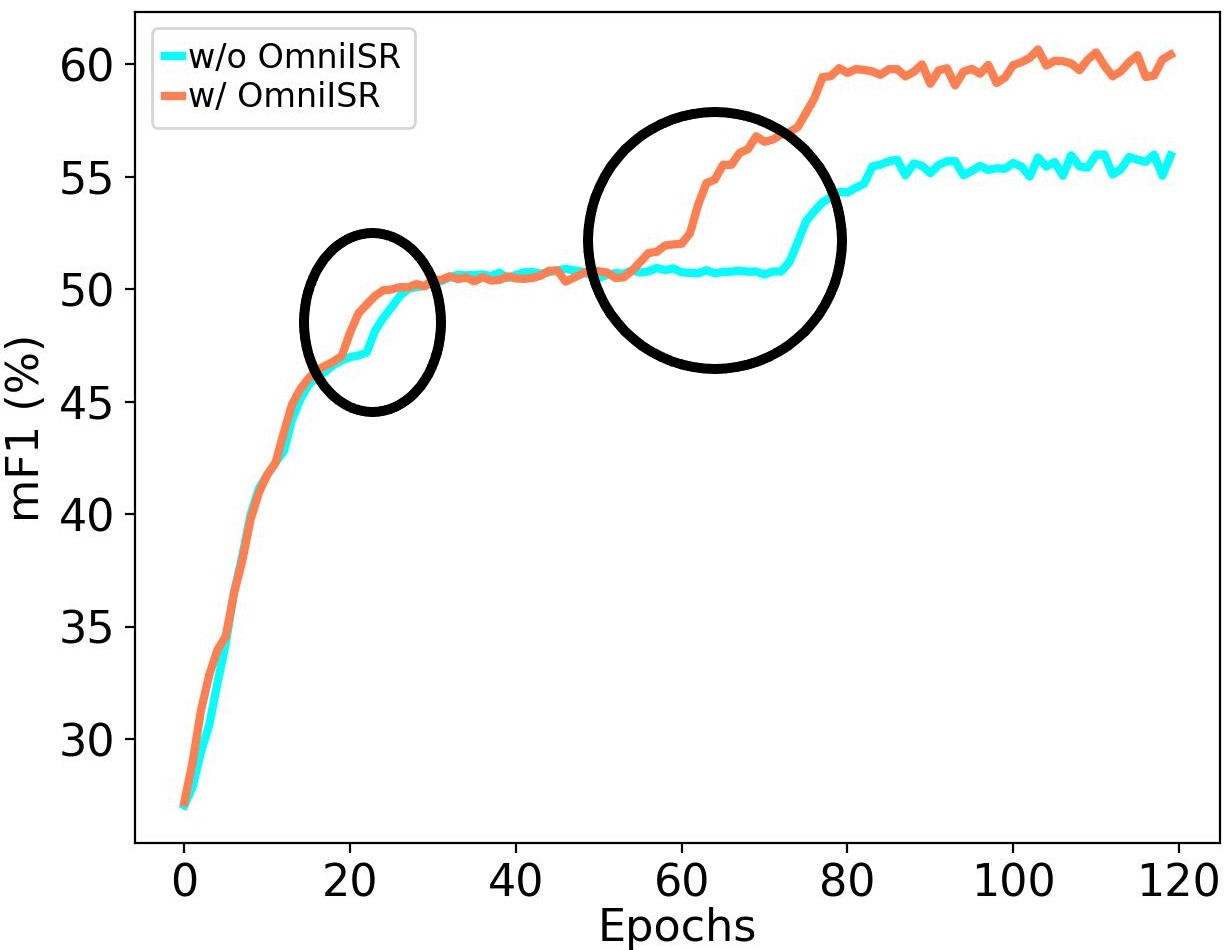}
\label{Fig.federated_mF1}
}
\vspace{-0.15cm}
\caption{Practical escape time comparison between OmniISR-enabled and OmniISR-disabled settings.}
\label{Fig.federated_Metrics}
\vspace{-0.5cm}
\end{figure}

\begin{figure*}[!t]
\includegraphics[width=\linewidth]{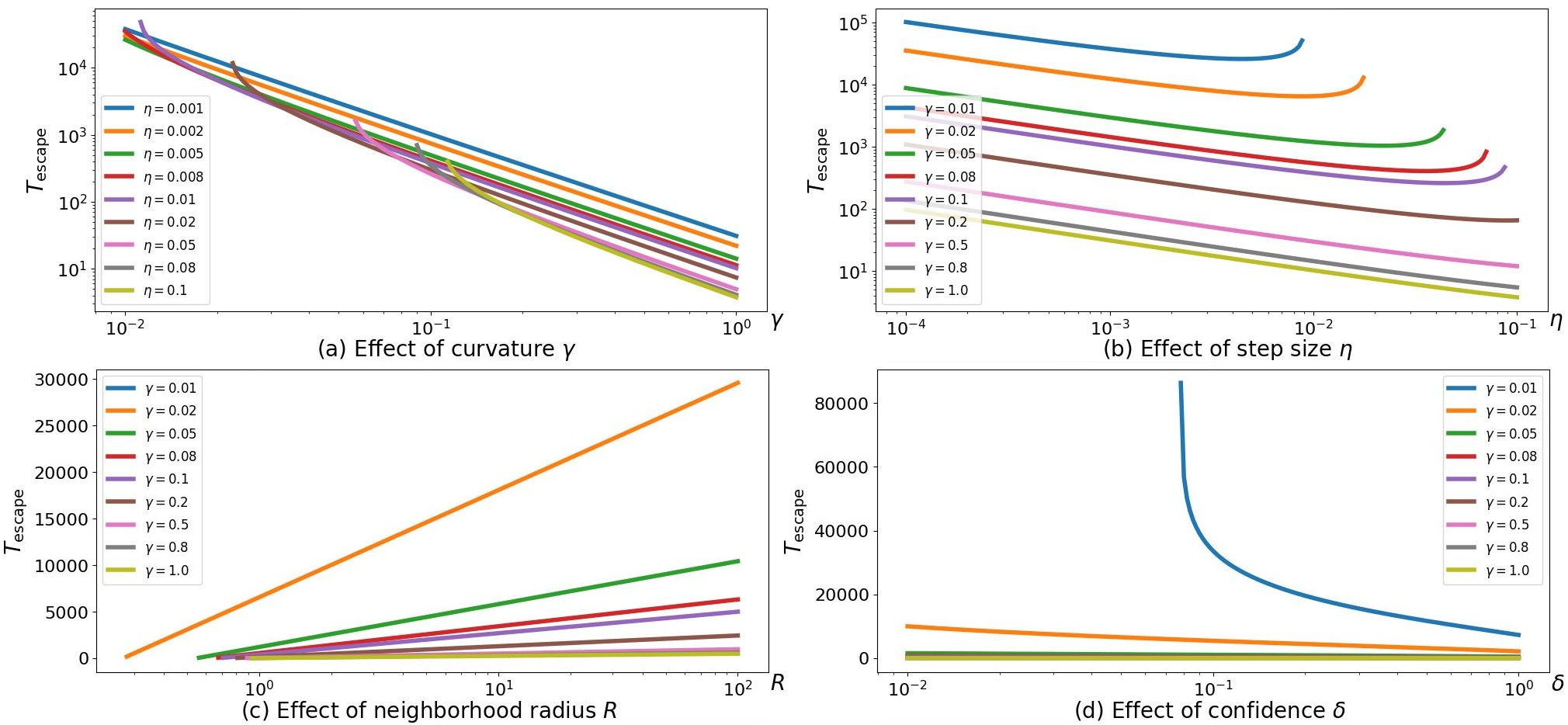}
\vspace{-0.7cm}
\caption{Evaluation of the time of escape saddle point for the proposed OmniISR framework.}
\label{Fig:escape_saddle_eval}
\vspace{-0.3cm}
\end{figure*}

\subsection*{RQ3: Is OmniISR architecture-agnostic or architecture-invariant?}

The evidence supports architecture-agnostic \emph{deployability} but not architecture-invariant \emph{effect size}. DeepLabv3+ shows the most stable gains, while Transformer-related models exhibit stronger condition dependence. For example, SeaFormer on SynthiaSF decreases slightly in CL (24.74 to 24.20 mIoU) but improves clearly in FL (16.29 to 18.43 mIoU). TopFormer gains are strong on CamVid (CL: 63.10 to 66.38; FL: 56.60 to 58.85) but almost neutral on SynthiaSF. This is a critical design message for practitioners: OmniISR can be inserted broadly, but point placement and coefficients of MI loss and NE regularizer  should be architecture-aware, especially for Transformer families.

\subsection*{RQ4: Does OmniISR improve cross-paradigm unification (reduce the CL--FL performance gap) beyond raising absolute accuracy?}

We quantify ``unification quality'' using the performance gap between CL and FL, defined as \(D = \mathrm{mIoU}_{\mathrm{CL}} - \mathrm{mIoU}_{\mathrm{FL}}\) (where a smaller \(D\) indicates superior unification). Across the nine evaluated model--dataset pairs, we analyze the individual gap \(D\) with and without OmniISR, the aggregate mean and median gaps, and the proportion of pairs exhibiting a reduced gap. Taking DeepLabv3+ as a representative example, OmniISR consistently reduces this dataset-level gap: from 4.15 to 2.71 on Cityscapes (\(\Delta = 1.44\), 34.7\% reduction), 3.24 to 2.89 on CamVid (\(\Delta = 0.35\), 10.8\%), and 6.63 to 5.25 on SynthiaSF (\(\Delta = 1.38\), 20.8\%). Consequently, the average gap for DeepLabv3+ decreases from 4.67 to 3.62 (\(\Delta = 1.05\), 22.6\%). Importantly, this improvement is \emph{not universal}. Across all nine architecture--dataset pairs, OmniISR reduces the gap in five instances but increases it in four. This variability highlights that the efficacy of CL--FL unification is highly sensitive to both the underlying data distribution and the specific model architecture, rather than being universally guaranteed.

\subsubsection{Qualitative Evaluation}
\Cref{tab:iMacRS_qualitive_comp} compares predictions with and without OmniISR on five representative urban scenes. Beyond visual "cleanliness", the key qualitative evidence is structural consistency. Specifically, OmniISR better preserves thin and boundary-sensitive regions (e.g., poles, object contours, and small foreground instances) while reducing fragmented misclassification in cluttered backgrounds. This observation is consistent with the design purpose of OmniISR: MI supervision strengthens semantic discriminability at intermediate layers, and NE regularization discourages overconfident early commitments that often produce brittle boundaries. The qualitative gains therefore corroborate the quantitative improvements by showing \textit{how} the improvement is realized in pixel space. \textit{In addition}, the quantitative pattern in \Cref{tab:OmniISR_quantitative_comp} shows that mPre often increases at least as much as mRec, suggesting that OmniISR primarily reduces false positives from ambiguous regions while maintaining recall, which is desirable for safety-critical AD perception pipelines.

\subsubsection{Escape Time Evaluation}
\Cref{Fig.federated_Metrics} reports the practical escape-time comparison between OmniISR-enabled and OmniISR-disabled training. The curves show that OmniISR reaches the post-saddle acceleration phase earlier on both mIoU and mF1 (black circles). This result is an optimization-speed gain, where fewer rounds to exit flat/unstable regions translate into lower communication and computation budgets. This behavior supports our theoretical claim of gradient synergy: intermediate supervision stabilizes useful directions, while regularization-induced stochasticity prevents overconfident trapping, jointly improving the probability of fast escape from poor stationary regions. 

To further illustrate the theoretical properties of the proposed OmniISR framework, \Cref{Fig:escape_saddle_eval} analyzes its saddle escape time \(T_{\mathrm{escape}}\). Specifically, the figure evaluates the sensitivity of \(T_{\mathrm{escape}}\) with respect to four key parameters: the curvature \(\gamma\), the step size \(\eta\), the neighborhood radius \(R\), and the confidence level \(\delta\). Each subfigure systematically sweeps one primary variable while holding the others constant, utilizing multiple curves to demonstrate the compounding influence of a secondary parameter.

\subsection*{(a) Effect of curvature \texorpdfstring{$\gamma$}{gamma}}

\Cref{Fig:escape_saddle_eval}(a) illustrates the effect of the curvature parameter \texorpdfstring{\(\gamma\)}{gamma}, on the saddle escape time \(T_{\mathrm{escape}}\). The log--log plot reveals that \(\gamma\) is the dominant factor governing the escape dynamics. Specifically, \(T_{\mathrm{escape}}\) exhibits a strong inverse dependence on the curvature, decreasing across the evaluated curvature range. Furthermore, the escape time scales inversely with the step size \(\eta\), which is evidenced by the vertical spacing between the curves. However, in the weak-curvature regime (i.e., for small values of \(\gamma\)), the curves bend sharply upward and terminate prematurely. This divergence indicates that stochastic noise begins to dominate the gradient information, causing the theoretical bound to break down. Overall, these findings emphasize that even a marginal increase in curvature drastically accelerates the saddle escape process.

\subsection*{(b) Effect of step size \texorpdfstring{$\eta$}{eta}}
\Cref{Fig:escape_saddle_eval}(b) illustrates the influence of the step size \texorpdfstring{\(\eta\)}{eta} on the escape time \(T_{\mathrm{escape}}\). For most curvature values, \(T_{\mathrm{escape}}\) decreases gradually as \(\eta\) increases, exhibiting a gentle downward slope on the logarithmic scale. However, in the weak-curvature regime (e.g., \(\gamma \in \{0.01, 0.02, 0.05\}\)), the dynamics display a pronounced non-monotonic behavior. Beyond a critical step size threshold, the escape time rises abruptly, forming a distinct U-shape with a well-defined minimum. The location of this minimum, which represents the optimal step size, shifts toward larger values of \(\eta\) as the curvature \(\gamma\) increases. Ultimately, these findings demonstrate a critical trade-off: while larger step sizes generally accelerate the escape process, exceeding the optimal \(\eta\) becomes highly detrimental when the curvature is small, leading to a severe divergence in escape time.

\subsection*{(c) Effect of neighborhood radius \texorpdfstring{$R$}{R}}
\Cref{Fig:escape_saddle_eval}(c) illustrates the impact of the neighborhood radius \texorpdfstring{\(R\)}{R}, on the escape time \(T_{\mathrm{escape}}\). When plotted on a semi-logarithmic scale (linear \(T_{\mathrm{escape}}\) versus logarithmic \(R\)), the curves exhibit an approximately linear trajectory, indicating that \(T_{\mathrm{escape}}\) grows only logarithmically with \(R\). This logarithmic dependence demonstrates that the neighborhood radius exerts a relatively weak influence on the escape dynamics. Instead, the overall magnitude of the escape time is governed predominantly by the curvature \(\gamma\). This is evident from the distinct vertical stratification of the curves: the weak-curvature case (\(\gamma = 0.02\)) dominates the plot, reaching \(T_{\mathrm{escape}} \approx 3 \times 10^{4}\) at \(R = 100\), whereas the curves corresponding to stronger curvatures (\(\gamma \gtrsim 0.1\)) remain essentially flat and near zero across the entire range of \(R\). Furthermore, in the extreme weak-curvature scenario (\(\gamma = 0.01\)), the curve is absent at small values of \(R\). This absence indicates that the theoretical bound becomes physically meaningless.

\subsection*{(d) Effect of confidence \texorpdfstring{$\delta$}{delta}}
\Cref{Fig:escape_saddle_eval}(d) illustrates the influence of the confidence level \texorpdfstring{\(\delta\)}{delta} on the saddle escape time \(T_{\mathrm{escape}}\). For moderate to large values of the curvature parameter \(\gamma\), the curves remain essentially flat across the evaluated range of \(\delta\). This indicates that the required confidence level has a negligible impact on the escape dynamics, with the corresponding escape times remaining compressed near the bottom of the plot and practically indistinguishable from zero on a linear scale. However, a stark contrast emerges in the extreme weak-curvature regime. Specifically, for \(\gamma = 0.01\), the escape time rises steeply as \(\delta\) decreases, peaking at approximately \(9 \times 10^{4}\) near \(\delta \approx 10^{-1}\). The curve corresponding to \(\gamma = 0.02\) exhibits a similar, albeit significantly less pronounced, upward trajectory. Ultimately, these observations demonstrate that demanding higher statistical confidence incurs almost no penalty in escape time, except when the curvature is exceptionally small, where the cost becomes severe.

\subsection*{Overall Synthesis}

\begin{table}[t]
  \centering
  \setlength{\tabcolsep}{5.0pt}
  \caption{Qualitative summary of the dependencies of the escape time \(T_{\mathrm{esc}}\).}
  \vspace{-0.2cm}
  \begin{tabularx}{\linewidth}{ccc}
    \toprule
    Parameter & Observed Trend & Practical Impact \\
    \midrule
    \(\gamma\)  & Pronounced log--log decay & \textbf{Dominant} \\ \hline
    \(\eta\)    & \makecell[c]{Non-monotonic \\U-shaped dependence} & \makecell[c]{Strong \\(exhibits an optimum)} \\ \hline
    \(R\)       & Mild logarithmic growth & Weak \\ \hline
    \(\delta\)  & \makecell[c]{Negligible variation; \\diverges  only for minimal \(\gamma\)} & Mostly negligible \\
    \bottomrule
  \end{tabularx}
  \label{tab:sensitivity}
  \vspace{-0.3cm}
\end{table}

\begin{table}[tp]
\centering
\setlength{\tabcolsep}{7.0pt}
\caption{Quantitative performance comparison of OmniISR-enabled case against OmniISR-disabled case across multiple FL algorithms}
\vspace{-0.2cm}
\begin{tabularx}{\linewidth}{c|c|cccc}
\hline
\multirow{1}{*}{FL Algorithm} & \multirow{1}{*}{OmniISR?} &mIoU    &mF1     &mPre     &mRec     \\ \hline
\multirow{2}{*}{FedAvg} & \xmark                &47.91   &56.31   &59.99    &55.58    \\ 
                          & \checkmark          &\textbf{50.49}   &\textbf{59.70}   &\textbf{65.00}    &\textbf{57.82}    \\ \hline
\multirow{2}{*}{FedProx (0.005)} & \xmark       &41.41   &47.52   &47.69    &48.31    \\
                          & \checkmark          &\textbf{42.45}   &\textbf{48.53}   &\textbf{48.61}    &\textbf{49.36}    \\ \hline
\multirow{2}{*}{FedProx (0.01)} & \xmark        &31.74   &35.74   &35.82    &37.31    \\
                         & \checkmark           &\textbf{32.84}   &\textbf{36.82}   &\textbf{36.35}    &\textbf{38.38}    \\ \hline
\multirow{2}{*}{FedDyn (0.005)} & \xmark        &28.63   &31.95   &31.93    &33.95    \\
                         & \checkmark           &\textbf{29.75}   &\textbf{33.05}   &\textbf{32.31}    &\textbf{35.01}    \\ \hline
\multirow{2}{*}{FedDyn (0.01)} & \xmark         &25.19   &27.92   &26.63    &29.43    \\
                         & \checkmark           &\textbf{26.19}   &\textbf{28.92}   &\textbf{27.64}    &\textbf{30.43}    \\ \hline
\multirow{2}{*}{FedAvgM (0.7)} & \xmark         &51.49   &60.74   &65.37    &58.96    \\
                         & \checkmark           &\textbf{52.59}   &\textbf{61.84}   &\textbf{66.66}    &\textbf{59.93}    \\ \hline
\multirow{2}{*}{FedAvgM (0.9)} & \xmark         &51.65   &60.91   &65.55    &59.12    \\
                         & \checkmark           &\textbf{52.74}   &\textbf{61.93}   &\textbf{66.63}    &\textbf{60.39}    \\ \hline
\multirow{2}{*}{FedGau} & \xmark                &52.03   &62.23   &68.98    &59.36    \\
                         & \checkmark           &\textbf{54.40}   &\textbf{64.43}   &\textbf{71.94}    &\textbf{61.49}    \\ \hline
\multirow{2}{*}{FedIR} & \xmark                 &\textbf{25.83}   &\textbf{28.33}   &\textbf{27.91}    &\textbf{29.02}    \\
                         & \checkmark           &25.81   &28.32   &27.90    &28.98    \\ \hline
\multirow{2}{*}{MOON} & \xmark                  &51.48   &60.75   &65.37    &59.02    \\
                         & \checkmark           &\textbf{52.49}   &\textbf{61.67}   &\textbf{66.63}    &\textbf{59.70}    \\ \hline
\multirow{2}{*}{SCAFFOLD} & \xmark              &\textbf{24.19}   &\textbf{27.22}   &26.32    &\textbf{28.27}    \\
                         & \checkmark           &23.63   &26.79   &\textbf{26.38}    &27.49    \\ \hline
\multirow{2}{*}{BalanceFL} & \xmark             &\textbf{52.15}   &\textbf{61.36}   &\textbf{65.82}    &\textbf{59.53}    \\
                         & \checkmark           &51.48   &60.78   &65.54    &58.95    \\ \hline
\end{tabularx}
\label{tab:Federated_OmniISR_perf}
\vspace{-0.3cm}
\end{table}

A summary of the four parametric studies, provided in \Cref{tab:sensitivity}, highlights the key factors of the escape time. Most importantly, the curvature \(\gamma\) is the dominant factor controlling the overall escape time \(T_{\mathrm{escape}}\). It determines the baseline magnitude of the escape time and causes the largest variations across all of our evaluations. Additionally, the step size \(\eta\) plays a critical role, possessing a clear optimal value for minimizing the escape time. If the step size exceeds this optimal threshold, performance drops sharply, especially when the curvature is weak. In contrast, the neighborhood radius \(R\) and the confidence level \(\delta\) act as minor factors. Their impact on \(T_{\mathrm{escape}}\) is mostly negligible, only becoming significant in extreme, weak-curvature scenarios where the theoretical bounds are highly sensitive.

\subsection{OmniISR-Augmented FL Algorithms}
To assess whether OmniISR is merely compatible with FedAvg-like updates or genuinely \emph{optimizer-agnostic}, we perform a stress test across 12 FL algorithms spanning proximal correction (FedProx), drift correction (SCAFFOLD), dynamic regularization (FedDyn), momentum aggregation (FedAvgM), contrastive alignment (MOON), imbalance-aware optimization (FedIR/BalanceFL), and our prior AD-oriented variant (FedGau). 

\textbf{Result 1: broad positive transfer.} Across 48 paired comparisons (12 algorithms \(\times\) 4 metrics), OmniISR demonstrates improved performance in 37 cases (77.10\%). Furthermore, the performance gains over strong representative baselines remain substantial. For instance, when compared to FedGau, OmniISR achieves improvements of 4.56\%, 3.54\%, 4.29\%, and 3.59\% across the mIoU, mF1, mPre, and mRec metrics, respectively. These consistent enhancements indicate that the proposed OmniISR's intermediate mechanism is highly adaptable and not restricted to any single FL aggregation algorithm.

\textbf{Result 2: gain stratification reveals when OmniISR is most useful.} 
We observe distinct tiers of performance improvement: (i) \emph{high-gain} methods (e.g., FedAvg, FedAvgM, FedGau, and MOON); (ii) \emph{moderate-gain} methods (e.g., FedProx and FedDyn); and (iii) \emph{marginal-to-negative} methods (e.g., FedIR, SCAFFOLD, and BalanceFL). This stratification indicates that OmniISR yields the most significant improvements when the base FL algorithm lacks robust, built-in representation-drift control mechanisms. Conversely, the benefits are less pronounced when drift correction is already heavily encoded into the base FL algorithm itself.

\begin{figure*}[t]
\centering
\subfloat[\footnotesize mIoU]{\includegraphics[width=0.24\linewidth]{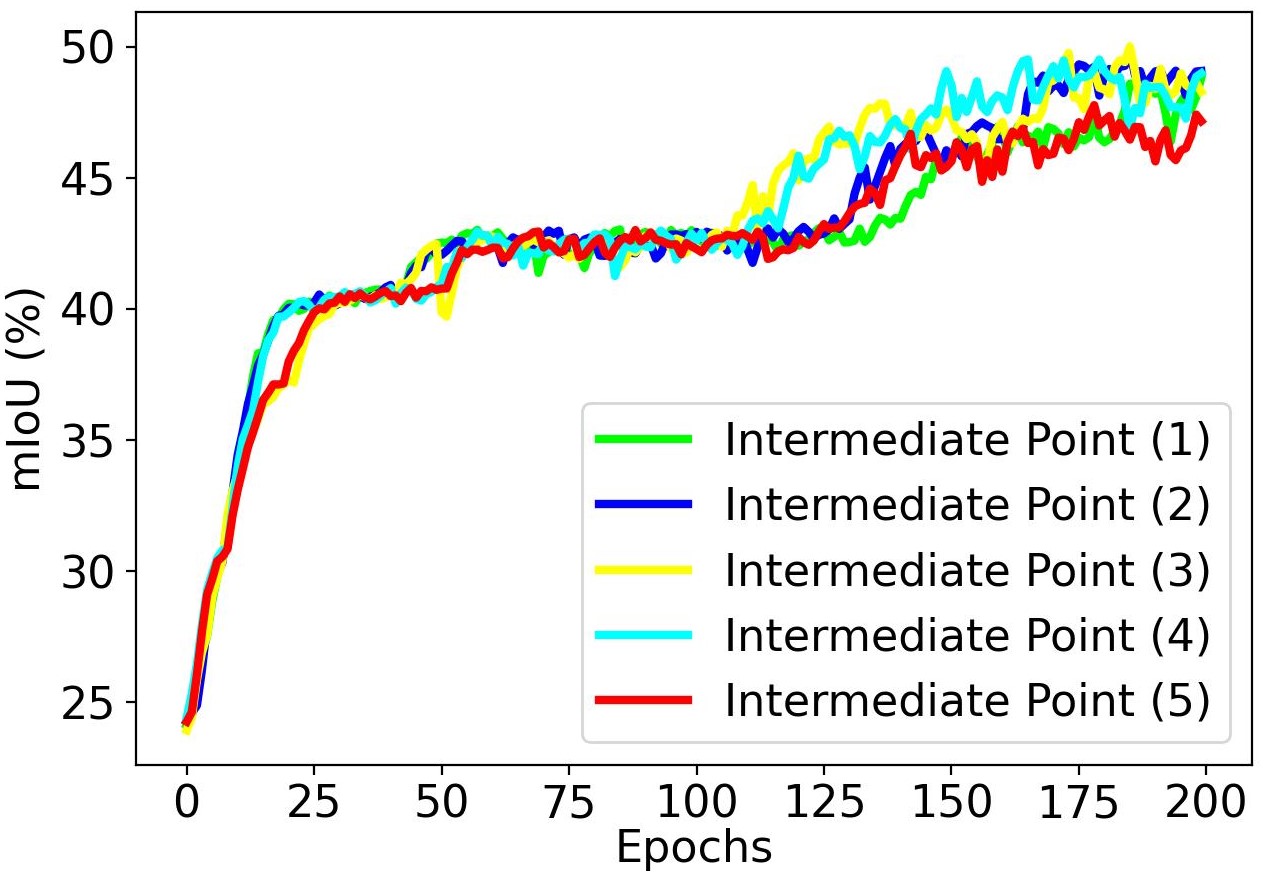}
\label{Fig.abl_number_Metrics_mIoU}
}
\subfloat[\footnotesize mPrecision]{\includegraphics[width=0.24\linewidth]{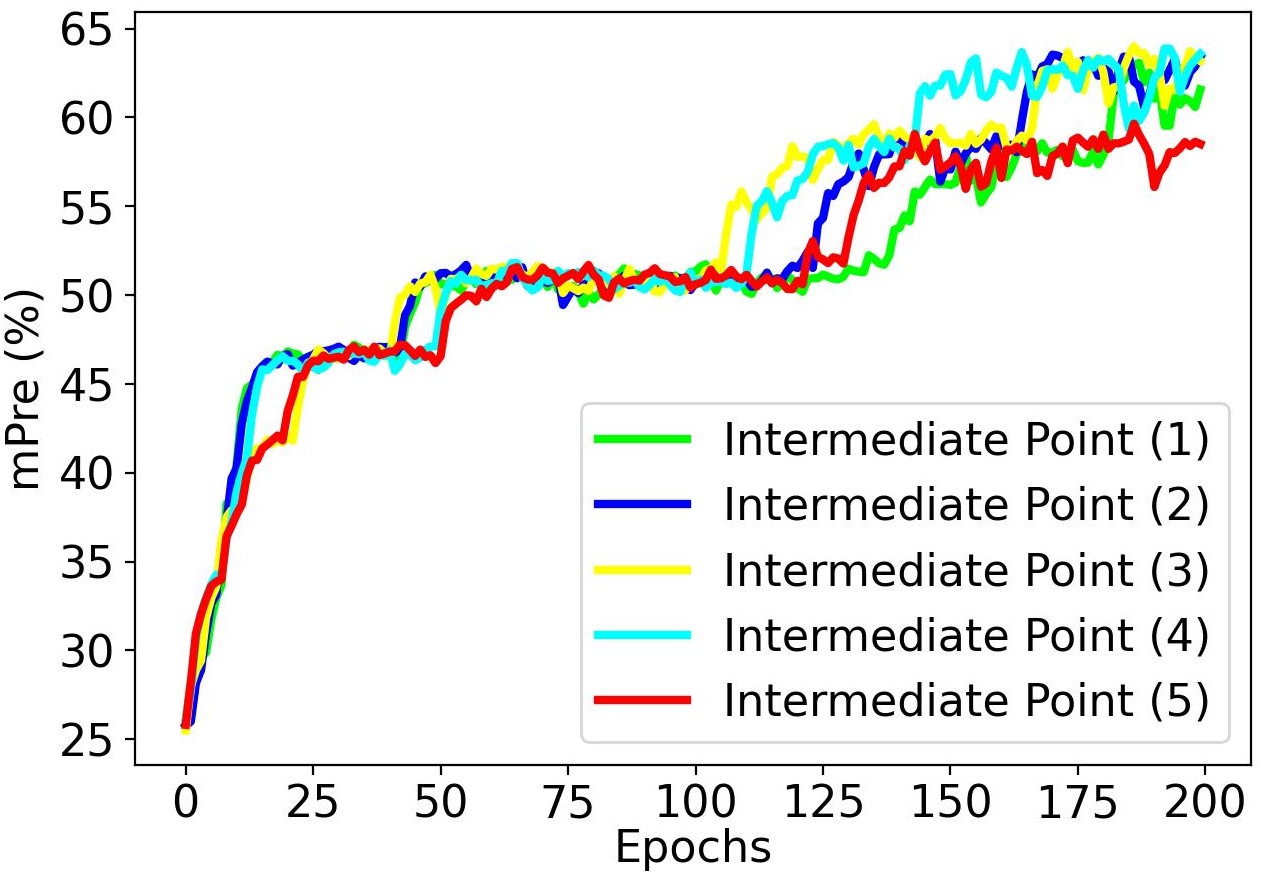}
\label{Fig.abl_number_Metrics_mPre}
}
\subfloat[\footnotesize mRecall]{\includegraphics[width=0.24\linewidth]{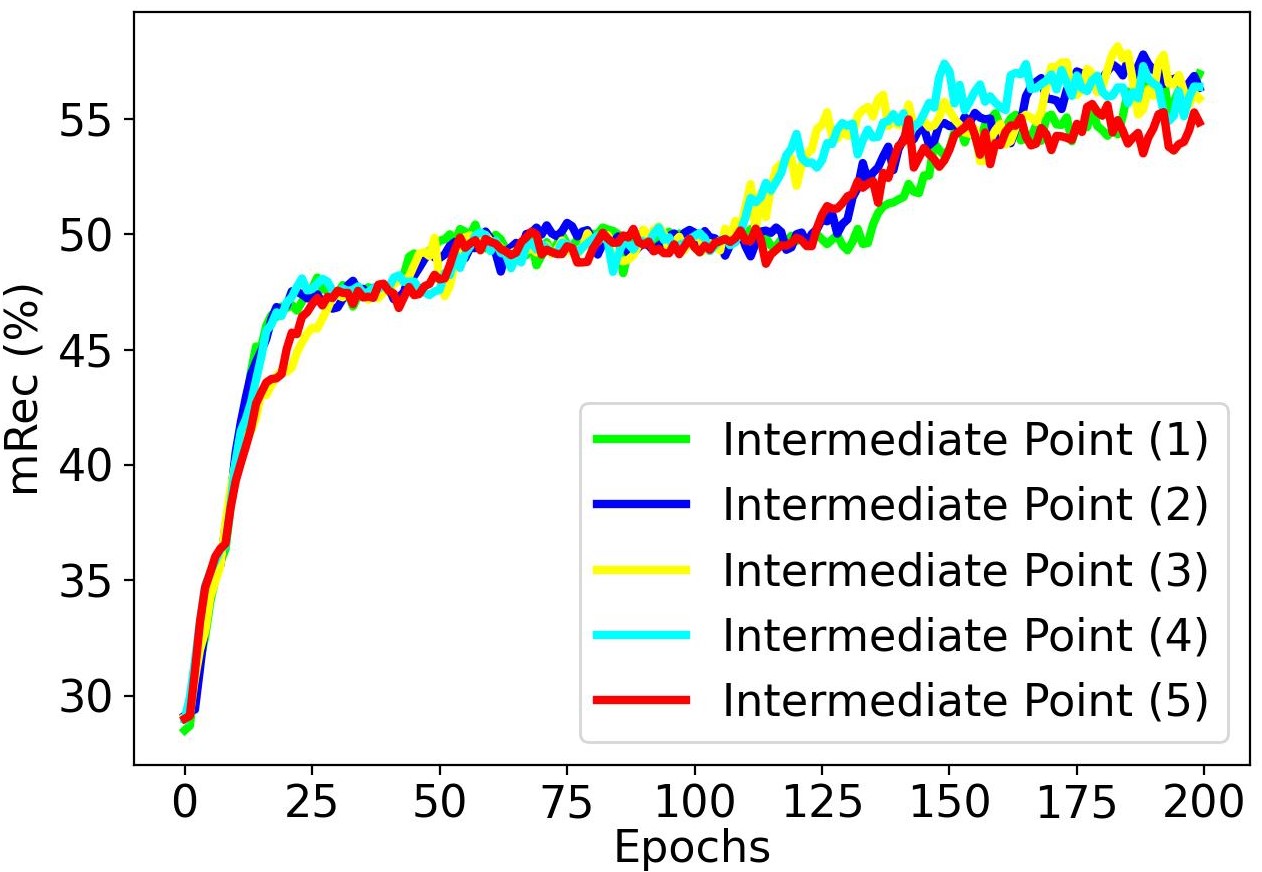}
\label{Fig.abl_number_Metrics_mRec}
}
\subfloat[\footnotesize mF1]{\includegraphics[width=0.24\linewidth]{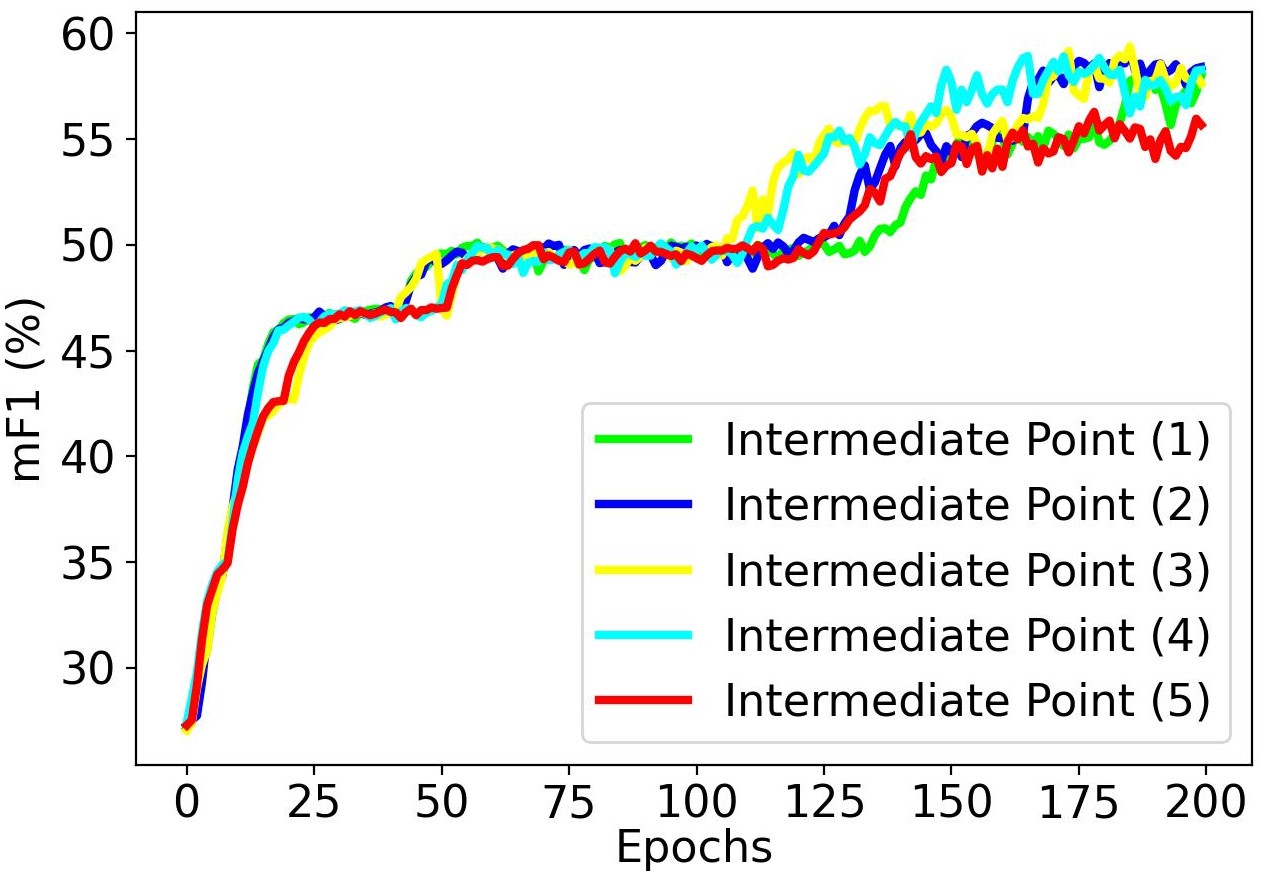}
\label{Fig.abl_number_Metrics_mF1}
}
\vspace{-0.15cm}
\caption{The impact of the number of intermediate points on OmniISR's training performance.}
\label{Fig.abl_number_Metrics}
\vspace{-0.5cm}
\end{figure*}

\subsection{Ablation Studies}
\label{sec:ablation}

This subsection details three ablation studies designed to evaluate the configuration of intermediate points in OmniISR. We assess how OmniISR's inference performance is affected by (i) the number of intermediate points deployed, (ii) the interval distance between adjacent points, and (iii) the structural placement of these points.

\subsubsection{Impact of the Number of intermediate points}
To quantify how supervision granularity affects OmniISR, we evaluate 1--5 intermediate points in both CL and FL settings (i.e., ``Intermediate Point (1)'' to ``Intermediate Point (5)''). As shown in \Cref{Fig.abl_number_Metrics}, performance peaks at a moderate number of points, revealing a clear bias--variance style trade-off in representation shaping: too few points under-constrain latent drift, while too many points over-constrain feature evolution and reduce representational flexibility. This observation is consistent with our core claim that intermediate MI supervision should guide rather than dominate hidden representations. In practical, OmniISR's deployment should therefore consider a moderate point count instead of maximal insertion.

\begin{figure*}[t]
\centering
\subfloat[\footnotesize mIoU]{\includegraphics[width=0.24\linewidth]{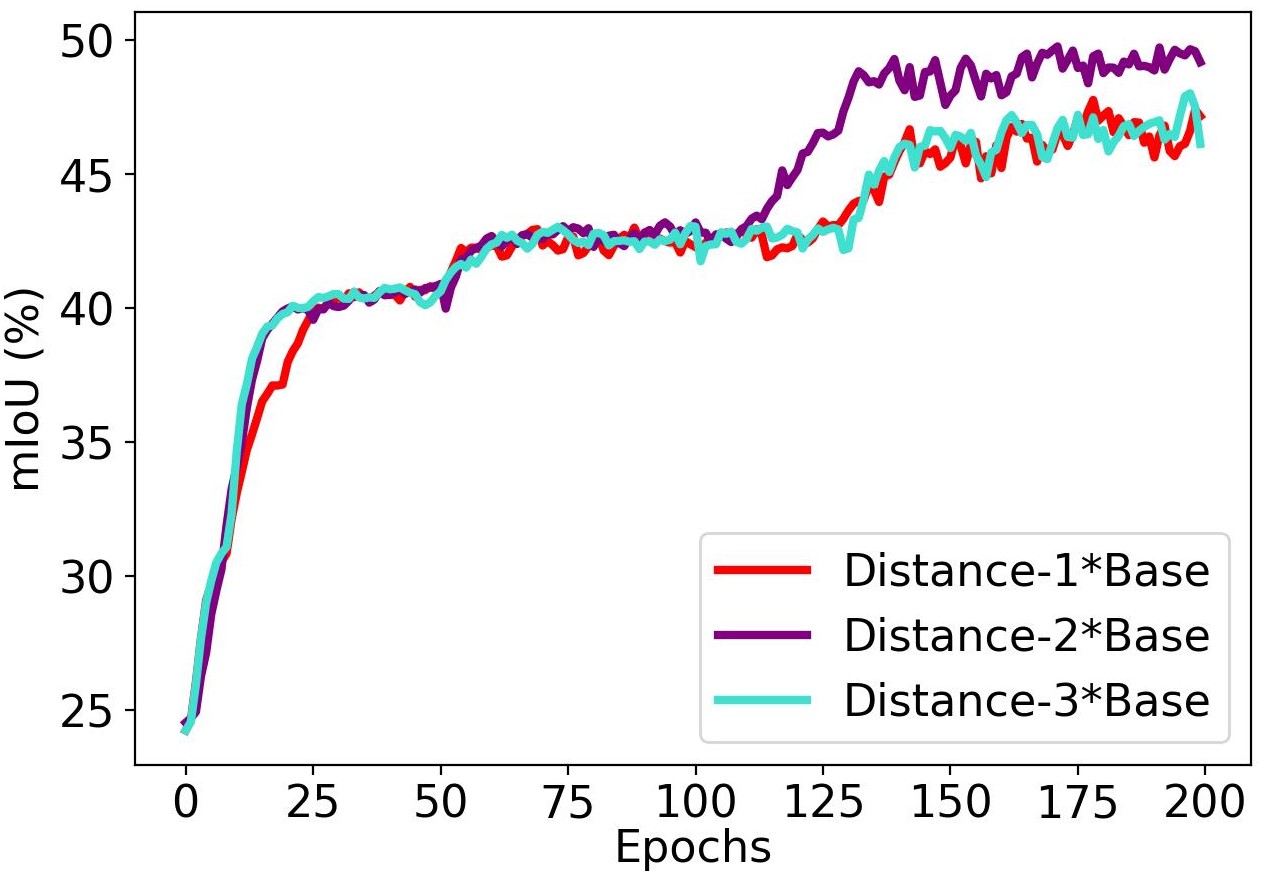}
\label{Fig.abl_dist_Metrics_mIoU}
}
\subfloat[\footnotesize mPrecision]{\includegraphics[width=0.24\linewidth]{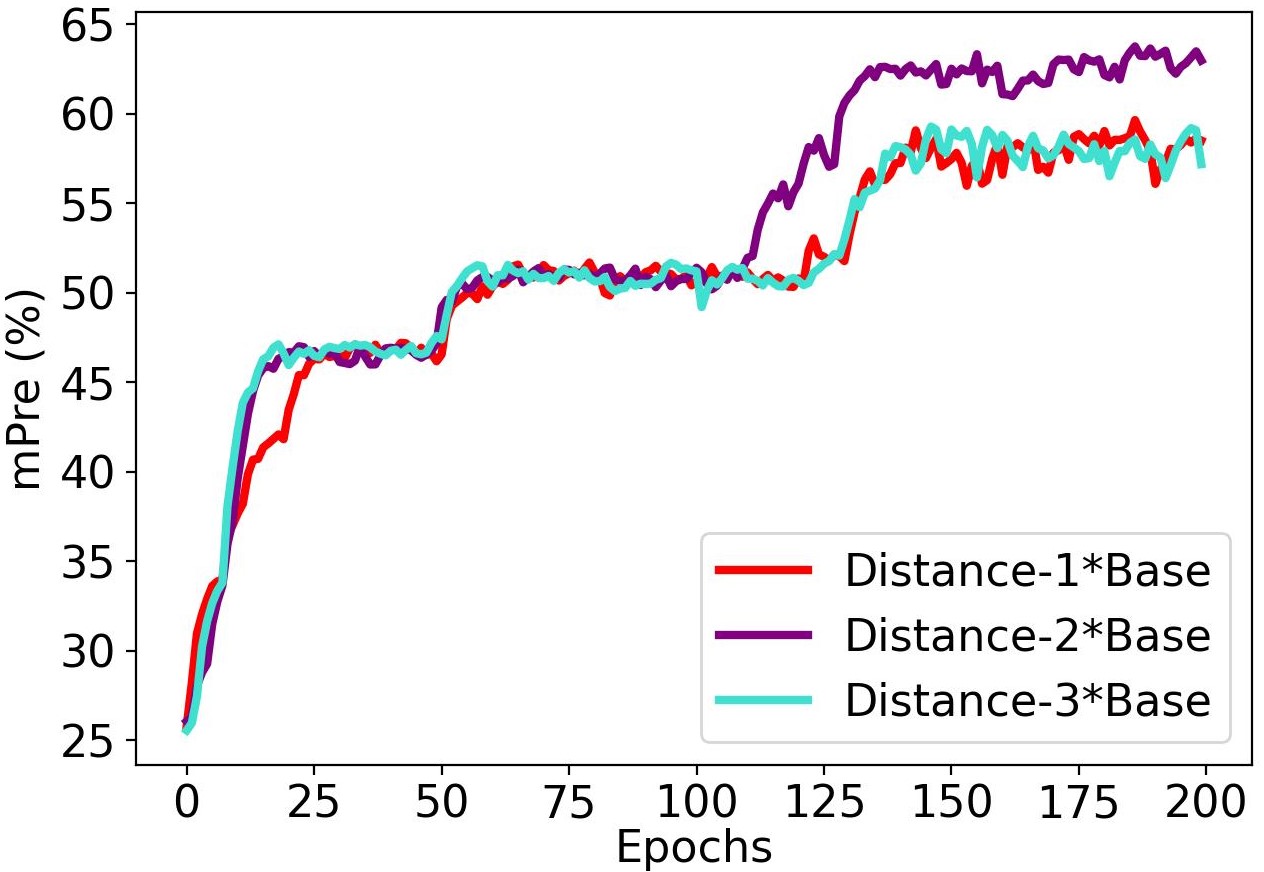}
\label{Fig.abl_dist_Metrics_mPre}
}
\subfloat[\footnotesize mRecall]{\includegraphics[width=0.24\linewidth]{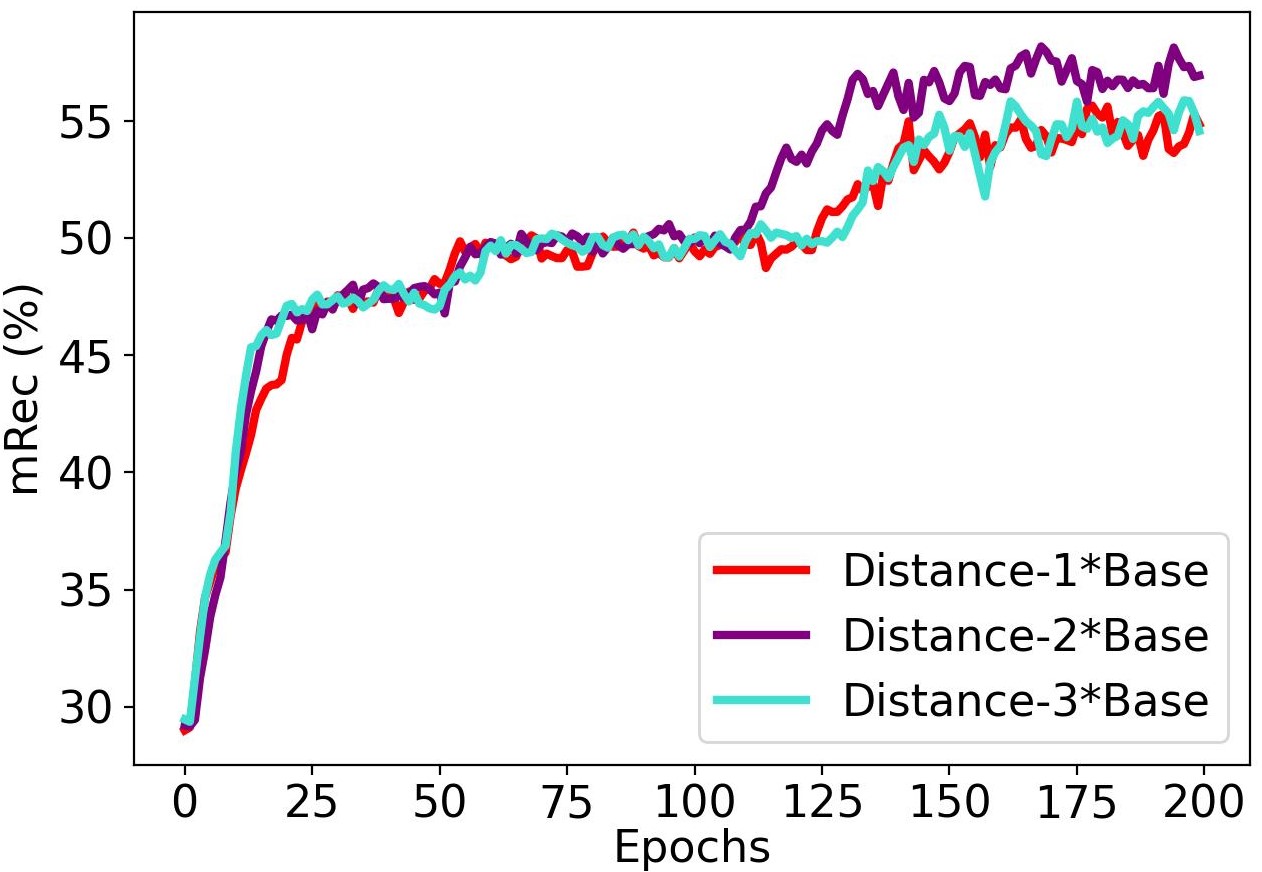}
\label{Fig.abl_dist_Metrics_mRec}
}
\subfloat[\footnotesize mF1]{\includegraphics[width=0.24\linewidth]{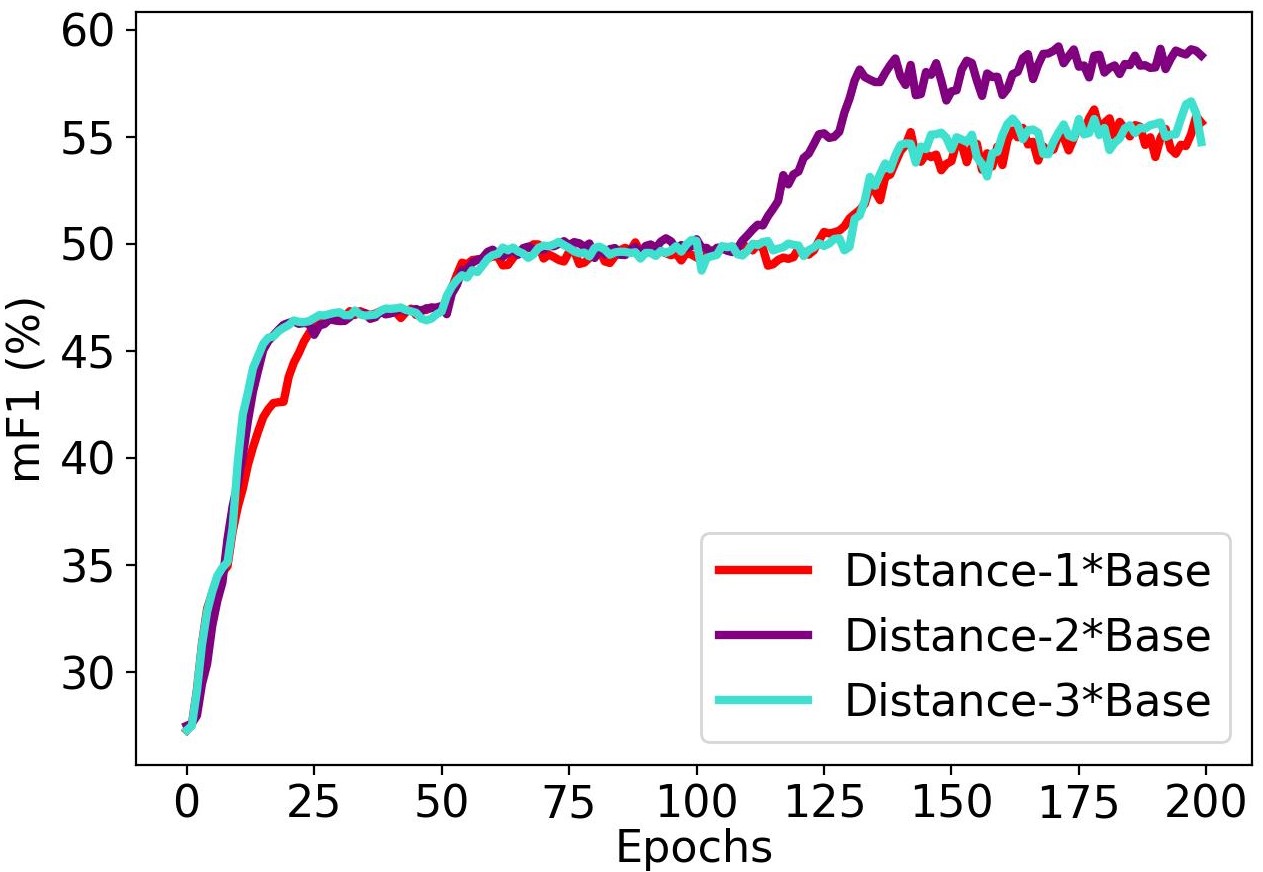}
\label{Fig.abl_dist_Metrics_mF1}
}
\vspace{-0.15cm}
\caption{The impact of the distance between adjacent intermediate points on OmniISR's training performance.}
\label{Fig.abl_dist_Metrics}
\vspace{-0.3cm}
\end{figure*}

\subsubsection{Impact of the distance between adjacent points}
To examine spatial coupling between supervision points, we define a base layer distance and test three configurations: 1-base, 2-base, and 3-base spacing for both CL and FL. \Cref{Fig.abl_dist_Metrics} shows that the 2-base spacing consistently yields the best performance. A plausible interpretation is that 1-base spacing introduces redundant constraints on highly correlated adjacent features, while 3-base spacing is too sparse to propagate stable intermediate guidance across depth. Thus, moderate spacing appears to best balance local feature consistency and global semantic abstraction, which is precisely where OmniISR is expected to be most effective.

\begin{table*}[tp]
\centering
\renewcommand{\arraystretch}{0.40}
\addtolength{\tabcolsep}{-0.4pt}
\caption{The impact of the position of intermediate points on OmniISR's training performance}
\vspace{-0.2cm}
\begin{tabularx}{\linewidth}{|c|cccc|}
\hline
& \textbf{mIoU} & \textbf{mPre} & \textbf{mRec } & \textbf{mF1} \\
\hline

\verticaltext[38pt]{\textbf{Single-Point}} &\hspace{-0.21cm}
\includegraphics[width=0.2365\linewidth]{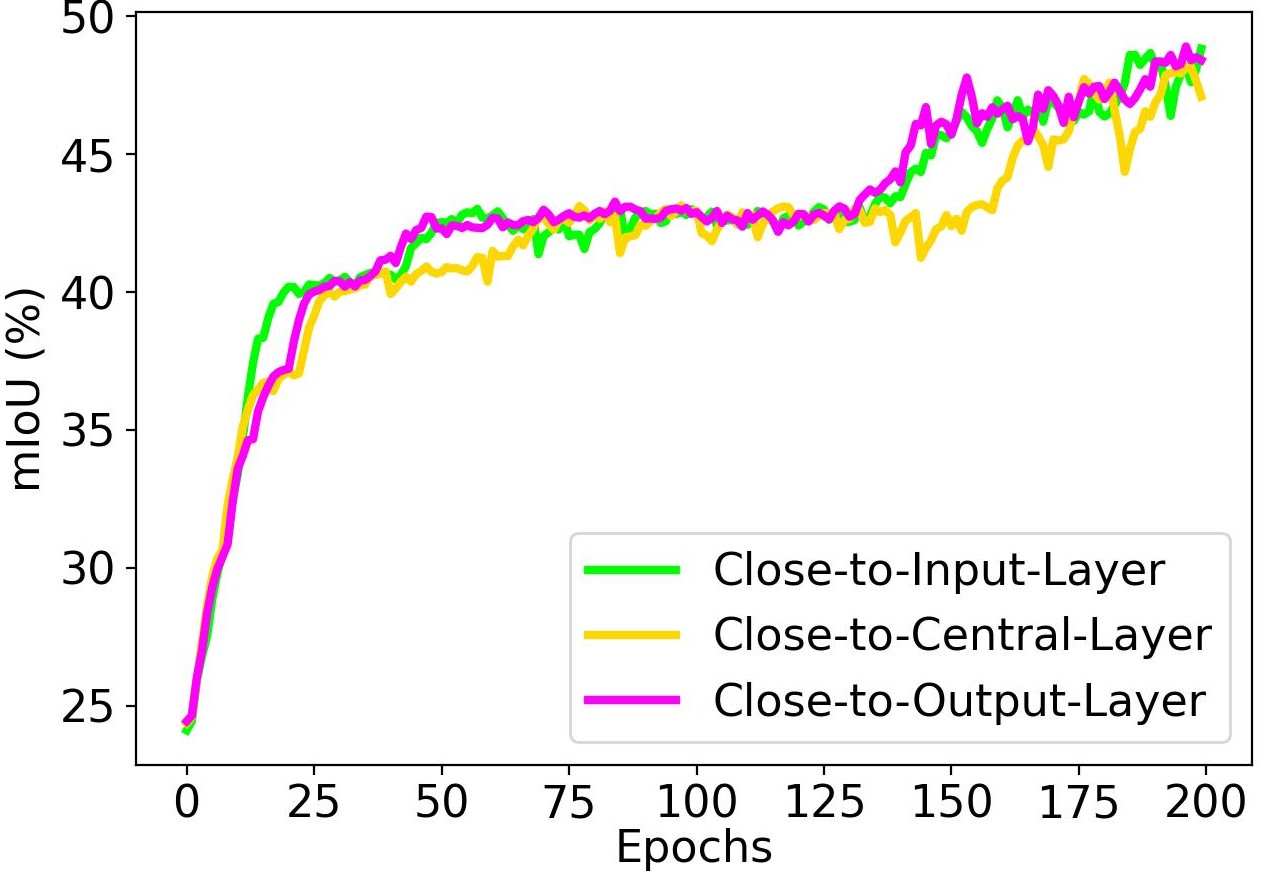} &\hspace{-0.47cm}
\includegraphics[width=0.2365\linewidth]{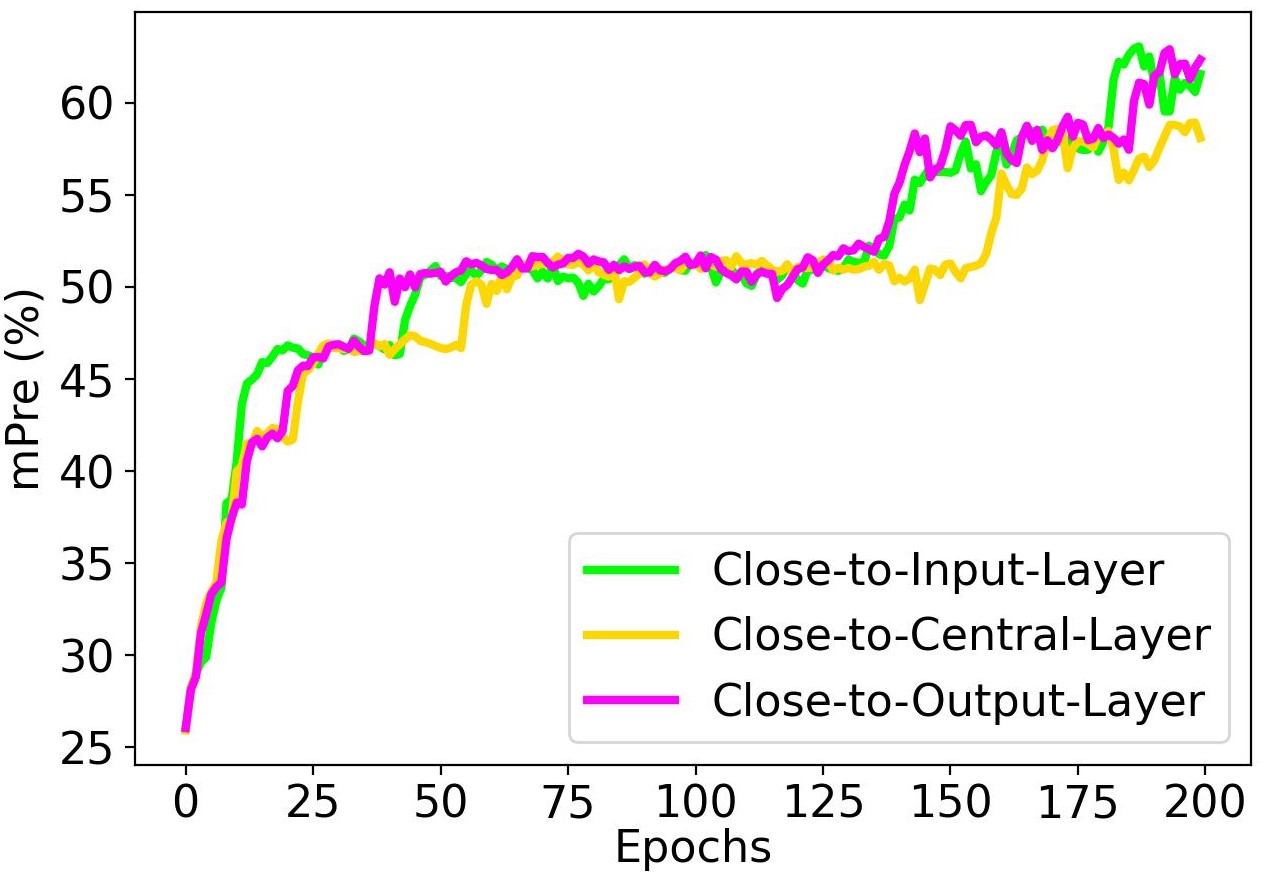} &\hspace{-0.47cm}
\includegraphics[width=0.2365\linewidth]{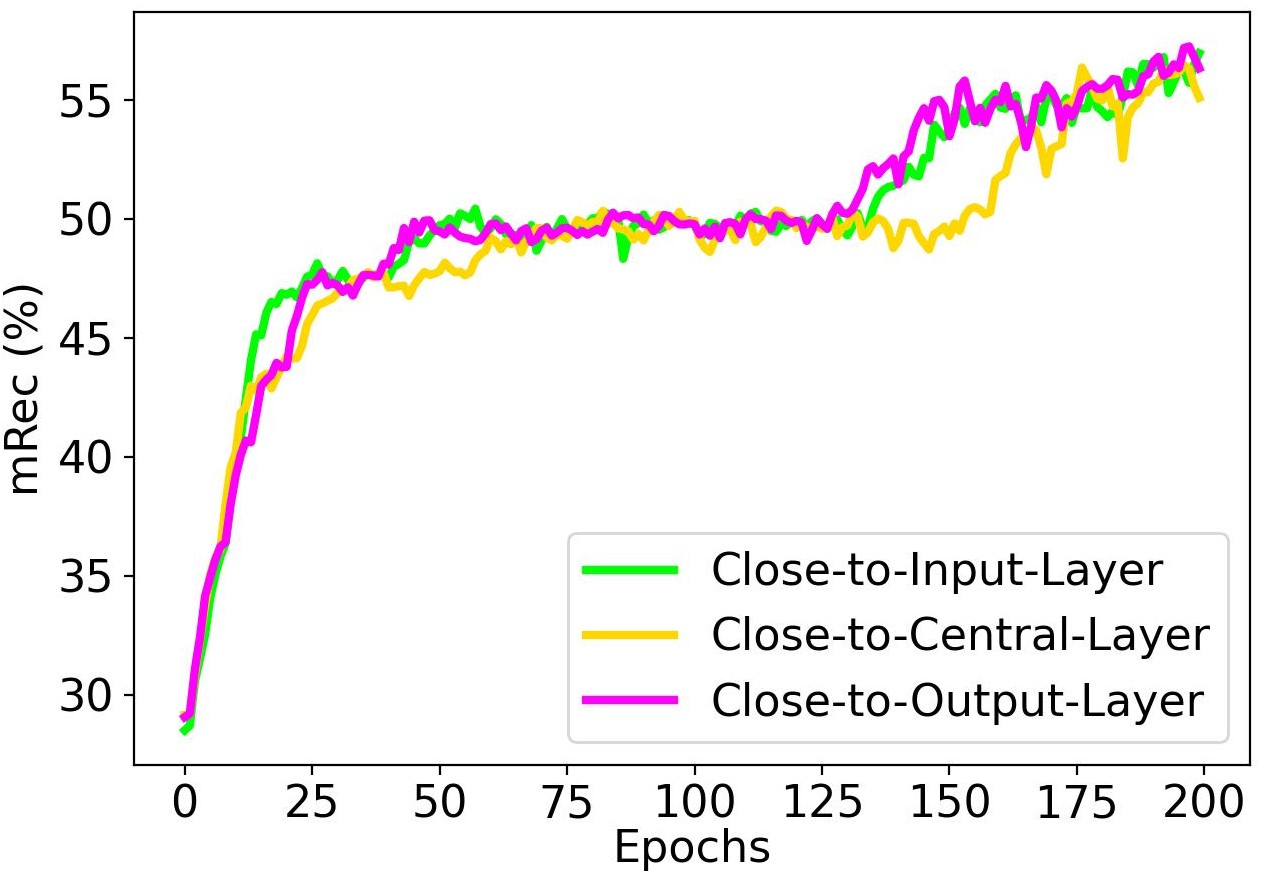} &\hspace{-0.47cm}
\includegraphics[width=0.2365\linewidth]{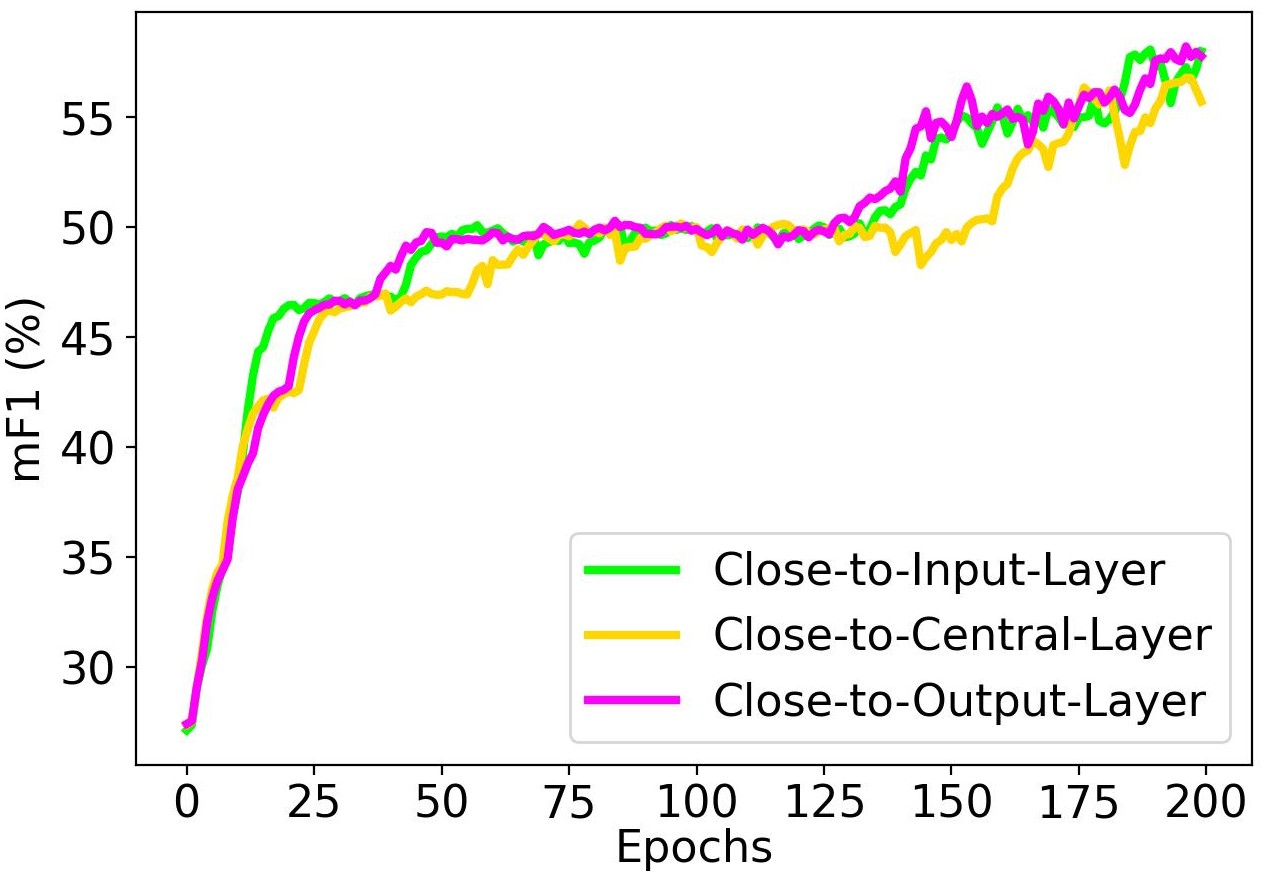 } \\
\hline

\verticaltext[38pt]{\textbf{Two-Point}} &\hspace{-0.21cm}
\includegraphics[width=0.2365\linewidth]{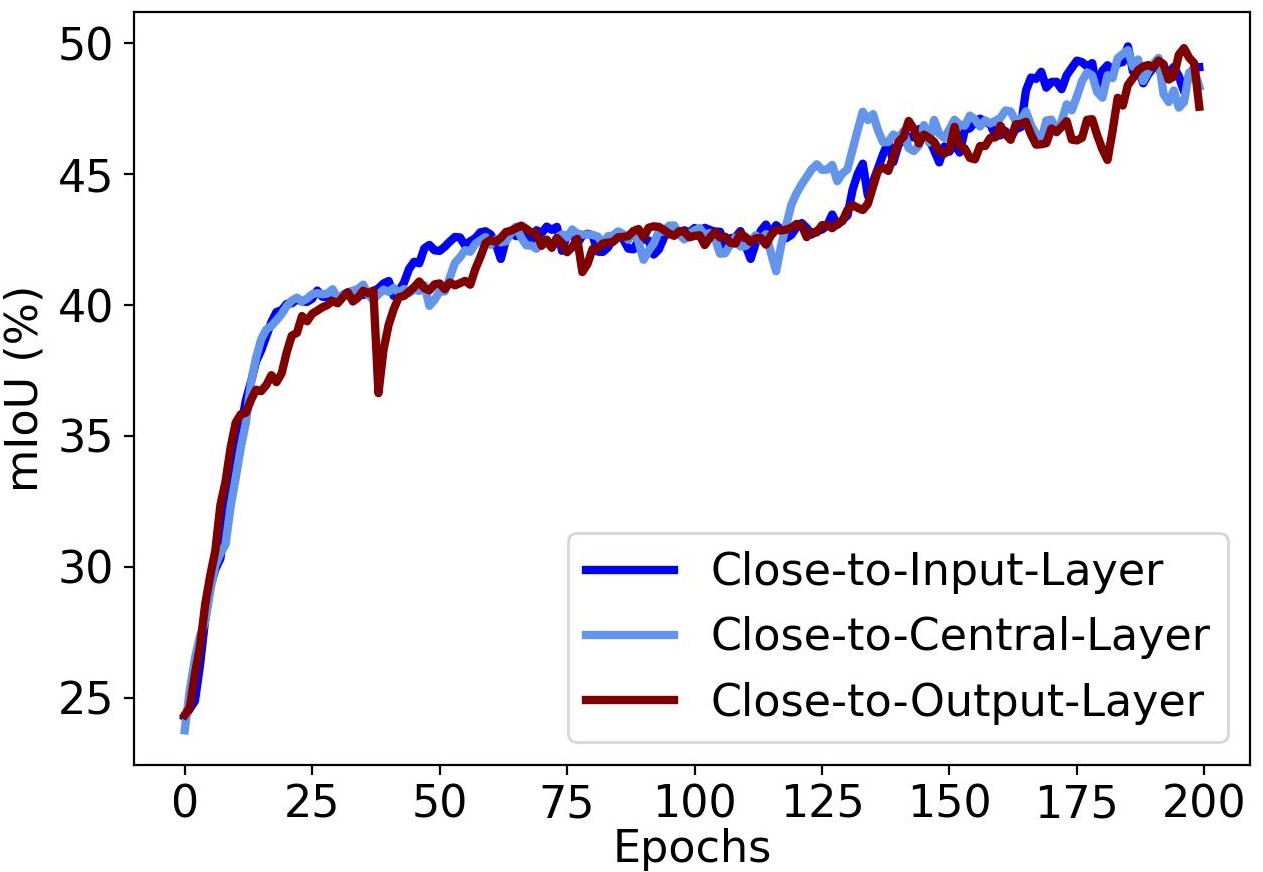} &\hspace{-0.47cm}
\includegraphics[width=0.2365\linewidth]{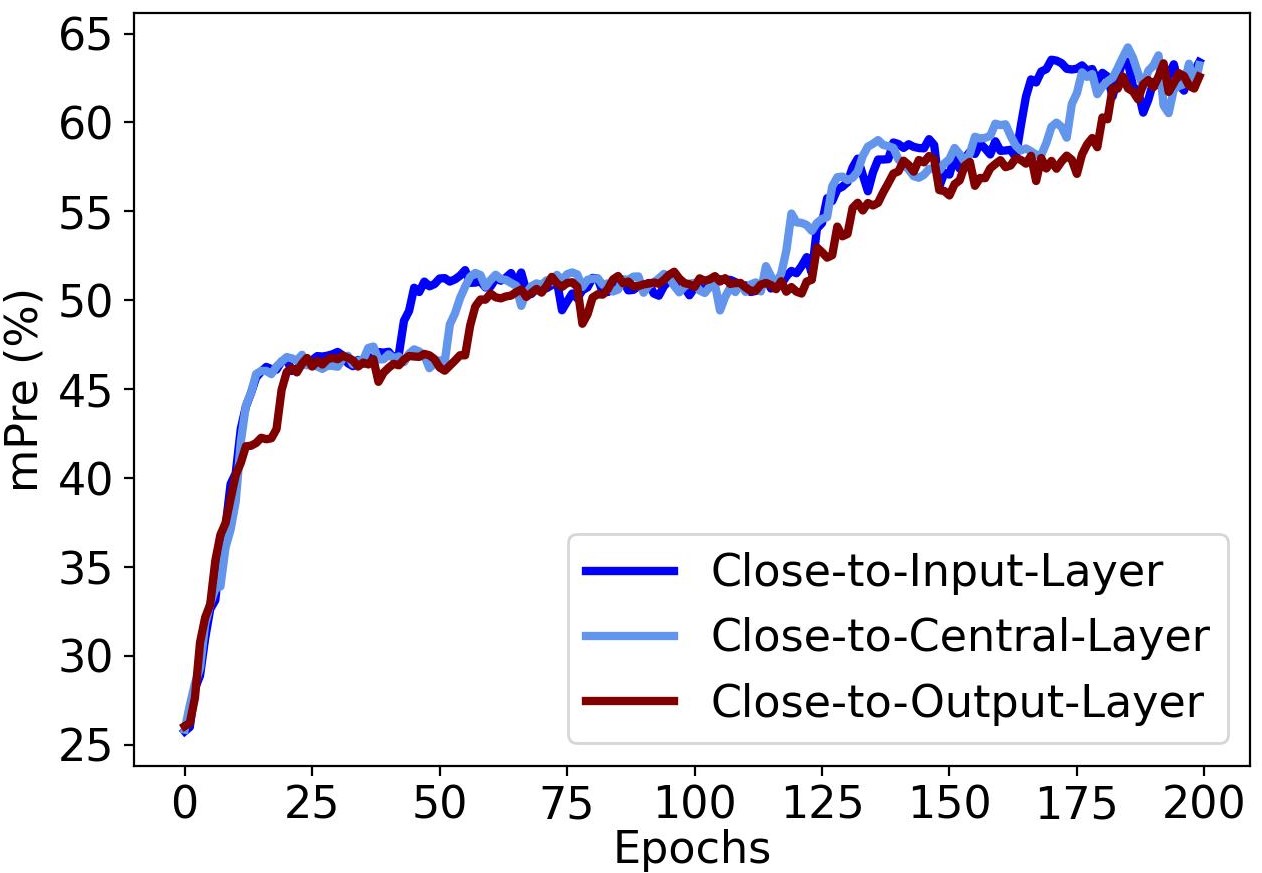} &\hspace{-0.47cm}
\includegraphics[width=0.2365\linewidth]{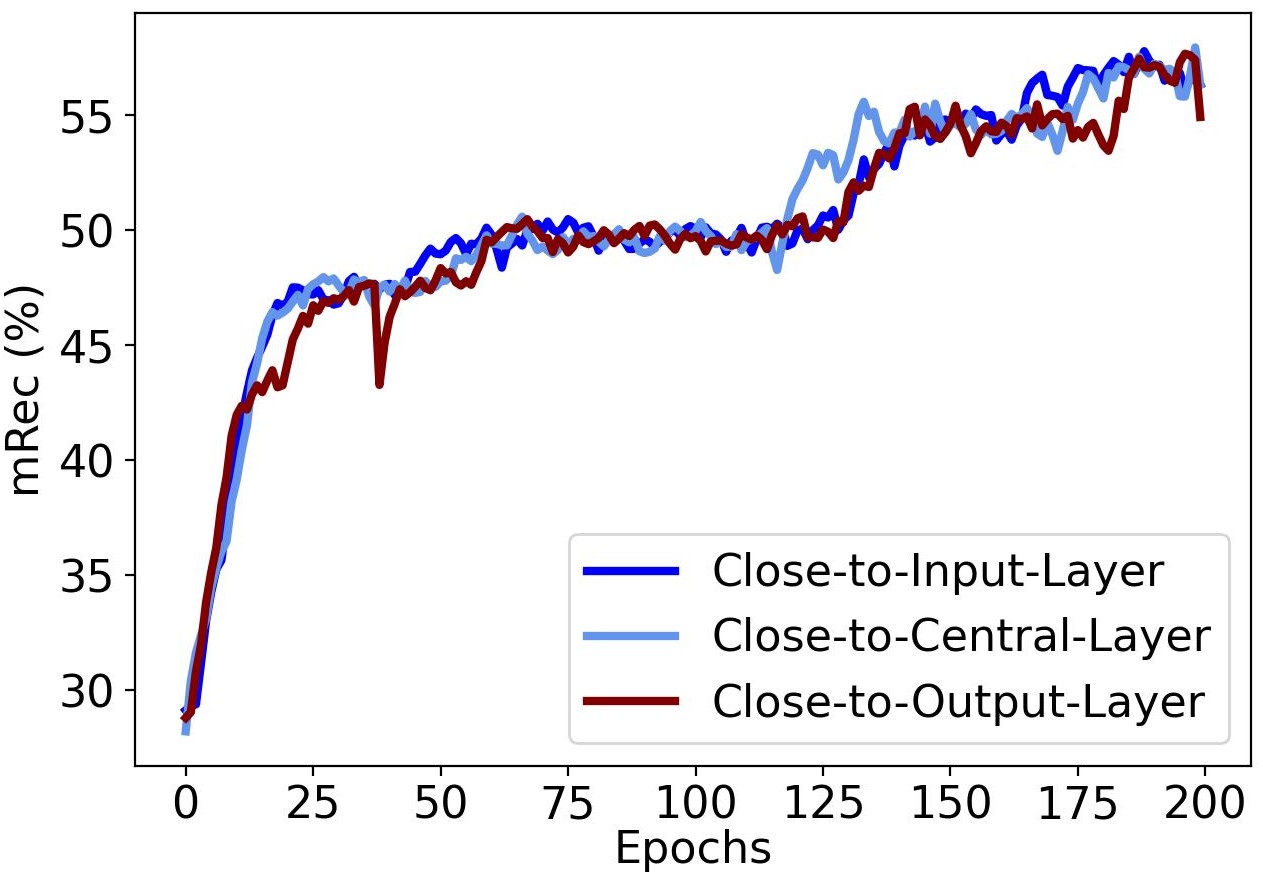} &\hspace{-0.47cm}
\includegraphics[width=0.2365\linewidth]{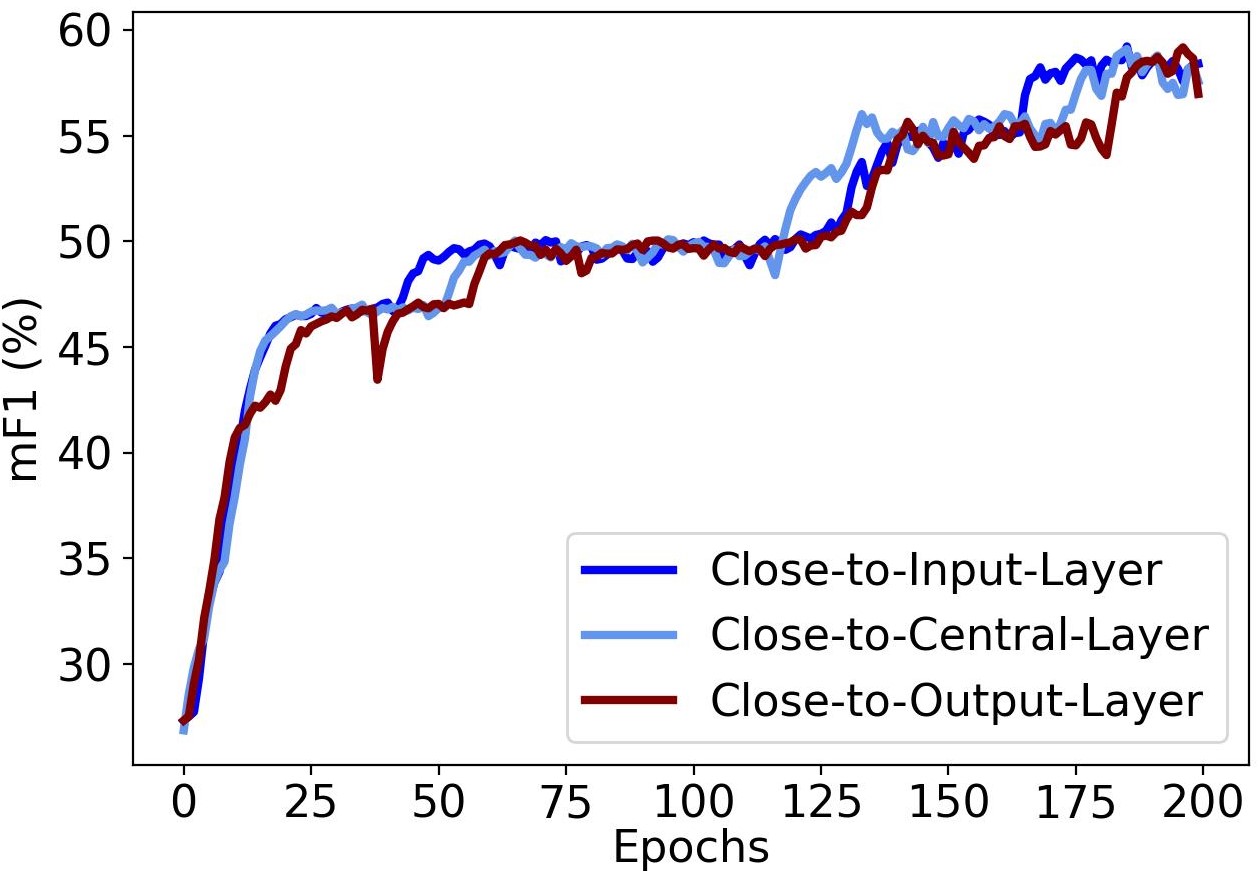 } \\
\hline

\verticaltext[38pt]{\textbf{Three-Point}} &\hspace{-0.21cm}
\includegraphics[width=0.2365\linewidth]{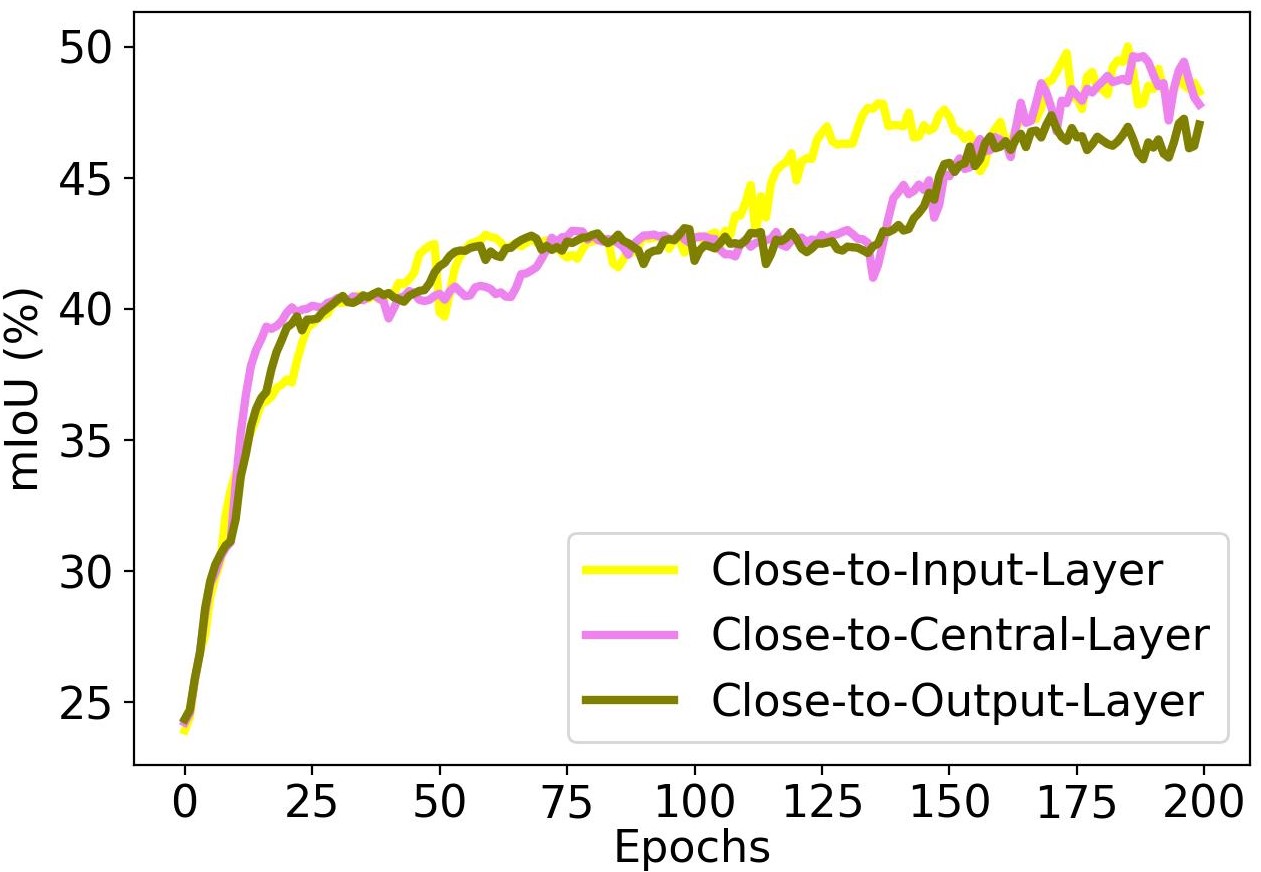} &\hspace{-0.47cm}
\includegraphics[width=0.2365\linewidth]{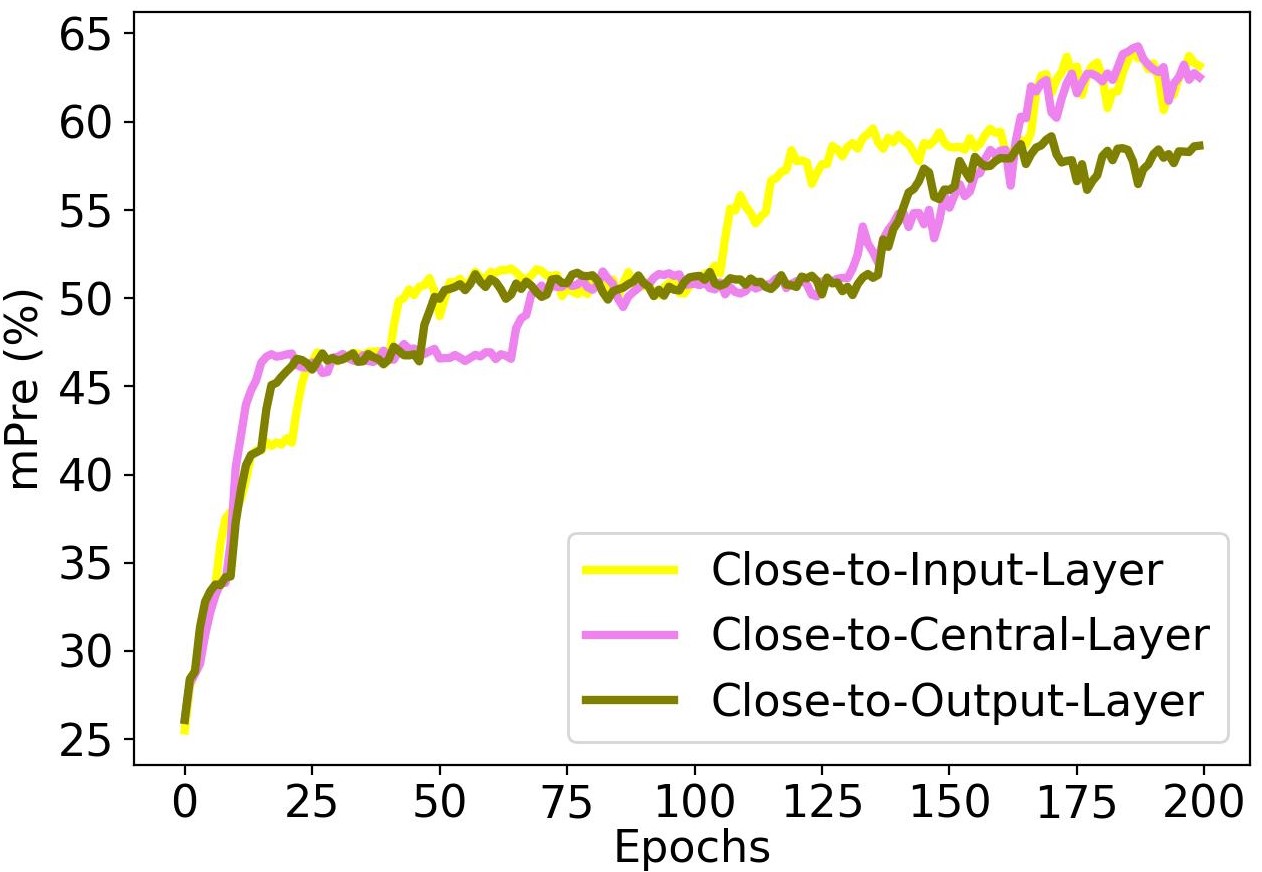} &\hspace{-0.47cm}
\includegraphics[width=0.2365\linewidth]{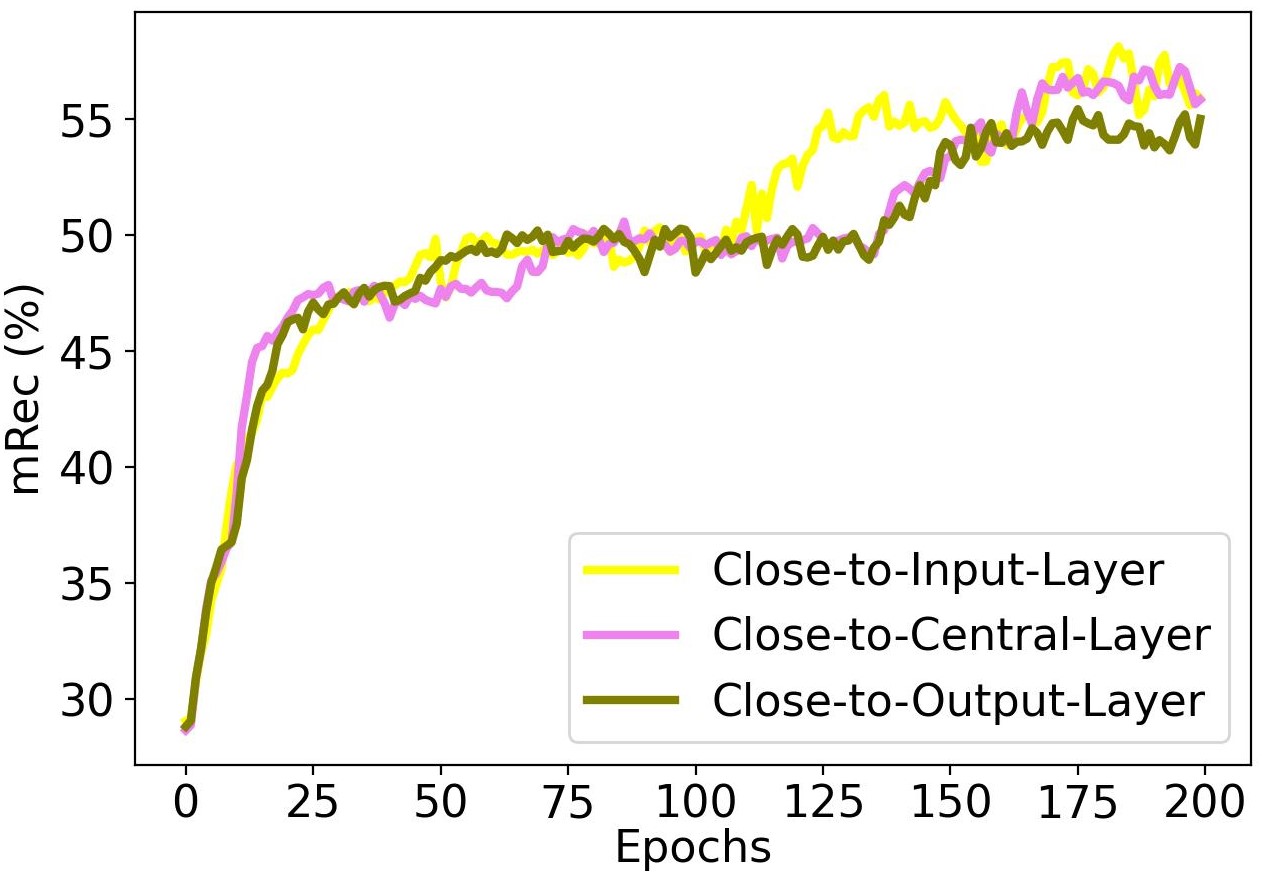} &\hspace{-0.47cm}
\includegraphics[width=0.2365\linewidth]{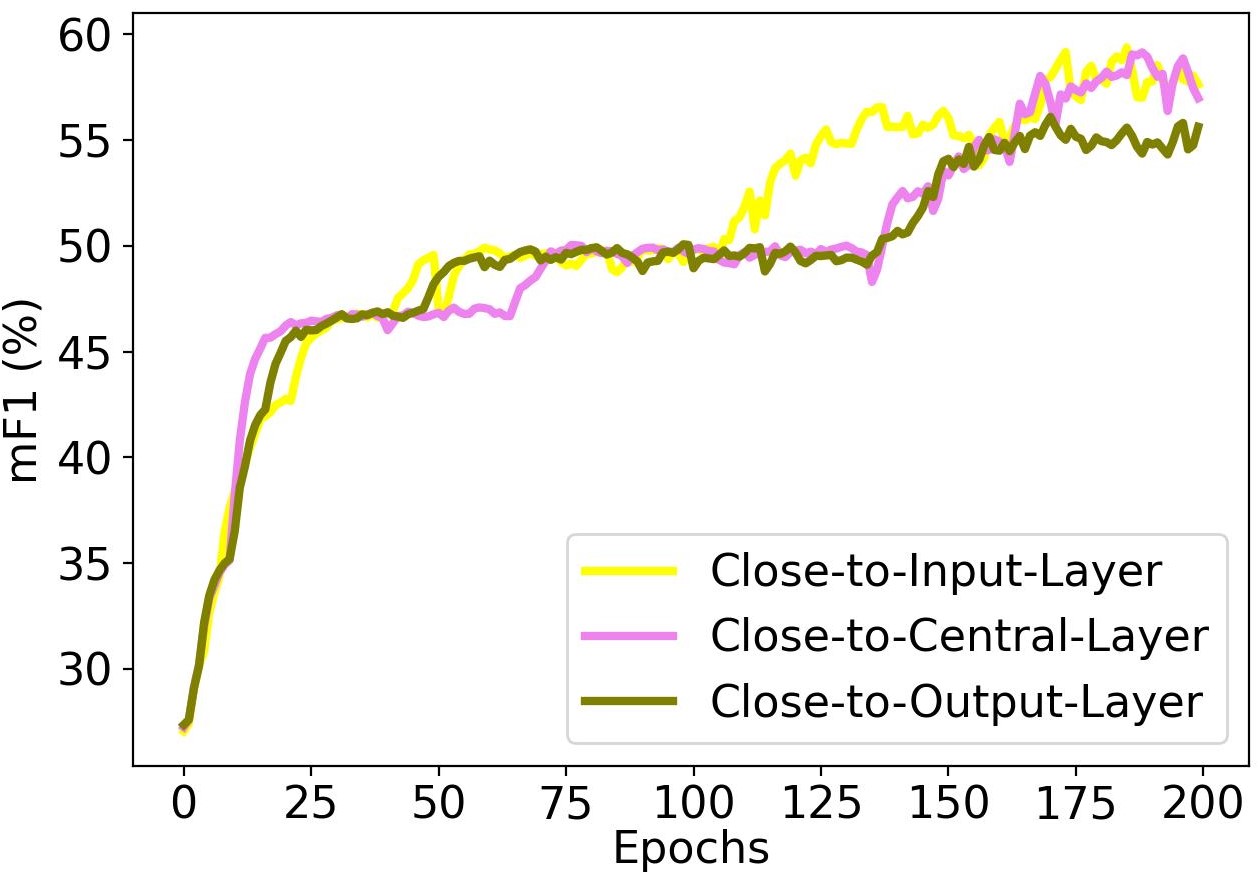 } \\
\hline
\end{tabularx}
\label{tab:abl_position_metric_comp}
\vspace{-0.3cm}
\end{table*}

\subsubsection{Impact of position of intermediate points}
To determine the most effective locations for intermediate constraints, we evaluate three placement strategies:
\begin{itemize}[leftmargin=*]
    \item \textbf{Series I:} A single intermediate point located near the input, middle, or output stage.
    \item \textbf{Series II:} Two intermediate points, following the same positional biases.
    \item \textbf{Series III:} Three intermediate points, again comparing input-, middle-, and output-biased configurations.
\end{itemize}

The results, summarized in \Cref{tab:abl_position_metric_comp}, reveal a consistent pattern regarding the network's learning dynamics. Across all three series, input-biased placement yields the best performance. This suggests that applying guidance at early layers effectively stabilizes shared, low-level feature representations before client-specific drift can accumulate in the deeper layers. Conversely, the performance of output-biased placement degrades as the number of constraint points increases. This decline likely occurs because excessive constraints at later stages interfere with task-specific adaptation and limit the flexibility of the final decision layers. 

Interestingly, middle-biased placement improves as more constraint points are added. This indicates that distributing supervision across the intermediate layers (the network's ``waist'') enhances semantic consolidation without overly restricting the final output logits. Ultimately, these findings offer a practical guideline for deployment: intermediate constraints should be prioritized in the early-to-middle stages of the network, while stacking multiple constraints near the output head should be avoided.

To move from ablation analysis to actionable practice, we summarize a suite of concrete configuration policies. \textbf{Step 1: choose point count in the mid-range} to avoid both under-guidance and over-constraint. \textbf{Step 2: choose moderate spacing} to balance local consistency and global abstraction. \textbf{Step 3: bias placement toward early-to-middle stages}, especially under strong non-IID drift. 

\section{Conclusion}
\label{sec:conclusion}

We presented OmniISR, a unified CL/FL/hybrid framework that couples intermediate MI supervision and NE regularization under one optimization objective. Rather than claiming novelty from individual components in isolation, we emphasize the innovation in their cross-paradigm coupling and corresponding optimization theory.

\begin{enumerate}
  \item We formulate a unified objective that can be instantiated in centralized, federated, and hybrid modes without architecture-specific redesign.
  \item We show that heterogeneous intermediate supervision plus uncertainty-aware regularization can jointly mitigate representation drift and preserve generalization, especially under non-IID federated conditions.
  \item Theoretically, we establish four key results: (i) a unified, ISR-agnostic, and non-asymptotic \(\mathcal{O}(1/\sqrt{T})\) convergence bound applicable across pure and hybrid CL--FL paradigms; (ii) a federated drift-bound quantifying the ISR-reduced client drift; (iii) a gradient-alignment guarantee ensuring update consistency; and (iv) an explicit time bound demonstrating that hybrid mixing accelerates the escape from strict saddle points.
  \item Empirically, OmniISR improves a broad benchmark matrix, narrows the CL--FL quality gap in representative settings, and exhibits cross-optimizer positive transfer.
\end{enumerate}

In the future, three extensions are most impactful: (i) comprehensive multi-seed statistical intervals, (ii) explicit hybrid-mode empirical sweeps across cloud/client data ratios and mixing policies, and (iii) full accuracy--communication--computation Pareto reporting under realistic edge constraints.